\documentclass[letterpaper,twocolumn,10pt]{article}
\usepackage{usenix_style}

\usepackage{booktabs} 
\usepackage{balance}
\usepackage{textcomp}
\usepackage{subcaption}
\usepackage{graphics, epstopdf, graphicx, epsfig, psfrag, epsf, wrapfig}
\usepackage{amssymb, amsfonts, amsmath}
\usepackage{mathtools}
\usepackage{multicol, multirow, verbatim, caption,float}
\usepackage{enumerate}

\usepackage{thmtools} 
\usepackage{thm-restate}

\usepackage[listings,skins]{tcolorbox}
\usepackage[ruled]{algorithm2e}
\usepackage{xspace}
\usepackage{xcolor, color}
\usepackage{hyphenat}
\usepackage{array, calc}

\newcommand{\ie}{{\em i.e., }}
\newcommand{\eg}{{\em e.g., }}
\newcommand{\etc}{{\em etc. }}

\newcommand{\textvtt}[1]{\texttt{#1}}

\DeclarePairedDelimiterX{\infdivx}[2]{(}{)}{%
  #1\;\delimsize\|\;#2%
}

\usepackage{xcolor}
\definecolor{heraldBlue}{rgb}{0.0,0.0,0.8}
\definecolor{heraldGreen}{rgb}{0.0,0.4,0.0}
\definecolor{heraldPurple}{rgb}{0.9,0.1,0.9}

\newcommand{\campus}{\textvtt{Campus Dataset}\xspace}
\newcommand{\radiocells}{\textvtt{Radiocells Dataset}\xspace}

\newcommand{\blue}[1]{#1}

\newcommand{\interval}{{time interval}\xspace}

\newcommand{\LocalBatch}{{LocalBatch}\xspace}

\newcommand{\algoname}{\textvtt{Diverse Batch}\xspace}
\newcommand{\improalgoname}{\textvtt{Farthest Batch}\xspace}

\newcommand{\aka}{a.k.a.\xspace}

\newcommand{\SP}{${}$\hspace{0.25cm}}

\newcommand\blfootnote[1]{%
  \begingroup
  \renewcommand\thefootnote{}\footnote{#1}%
  \addtocounter{footnote}{-1}%
  \endgroup
}

\usepackage[skipbelow=\topskip,skipabove=\topskip]{mdframed}

\usepackage{url}
\usepackage{breakurl}

\makeatletter
\def\url@leostyle{%
	\@ifundefined{selectfont}{\def\UrlFont{\sf}}{\def\UrlFont{\small\ttfamily}}}
\makeatother

\AtBeginDocument{%
  \providecommand\BibTeX{{%
    \normalfont B\kern-0.5em{\scshape i\kern-0.25em b}\kern-0.8em\TeX}}}

\usepackage{hyperref}
\usepackage{cleveref}

\begin{document}

\title{Location Leakage in Federated Signal Maps}


\author{
{\rm Evita Bakopoulou$^1$$\dagger$~~~}
{\rm Mengwei Yang$^1$$\dagger$~~~}
{\rm Jiang Zhang$^2$~~~}
{\rm Konstantinos Psounis$^2$~~~}
{\rm Athina Markopoulou$^1$~~~}
\smallskip 
\\
$^1$University of California Irvine
\{ebakopou, mengwey, athina\}@uci.edu
\\
$^2$University of Southern California
\{jiangzha, kpsounis\}@usc.edu
}

\maketitle

\begin{abstract}
{%

We consider the problem of predicting cellular network performance (signal maps)
from measurements collected by several mobile devices. We formulate the problem within the online federated learning framework: (i) federated learning (FL) enables users to collaboratively train a model, while  keeping their training data on their devices; (ii) measurements are collected as users move around over time and are used for local training in an online fashion. 
We consider an honest-but-curious server, who observes the updates from target users participating in FL and infers their location using a deep leakage from gradients (DLG) type of attack, originally developed to reconstruct  training data of DNN image classifiers. We make the key observation that a DLG attack, applied to our setting, infers the average location of a batch of local data, and can thus be used to reconstruct the target users' trajectory at a coarse granularity. We build on this observation to protect location privacy,  in our setting, by revisiting and designing mechanisms within the federated learning framework including: tuning the FL parameters for averaging, curating local batches so as to mislead the DLG attacker, and aggregating across multiple users with different trajectories. 
We evaluate the performance of our algorithms through both analysis and simulation based on real-world mobile datasets, and we show that they achieve a good privacy-utility tradeoff. %

} %
\end{abstract}

\blfootnote{\blue{$\dagger$The first two authors made equal contributions.} E. Bakopoulou was with the University of California, Irvine, when this work was conducted; she is currently with Google, Mountain View, CA. 
This work has been supported by NSF Awards 1956393, 1900654,  1901488, and 1956435.
}

\blfootnote{Accepted in IEEE Transactions on Mobile Computer in October 2023.}

\section{Introduction}\label{sec:introduction}

Mobile crowdsourcing is widely used to collect data from a large number of mobile devices, which are useful on their own and/or used to train models for properties of interest, such as cellular/WiFi coverage, sentiment, occupancy, temperature, COVID-related statistics, etc. Within this broader class of spatiotemporal models trained by mobile crowdsourced data \cite{carmela}, we focus on the representative and important case of cellular signal maps. %
Cellular operators rely on key performance indicators (\aka KPIs) to understand the performance and coverage of their network, in their effort to provide the best user experience. These KPIs include wireless signal strength measurements, especially Reference Signal Received Power (RSRP), which is going to be the focus of this paper, and other performance metrics (\eg coverage, throughput, delay) as well as information associated with the measurement (\eg  location, time, frequency band, device type, etc.). 

\begin{figure*}[t!]
	\centering
	\begin{subfigure}{0.33\textwidth}
    	\includegraphics[width=0.99\textwidth]{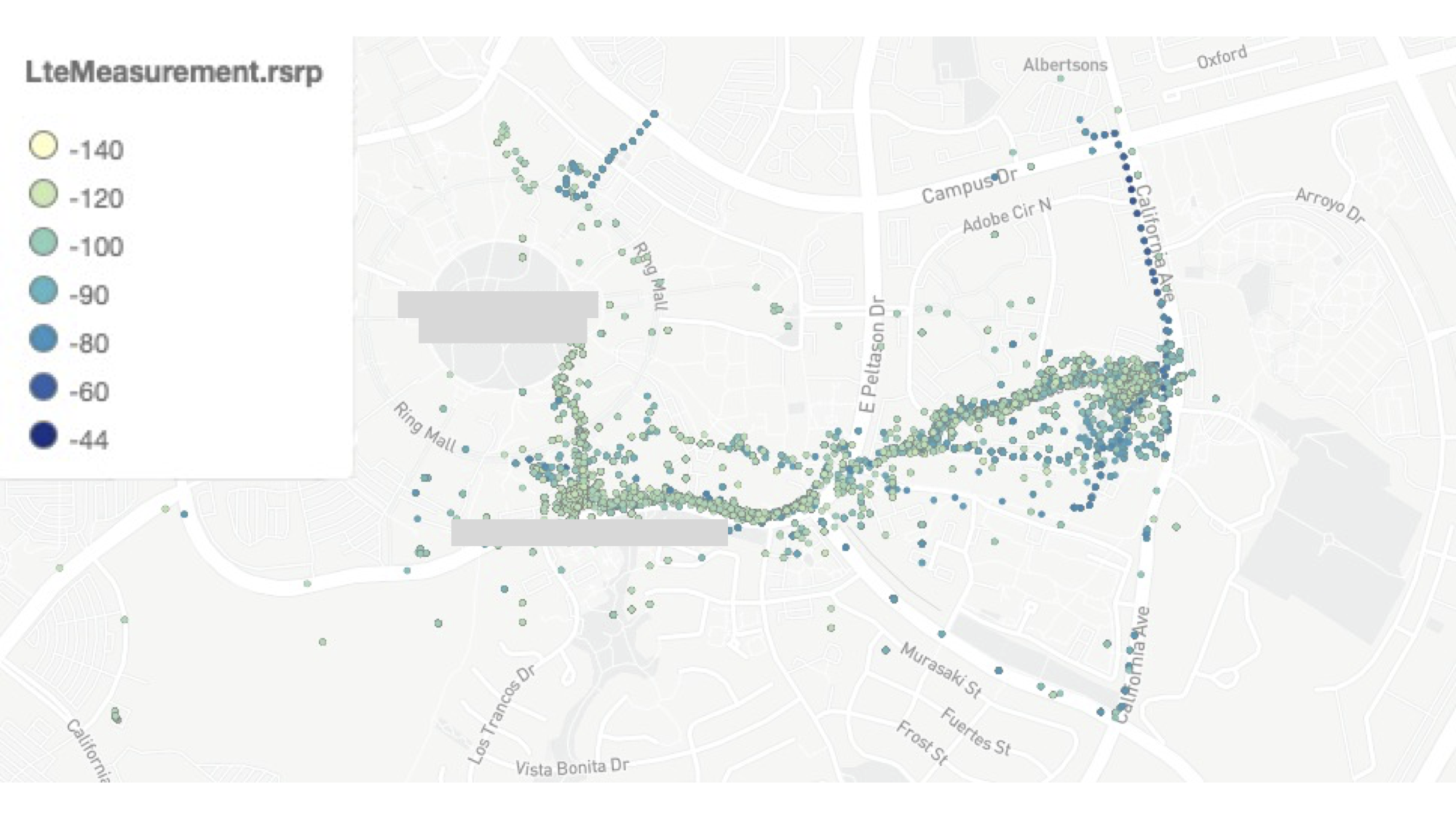}
        \caption{RSRP measurements of one user.}
     \label{fig:all_cells_density_maps_user0}
	\end{subfigure}
	\begin{subfigure}{0.33\textwidth}
    	\includegraphics[width=0.99\textwidth]{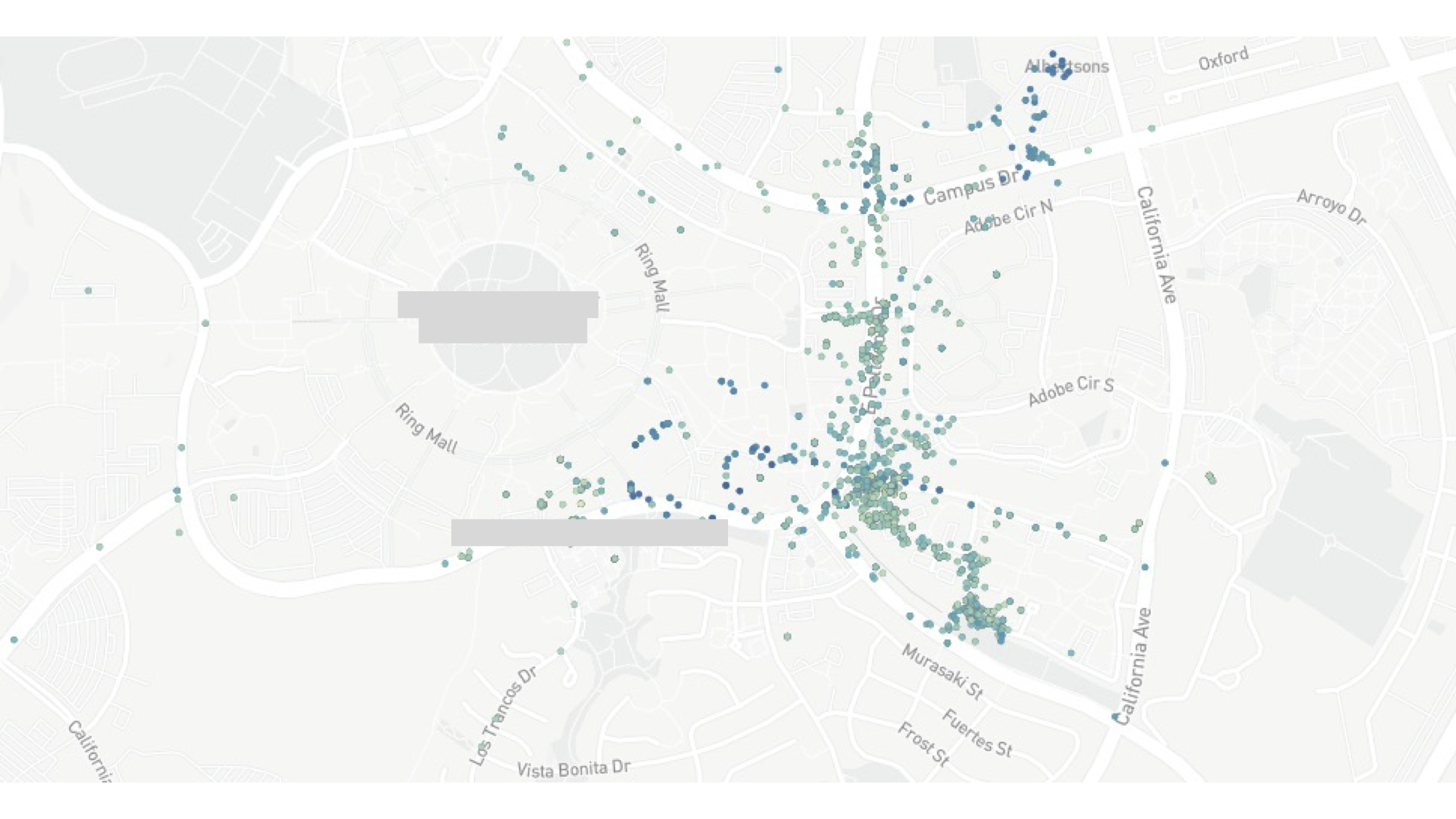}
        \caption{ RSRP measurements of another user.}
     \label{fig:all_cells_density_maps_user1}
	\end{subfigure}
	\centering
	\begin{subfigure}{0.33\textwidth}
    	\includegraphics[width=0.99\textwidth]{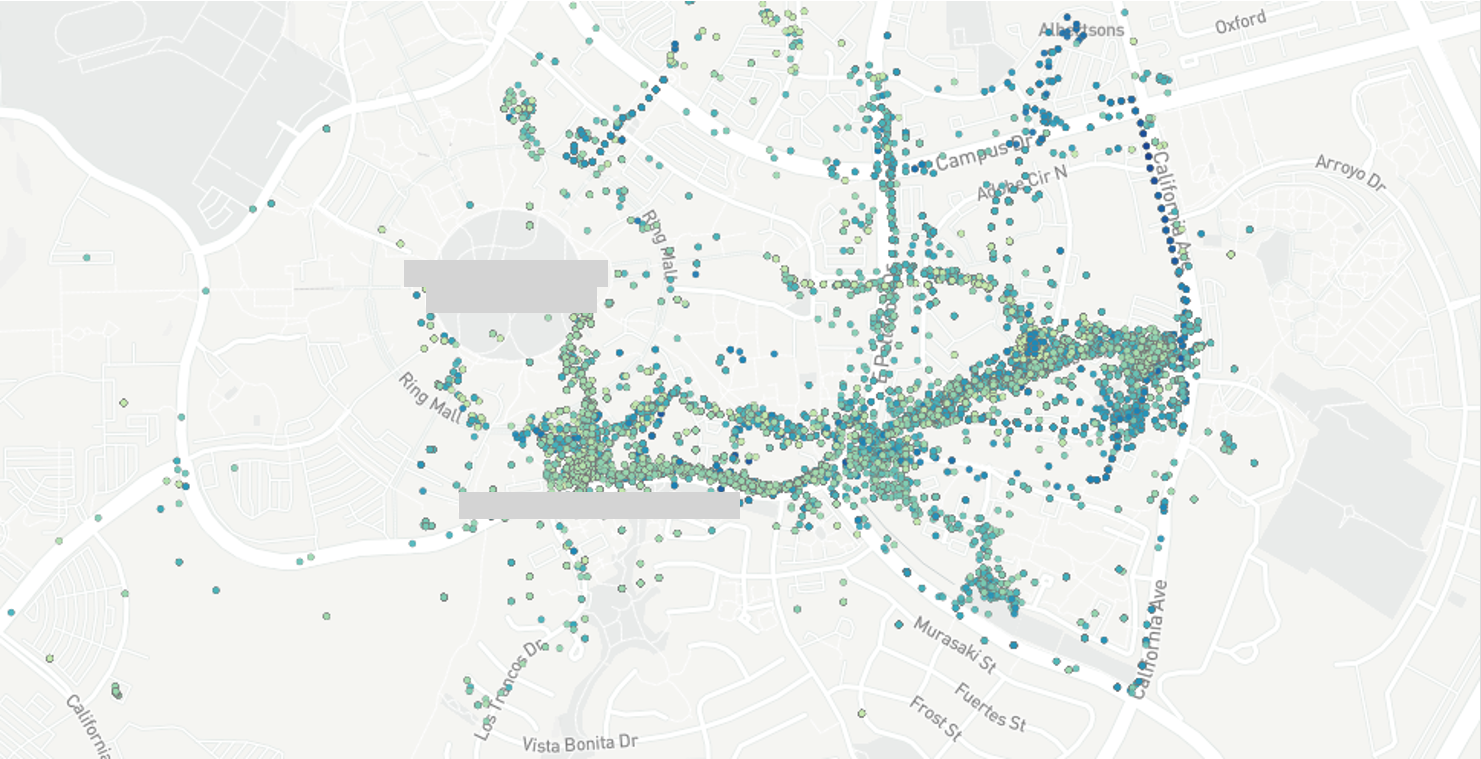}
    	\caption{ RSRP measurements from all users.}
    	\label{fig:all_cells_density_maps_merged}
	\end{subfigure}
	\caption{{\blue{\small Examples from the \campus. We see the locations where measurements are collected and the color indicates the values of signal strength (RSRP) in those locations. Fig. (a) and (b) depict measurements collected by two different users. Fig. (c) shows the union of all measurements from all users in the \campus. } }}
	\label{fig:density_maps_uci}
\vspace{-5pt}
\end{figure*}

Cellular signal strength maps consist of KPIs in several locations. Traditionally, cellular operators collected such measurements by hiring dedicated vans (\aka wardriving~\cite{yang:10}) with special equipment, to drive through, measure and map the performance in a particular area of interest. However, in recent years, they increasingly outsource the collection of signal maps to third parties~\cite{alimpertis2019city}. Mobile analytics companies (\eg OpenSignal \cite{opensignal:11}, Tutela \cite{tutela}) crowdsource measurements directly from end-user devices, via standalone mobile apps or SDKs integrated into popular partnering apps, typically games, utility or streaming apps. %
 The upcoming dense deployment of small cells for 5G at metropolitan scales will only increase the need for accurate and comprehensive signal maps~\cite{5gonapkpis,imran:14}.
Because cellular measurements are expensive to obtain, they may not be available for all locations, times and other parameters of interest, thus there is a need for signal maps prediction based on limited available such measurements.  

Signal maps prediction is an active research area and includes: propagation models~\cite{raytracing:15, winnerreport}, data-driven approaches~\cite{fidaZipWeave:17,specsense:17,heBCS:18},  combinations thereof~\cite{phillips:12}, and increasingly sophisticated ML models for RSRP %
~\cite{ray:16,raik:18,alimpertis2019city, Krijestorac2021SpatialSS} and throughput~\cite{infocomLSTM:17,milanMobihocDeepnets:18}. All these prediction techniques consider a  centralized setting: mobile devices upload  measurements to a server, which then trains a global model to predict cellular performance (typically RSRP) based on at least location, and potentially time and other features. %

Fig. \ref{fig:density_maps_uci} depicts an example of a dataset collected on a university campus, which is one of the datasets we use throughout this paper. 
Fig. \ref{fig:density_maps_uci}(a) and (b) show the locations where measurements of signal strength (RSRP) were collected by two different volunteers as they move around the campus. The measurements from all users are uploaded to a server, which then merges them and creates a signal map for the campus (shown in Fig. \ref{fig:density_maps_uci}(c)); and/or may train a global model for predicting signal strength based on  %
 location and potentially other features. However, this utility comes at the expense of privacy: as evident in Fig. \ref{fig:density_maps_uci},  frequently visited locations may reveal users' home, work, or other important locations, as well as their mobility pattern; these may further reveal sensitive information such as their medical conditions, political beliefs, etc \cite{nytimesarticle}. The trajectories of the two example users are also sufficiently different from each other, and can be used to distinguish between them, even if their identifiers are removed from the dataset.

\blue{
In this paper, we make three contributions (in the problem setup, privacy attack and defense mechanisms), all leveraging the patterns of human mobility underlying our data. First, we design a lightweight online federated learning framework, specifically for the signal strength prediction {\em problem}. Second, we introduce a   {\em privacy attack}, specifically for this framework: an honest-but-curious server employs gradient inversion to infer the location of users participating in the federated signal map framework. This DLG-based attack is specifically designed to reconstruct the average location in each round; this is in contrast to state-of-the-art DLG attacks on images or text, which aims at fully reconstructing all training data points. Third, we propose a {\em defense} approach that selects local batches so that the inferred location is far from the true average location, thus misleading the DLG attacker. Evaluation results show that our defense mechanisms achieve better privacy-performance trade-off compared to state-of-the-art baselines.}

First, w.r.t. the {\em signal maps prediction} based on crowdsourced data: we formulate a simple version that captures the core problem. We train a DNN  to predict  signal strength (RSRP) based on GPS location (latitude, longitude), while local training data arrive in an online fashion. The problem lends itself naturally to Federated Learning (FL):  training data are collected by mobile devices, which want to collaborate without uploading sensitive location information.  FL enables mobiles to do that by exchanging model parameters with the server but keeping their training data local \cite{original_federated}. The problem further lends itself to online learning because the training data are collected over time \cite{bottou2004large, li2014efficient} as users move around. We design  a lightweight online FL scheme, which trains only on data collected during the current round, and we show that it performs well in this setting.

Second, w.r.t. the {\em location privacy}: we consider an honest-but-curious server, which implements online FL accurately but attempts to infer the location of users. %
Since gradient updates are sent from users to the server in every FL round, %
 FL lends itself naturally to inference attacks from gradients.  We adapt the DLG attack, originally developed to reconstruct training images and text used for training DNN classifiers \cite{dlg,geiping2020inverting}.  A key observation, that we confirm both empirically and analytically, is that a DLG attacker who observes a single gradient (SGD) update from a target user, can reconstruct the  {\em average} location of points in the batch. %
Over multiple rounds of FL, this allows the reconstruction of the target(s)' mobility pattern. %

Third, on the defense side, we leverage  this intuition to design local mechanisms that are inherent to FL (which we refer to as "FL-native") specifically to mislead the DLGs attacker and protect location privacy. In particular, 
we show that the averaging of gradients inherent in FedAvg provides a moderate level of protection against DLG, while simultaneously improving utility; we systematically evaluate the effect of multiple federate learning parameters ($E, B, R, \eta$) on the success of the attack. Furthermore, we design and evaluate two algorithms for local batch selection, \algoname and \improalgoname, that a mobile device can apply locally to curate its local batches so that the inferred  location is far from the true average location, thus misleading the DLG attacker and protecting location privacy. %
(3) We also show that the effect of multiple users participating in FL, w.r.t. the success of the DLG attack, depends on the similarity of user trajectories.

Throughout this paper, we use two real-world datasets: (i) our own geographically small but dense \campus  \cite{alimpertis2017system} we introduced in Fig. \ref{fig:density_maps_uci}); and (ii) the larger but sparser publicly available \radiocells  \cite{radiocells}, especially its subset from the London metropolitan area. 
We show that we can achieve good location privacy, without compromising prediction performance, through the privacy-enhancing design of the aforementioned FL-native mechanisms (\ie tuning of averaging, curation of diverse and farthest local batches, and aggregation of mobile users with different trajectories).   
 Add-on privacy-preserving techniques, such as Differential Privacy (DP) \cite{dwork2011differential, naseri2020toward} or Secure Aggregation (SecAgg) \cite{fed_sec_aggregation},  are orthogonal and can be added on top of these FL mechanisms, if stronger privacy guarantees are desired, at the expense of computation or utility. Our evaluation suggests that \algoname and \improalgoname alone are sufficient to achieve  a great privacy-utility tradeoff in our setting.

The outline of the paper is as follows.  Section \ref{sec:methodology} formulates the federated online signal maps prediction problem and the corresponding DLG attack, and provides key insights. Section \ref{sec:evalsetup} describes the evaluation setup, including datasets and privacy metrics. Section \ref{sec:results} presents the evaluation results for DLG attacks without any defense, as well as with our privacy-enhancing techniques, for a range of simulation scenarios.  Section \ref{sec:related_work} discusses related work.  Section 6 concludes and  outlines future directions. The appendix -- uploaded under supplemental materials -- provide additional details on datasets, parameter tuning, analysis, and evaluation results. %

\section{Location DLG Attack}\label{sec:methodology}

In Section \ref{sec:setup}, we model the problem within the online federated learning framework and we define the  DLG attack that allows an honest-but-curious server to infer the whereabouts of the target user(s). In Section \ref{sec:analysis}, we provide analytical insights that explain the performance of the DLG attack for various user trajectories and tuning of various parameters, and also guide our algorithm design choices.

\subsection{Problem Setup\label{sec:setup}}

\textbf{Signal Maps Prediction.} Signal maps prediction typically trains a model to predict a key performance indicator (KPI) $y_i$ based on the input feature vector $x_i=[x_{i,1},x_{i,2}, ....x_{i,m}]^T$, where $i$ denotes the $i$-th datapoint in the training set. W.l.o.g. we consider the following: $y$ is a metric capturing the signal strength and we focus specifically on Reference Signal Received Power  (RSRP), which is the most widely used KPI in LTE. \blue{For the features $x$ used for the prediction of $y$, we focus on the spatial coordinates (longitude, latitude), \ie $m=2$.}\footnote{\blue{In prior work on centralized signal map prediction \cite{alimpertis2019city, alimpertis2022unified}, we assessed the feature importance  on the campus and other datasets, and found location, time, and cell tower id to be the most important, while device type, frequency and outdoor/indoor location had negligible effect on our campus datasets.  In this paper, we focus on the most important features, \ie spatial coordinates, we handle time within the online learning framework, and we train one DNN per cell id.}} %
\blue{We train a DNN model with weights $w$, per cell tower, $y_i=F(x_i, w)$; the  loss function $\ell$ is the Mean Square Error, and we report the commonly used  Root Mean Squared Error (RMSE).} %

We consider a general DNN architecture that, unlike prior work \cite{dlg},  is quite general. We tune its hyperparameters (depth, width, type of activation functions, learning rate $\eta$) via the Hyperband tuner \cite{hyperband_tuner} to maximize utility. Tuning the DNN architecture %
 can be done using small datasets per cell tower, which are collected directly or contributed by users or third parties willing to share/sell their data.

\begin{figure*}[t!]
		\centering
    \includegraphics[width=0.85\linewidth]{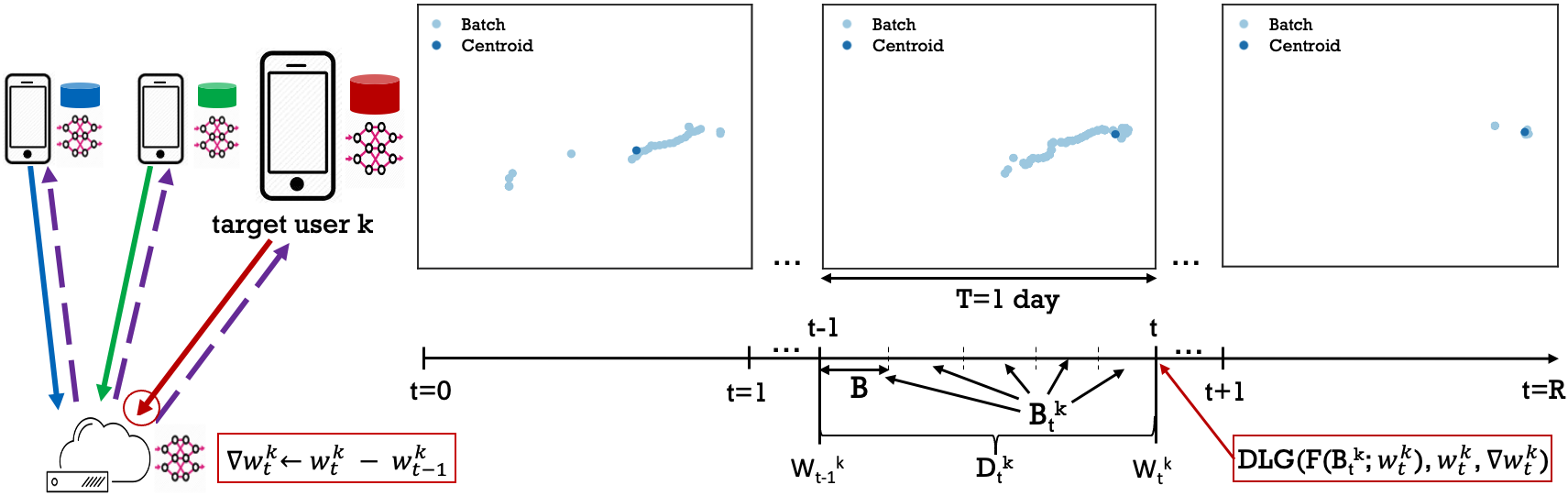}
\vspace{-5pt}
\caption{\blue{\small {\bf Example of the Online Federated Learning Framework with DLG Attack, using data from the campus.}} The target user $k$  collects data in an online fashion, and processes them in intervals/rounds of duration $T$=1 day. The right part of the picture shows real locations (in light blue)  the target user visited  on different days.  During round $t$, the target collects local data $D_t^k$, uses it for local training,  updates the local model weights $w_t^k$ and shares them with the server. (During local training, the local data $D_t^k$ is further split into a list $B_t^k$ of mini-batches, each of size B.) The server observes the model parameter update $w_t^k$ at time t, computes the gradient ($\nabla w=w_t^k-w_{t-1}^k$) and launches a DLG attack using Algorithm \ref{alg:DLG}. For each day $t$, it manages to reconstruct the centroid (average location) of the points in $D_t^k$ (shown in dark blue color). During the last day, where the user did not move much, the centroid conveys quite a a lot of information.} %

\vspace{-5pt}
\label{fig:example_local_data_per_round}
\end{figure*}

\textbf{Measurements over time and space.} We consider several users, each with a mobile device, who collect several signal strength measurements ($\{x_i, y_i\}_{i=1}^{i=N}$), as they move around throughout the day and occasionally upload some information to the server. Fig.\ref{fig:density_maps_uci} shows examples of users moving around on a university campus; Fig. \ref{fig:example_local_data_per_round} shows a single such user and the locations where measurements were collected for three different days. 
It is important to note that  the  measurement data are not static but collected in an online fashion. \blue{Users continuously collect such measurements as they move around throughout the day,  and they periodically upload them to the server, \eg every night when the mobile is plugged into charge and connected to WiFi.} \blue{This is a special case of mobile crowdsourcing (MCS) \cite{carmela}.} 

Let the time be divided into \blue{time intervals or "rounds"} indexed by $t=1,\dots,R$. \blue{All rounds have the same duration $T$;} in the previous example, \blue{$T$ was one day, but we also consider other values: 1-3 hours, one day, one week, etc.}
\footnote{\blue{Values of $T$ were chosen to match the time scales of human mobility (on the order of hours, days or weeks)  and not the dynamics of the wireless channel (much shorter time scales, \eg below a second). Selecting $T$ to be one day or one week also allows for enough datapoints in a round (see Fig. \ref{fig:uci_stats_top_cells} in App. \ref{sec:appendix_data}). It also reflects common practices in crowdsourcing signal strength (apps collect measurements throughout the day but upload once, at night, when the phone is charging).}}
At the end of each such round, user $k$ processes the set of measurement data $D_t^k$  collected during that round and sends an update to the server.  We also refer to $D_t^k$ as the local data  that ``arrive'' at user client $k$ in round $t$. Collected over  $t=1,\dots,R$, $D_t^k$ reveals a lot about user k's whereabouts, as evident by the examples of Fig. \ref{fig:density_maps_uci}  and Fig. \ref{fig:example_local_data_per_round}. \blue{Human mobility is well known to exhibit strong patterns: people spend several hours a day in a few important places (\eg their home, work, and other important locations), and move between them in continuous trajectories.} The locations $x$ collected as part of the signal maps measurements $({x_i, y_i})_{i=1}^{i=N}$ essentially sample the real user's trajectory. %

\begin{table}[t!]
\centering
\smallskip\noindent
\resizebox{\linewidth}{!}{%
\begin{tabular}{@{}cl@{}}
\toprule
\centering
\textbf{Notation} & \textbf{Description}                                \\ \midrule
$x$                & Input features: (lat, lon) coordinates (and potentially more)                       \\
$y$                & Prediction label for RSRP \\
$\ell$             & MSE loss for RSRP prediction (RMSE is reported as utility)\\
$\eta$             & Learning rate   \\  
$i$                & $i^{th}$ measurement used for training: $(x_i, y_i)$ 
\\
 \hline
\blue{$T$}           & \blue{Duration of time interval/round over which users process the online data;} \\
~             & one interval corresponds to one round in FL and one local batch \\
\blue{$t$}                & \blue{Index of FL round, $t=1,2.,..R$; each round has duration $T$}  \\
$D_t^k$            & Local data arriving to user $k$ in round $t$               \\
\LocalBatch      & Subset of all local data $D_t^k$ used as a local batch in FL \\
~               & In FedAvg: \LocalBatch=$D_t^k$; \\
~                & In \algoname and \improalgoname : \LocalBatch $\subset D_t^k$ \\
$w_t$              &Global model weights at round $t$\\
$w_t^k$            & Local model weights at round $t$ from user $k$\\
$B$                & Mini-batch size; if $B=\infty$ then mini-batch = \LocalBatch\\         
$B_t^k$            & Partition of user k's \LocalBatch at round $t$ into mini-batches     \\
$E$                & Number of local epochs                                         \\
 \hline
 $\nabla{w_t^{target}}$        & Gradient of target's model weights at round $t$ \\
$\mathbb{D}$       & Cosine loss used in DLG attack (Algorithm \ref{alg:DLG})   \\
$x_{DLG}$          & Reconstructed location by DLG attack (Algorithm \ref{alg:DLG}) \\
$\bar{x_t}$        & Centroid: average location of data in a local batch: $\bar{x_t}=\sum_{i=1}^{i=N}{x_i}$  \\
$eps$        & DBSCAN parameter that controls total clusters for \algoname \\
$num$   & \improalgoname parameter that controls the number of \\
~             & measurements selected in each LocalBatch.\\
\bottomrule
\end{tabular}
} %
\caption{\small Main  parameters and notation. See also Fig. \ref{fig:example_local_data_per_round}.}
\vspace{-5pt}
\label{tab:params_table}
\end{table}

\textbf{Federated Signal Maps.}  State-of-the-art %
mobile crowdsourcing  practices \cite{carmela,tutela} rely on the  server to collect the raw measurements (in our context locations and associated RSRP measurements) from multiple users,  aggregate them into a single map, and maybe train a centralized prediction model, with the associated location privacy risks. 
In this paper, we apply for the first time FL \cite{original_federated} to the signal maps prediction problem, which allows users to keep their data %
on their device, while still collaborating to train a global model, by exchanging model parameter updates with the server.

In the federated learning framework \cite{original_federated, fl_survey}, %
the server and the users agree on a DNN architecture and they communicate in rounds ($t=1,\dots,R$) to train it.  In every round $t$, the server initializes the global parameters $w_t$ and asks for a local update from  %
a subset (fraction $C$) of the users. The user $k$ trains a local model on its local data and sends its update for the local model parameters $w_t^k$ to the server. The server averages all  received updates, updates the global parameters to $w_{t+1}$ and initiates another round; until convergence. If a single gradient descent step is performed on the local data, the scheme is called Federated SGD (FedSGD). If there are multiple local steps, (\ie  local data are partitioned into mini-batches of size $B$ each, there is one gradient descent step per mini-batch, and  multiple passes $E$ epochs) on the local data), the scheme is called Federated Averaging (FedAvg) \cite{original_federated}. FedSGD is  FedAvg  for $E=1, B=\infty, C=1$.  $B=\infty$ indicates that the entire local batch is treated as one mini-batch.

{\bf Online Federated Learning.} Differently from the classic FL setting \cite{original_federated}, the local data of user $k$ are not available all at once, but arrive in an online fashion as the user collects measurements. We consider that the interval $T$ (for processing online data)  coincides with one round $t$ of federated learning, at the end of which, the user processes the local data  $D_t^k$ that arrived during the last time interval $T$; it then updates its local model parameters $w_t$ and sends the update to the server. \blue{We introduce a new local pre-processing step in line 16 in Algorithm \ref{alg:FedAvg}: the user may choose to use all recent local data $D_t^k$ or a subset of it as its \LocalBatch.}  (Unless explicitly  noted, we mean \LocalBatch= $D_t^k$, except for Section \ref{sec:manipulation} where \algoname is introduced to pick \LocalBatch $\subset D_t^k$ so as to increase location variance and privacy.) Once \LocalBatch is selected, FedAvg can  further partition it into a set of mini-batches ($B_t^k$) of size $B$.  An example is depicted in Fig. \ref{fig:example_local_data_per_round}, where data are collected and processed by user $k$ in rounds of $T=$ one day.

How to update the model parameters based on the stream of local data  ($\{D_t^k\}, t=1,\dots,R$) is the subject of active research area on online 
learning \cite{bottou2004large, li2014efficient}. 
We adopt the following approach. In every round $t$, user $k$ uses only its latest batch $D_t^k$ for local training and for computing $w_t^k$. The data collected in previous rounds ($D_1^k,...D_{t-1}^k$) have already been used to compute the previous local ($w_{t-1}^k$) and global ($w_{t-1}$) model parameters but is not used for the new local update. This is one of the state-of-the-art approaches in federated learning  \cite{chen2020asynchronous, liu2020fedvision, damaskinos2020fleet}. 
\blue{Our design choice to discard data from previous rounds, raises a concern about catastrophic forgetting \cite{french1999catastrophic,kemker2018measuring}.  Our intuition is that this will {\em not} happen in our datasets because of the predictable and repeated patterns in human mobility data. As users visit the same locations and follow the same trajectories over days and weeks, they contribute similar data over time. This intuition was, indeed, confirmed by the model evaluation.\footnote{\blue{Due to lack of space, quantitative comparison to alternative approaches (\ie ``Cumulative Online FL'', which accumulates all training data as they arrive, and ``Testing on Past Data'', which trains on the current round but tests on the past data) are deferred to Appendix \ref{appendix:alldata}, under supplementary materials.
The consistency in performance across all evaluation scenarios confirms that there is no catastrophic forgetting.}} Therefore, the design choice of discarding past data allows us to train a good signal strength model, while keeping storage and computation light.}

\begin{algorithm}[t]
	\small
    \DontPrintSemicolon
    
    \SetAlgoNoEnd
	\SetAlgoNoLine
	
	Given: $K$ users (indexed by $k$); $B$ local mini-batch size; $E$ number of local epochs; $R$ number of rounds of duration $T$ each; $C$ fraction of clients; 
	$n_t$ is the total data size from all users at round t, $\eta$ is the learning rate; the server aims to reconstruct the local data of target user $k$.\\
	{\bf Server executes:}\\
	\SP Initialize $w_0$\\
	\SP \For{each round t = 1,2, ... $R$} {
		\SP 	$m \leftarrow max(C\cdot K,1)$\\
		\SP 	$S_t \leftarrow$ (random subset ($C$)  of users)\\
		\SP 	\For{each user $k \in S_t$ in parallel} {
			\SP 		$w_{t}^k \leftarrow UserUpdate(k, w_{t-1}, t, B)$\\
			\SP \If{$k$=={\it target}}{
			\SP $\nabla{w_{t}^{target}} \leftarrow w_{t}^k - w_{t-1}^k$ \\
			\SP $\mathbf{DLG}$($F(x; w_{t}^k)$, $w_{t}^k$, $\nabla{w_{t}^{target}}$)} %
			} %
		\SP $w_{t} \leftarrow \sum_{k=1}^K \frac{n_t^k}{n_t}w_{t}^k$\\
		\SP 
		}
	{\bf UserUpdate($k,w,t, B$):}\\
	\SP  \blue{Local data $D_t^k$ are collected by user $k$ during round $t$}\\
	\SP  \blue{Select LocalBatch $\subseteq D_t^k$ to use for training}\\
    \SP $n_t^k=|\text{LocalBatch}|$: training data size of user $k$ at round $t$\\
	\SP $B_t^k \leftarrow$ (split LocalBatch into mini-batches of size $B$)\\
	\SP  \For{each local epoch i: 1...$E$}{%
		\SP 	\For{mini-batch $b\in B_t^k$} {
			\SP $w \leftarrow w - \eta\nabla\ell(w; b)$\\
		}
	}
	\SP  \Return{$w$ to server}
	
\caption{Online FedAvg with DLG Attack.}
\label{alg:FedAvg}	
\end{algorithm}

\begin{algorithm}[t]
	\small
	\SetAlgoNoEnd
	\SetAlgoNoLine
	\DontPrintSemicolon
	
	{\bf Input:} $F(x; w_t)$: DNN model at round $t$; $w_t$: model weights, $\nabla{w_t}$:model gradients, after target trains on a data batch of size B at round t,  learning rate $\eta$ for DLG optimizer; $m$: max DLG iterations; $a$: regularization term for cosine DLG loss.\\
	{\bf Output:} reconstructed training data (x, y) at round t \\
	\SP Initialize $x'_0 \leftarrow \mathcal{N}$(0,1), $y'_0 \leftarrow $ $\bar{y}$ \tcp*{mean RSRP}
	\SP \For{i$\leftarrow$ 0,1, ... $m$} {
		\SP 	$\nabla{w'_i} \leftarrow \partial{\ell({F(x'_i, w_t), y'_i)}}$/$\partial{w_t}$\\
		\SP 	$\mathbb{D}_i \leftarrow 1 - \frac{\nabla{w}\cdot\nabla{w'_i}}{\parallel\nabla{w}\parallel\parallel\nabla{w'_i}\parallel} + \alpha$   \tcp*{cosine  loss}
	    \SP 	$x'_{i+1} \leftarrow x'_i - \eta\nabla_{x'_i}\mathbb{D}_i$, $y'_{i+1} \leftarrow y'_i - \eta\nabla_{y'_i}\mathbb{D}_i$
      }
	\SP \Return{$x_{DLG} \leftarrow x'_{m+1}$}
	\caption{\blue{DLG Attack.}}
\label{alg:DLG}
\end{algorithm}

{\bf Honest-but-Curious Server.} We assume an honest-but-curious server who receives and stores model updates  %
from each user, and whose goal is to infer the user's locations. The server may be interested in various  %
location inference goals: \eg the user trajectory at various spatiotemporal granularities, important locations (\eg home or work), presence near points-of-interest (\eg has the user been at the doctor's office?). %
W.l.o.g., %
the server targets a user $k$ who participates in round $t$: it compares updates across successive rounds $w_{t-1}^k$ and $w_t^k$ and  computes the gradient for round $t$; see Algorithm \ref{alg:FedAvg} and  Fig.\ref{fig:example_local_data_per_round}. It then uses this gradient to infer user $k$'s location in round $t$, as described next. %

\begin{figure}[t!]
		\centering
    \includegraphics[width=0.65\linewidth]{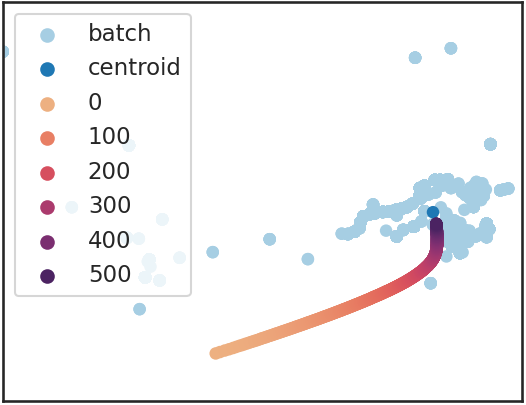}
\caption{\small {\blue{\bf Example of DLG attack (Algorithm \ref{alg:DLG}), using a user from the \campus.} The target has been in the light blue locations %
during an interval of $T$=1 week. The true average location (centroid, $\bar{x}$) of points %
is shown in dark blue. The DLG attack starts (iteration "0") with a dummy %
location (shown in yellow) and it gets closer to the centroid with more iterations (darker color indicates progression in iterations), by minimizing the cosine loss $\mathbb{D}$ of the true gradient and the gradient of the reconstructed point. After $m=500$ iterations, the distance between the reconstructed location $x_{DLG}$ (dark purple) and $\bar{x}$ %
(dark blue) is only 20 meters.}}
\vspace{-10pt}

\label{fig:example_DLG_attack_1_week_batch}
\end{figure}

\textbf{DLG Attack against a Single Update.} %
At Line 11 of Algorithm \ref{alg:FedAvg}, the server launches a DLG attack to infer the location of user $k$ in round $t$. The DLG attack is defined in Algorithm \ref{alg:DLG} and an example is shown in Fig. \ref{fig:example_DLG_attack_1_week_batch}. In each iteration $i$, the DLG algorithm: 1) randomly initializes a dummy location $x_0'$ (shown in yellow), 2) obtains the gradient at dummy location, $\nabla{w'_i}$, \aka dummy gradient, 3) updates the dummy location towards the direction that minimizes the cosine distance between the dummy and true gradient. We choose to minimize cosine, as opposed to euclidean loss %
to match the direction, not the magnitude, of the true gradient \cite{geiping2020inverting}. 

{\em Implementation details.} 
 (1)  The attacker reconstructs both the location $x$ (\ie latitude and longitude coordinates), and the RSRP value $y$; we cannot use the analytical reconstruction of the label proposed in \cite{idlg}, since we have regression instead of classification. %
 (2) We observe that different location initializations converge to the same point in practice, as shown in Fig. \ref{fig:fed_sgd_initializations_one_round}. We initialize the prediction label with the mean RSRP from the training data, which is  realistic: the attacker can have access to public cellular signal strength measurements or collect a few  measurements around each cell tower. \blue{(See Appendix \ref{sec:additional_results} on ``Analysis of DLG Label Initialization'', for a discussion on different RSRP initializations)}
 (3) We set the maximum number of iterations to $m=400,000$, and add an early stopping condition: if the reconstructed location does not change for 10 DLG iterations, then we declare convergence.

\blue{{\bf Key observation O1: DLG on one batch.}}
An example of our DLG attack on one SGD update from user $k$  is depicted in Fig. \ref{fig:example_DLG_attack_1_week_batch}: it reconstructs $x_{DLG}$, which ends up being %
close to the average location $\bar{x}$ in  batch $D_t^k$. We experimented with multiple %
initializations and we found that to be true in practice, independently of initialization; see Fig. \ref{fig:fed_sgd_initializations_one_round}. Therefore,  when applied to a single local update, {\em   Algorithm \ref{alg:DLG} reconstructs one location $x_{DLG}$, which is close to the average location $\bar{x}$ of the $N$ points in that batch.} This is in contrast to the original DLG attacks on images %
\cite{dlg, geiping2020inverting}, which aimed at reconstructing all $N$ points in the batch. 

Since local data arrive in an online fashion, %
the server can launch one DLG attack per round and reconstruct one (the average) location in each round. All  reconstructed locations together reveal the user's whereabouts, as discussed next. 

\blue{{\bf Key observation O2: DLG across several rounds.}} Human mobility is well-known to exhibit strong spatiotemporal patterns, \eg (i) people move in continuous trajectories and (ii)  they frequently visit a few important locations, such as home, work, gym, etc \cite{ isaacman2011identifying, becker2013human}. Because of (i), inferring the average location in successive rounds essentially reveals the trajectory at the granularity of interval $T$. Because of (ii), there is inherent clustering around these important locations, as exemplified in Fig. \ref{fig:1h_strongest_attack_points}. These can reveal sensitive information \cite{nytimesarticle} and help identify the user \cite{de2013unique}.

\subsection{Analytical Insights \label{sec:analysis}}

In this section, we are interested in analyzing how close is the location reconstructed by the DLG attack ($x_{DLG}$) to the true average location of the batch ($\bar{x}$).
 The analysis explains analytically our empirical observations, %
  provides insights into the performance of DLG attacks depending on the input data characteristics, and guides our design choice to maximize privacy.

\begin{restatable}{lemma}{dlglemma}
Given a data mini-batch of size $B$: $\{(x_i,y_i)\}_{i=1}^{i=B}$, the DLG attacker can reconstruct a unique user location $x_{DLG}$, which satisfies:
\begin{equation}
    x_{DLG}=\frac{1}{B}\sum_{i=1}^{i=B}\frac{g_i(w)}{\bar g(w)}\cdot x_i,
\end{equation}
where $g_i(w)=\frac{\partial \ell(F(x_i,w),y_i)}{\partial b^1_h}\in\mathbb{R}$ is an element in the gradient $\nabla w$ representing the partial derivative of loss $\ell$ w.r.t $b^1_h$, and $\bar g(w)=\frac{1}{B}\sum_{i=1}^{i=B}g_i(w)$.
\label{lemma1}
\end{restatable}

\noindent{\bf Proof:} We are able to prove this lemma under two assumptions on the DNN model architecture: (1) the DNN model starts with a biased fully-connected layer (Assumption \ref{assumption1} in Appendix \ref{proof-lemma1});  and (2) the bias vector of its first layer ($b^1_h$) has not converged (Assumption \ref{assumption2} in Appendix \ref{proof-lemma1}). %
$\square$

Next, we bound the distance between the reconstructed user location by the DLG attacker and the centroid of user locations in a data mini-batch as follows:
\begin{restatable}{theorem}{oldtheorem}
Suppose that a data mini-batch of size $B$: $\{(x_i,y_i)\}_{i=1}^{i=B}$ is used to update the DNN model $y_i=F(x_i, w)$ during a gradient descent step. Then, the reconstruction error of the DLG attacker, defined as the $L_2$ distance between the user location reconstructed by DLG attacker $x_{DLG}$ and the centroid of user locations in this mini-batch $\bar{x}=\frac{1}{B}\sum_{i=1}^{i=B}x_{i}$, can be bounded by the following expression: 
\begin{equation}
\begin{split}
    &||x_{DLG}-\bar{x}||_2 = \frac{1}{B|\bar{g}(w)|}||\sum_{i=1}^{i=B}\big(g_i(w)-\bar{g}(w)\big)\cdot\big(x_{i}-\bar{x}\big)||_2 \\
    &\leq \frac{1}{2B|\bar{g}(w)|}\sum_{i=1}^{i=B}\Big(\big(g_i(w)-\bar{g}(w)\big)^2+||x_{i}-\bar{x}||_2^2\Big).
\end{split}
\vspace{-10pt}
\end{equation}
\label{theorem1}
\end{restatable}
\vspace{-10pt}
\noindent{\bf Proof:} See Appendix \ref{proof-theorem1}. $\square$

Theorem \ref{theorem1} says that the reconstruction error of the DLG attacker is equal to the $L_2$ norm of the sample co-variance matrix between the partial derivative over the bias $g_i(w)$ and the user location $x_i$, divided by the (absolute value of the) average partial derivative within a mini-batch $\bar g(w)$. Moreover, this error can be upper bounded by the sum of the sample variance of $g_i(w)$ and the sample variance of $||x_i||_2$ in the mini-batch. \blue{This is a tight bound that can be achieved if, for any $i$ and $j$, it is: $|x_i - \bar x| = |x_j - \bar x| =  |g_i(w) - \bar g(w)| =  |g_j(w) - \bar g(w)|$}.

The above theorem involves the partial derivatives $g_i(w)$ whose values are hard to predict when $x_i$ varies. To bound the error of the DLG attacker without involving such derivatives, we further make the mild assumption that the gradient function is Lipschitz continuous (see Assumption \ref{assumption3} in Appendix \ref{proof-theorem2}), and state the following theorem:

\begin{restatable}{theorem}{newtheorem}
Subject to the Lipschitz continuity assumption about $\nabla\ell(F(x_i,w),y_i)$, the reconstruction error of the DLG attacker can be bounded by:
\begin{equation}
\begin{split}
    ||x_{DLG}-\bar x||_2 \leq \frac{L^2}{B|\bar{g}(w)|}\sum_{i=1}^{i=B}\big(\alpha||x_{i}-\bar x||^2+||y_{i}-\bar y||^2\big),
\end{split}
\label{theorem1-eq3}
\end{equation}
where $\bar{y}=\frac{1}{B}\sum_{i=1}^{i=B}y_{i}$ and $\alpha=1+\frac{1}{2L^2}$.
\label{theorem2}
\end{restatable}
\noindent{\bf Proof:} See Appendix \ref{proof-theorem2}. $\square$

Theorem \ref{theorem2} says that %
the reconstruction error can be bounded by the weighted sum of the sample variance of the user data $||x_i||_2$ and the labels $y_i$ in the mini-batch. \blue{To achieve the bound of Eq. (\ref{theorem2}), Assumption \ref{assumption3} should hold in addition to the condition for achieving the bound of Eq. (\ref{theorem1}). For instance, when the mini-batch size $B$ is equal to 1, the equality in Eq. (\ref{theorem1}) and Eq. (\ref{theorem2}) can be achieved.}

{\bf (I1) Impact of data mini-batch variance.} Theorem \ref{theorem1} shows that the variance of user locations affects the upper bound of the DLG attacker's reconstruction error. 
The smaller the data mini-batch variance is, the smaller the upper bound of the DLG attacker's reconstruction error is. 
Theorem \ref{theorem1} also shows that the DLG error depends on the variance of the gradients. One may intuitively argue that since
the randomness of the gradients comes from the randomness of the data, 
the larger the data variance the larger the gradients variance too, thus the larger the error.
Theorem \ref{theorem2} does not involve gradients. It directly shows how the variance of local user data and associated labels affect the upper bound of the reconstruction error. 
Motivated by the above discussion, we propose an algorithm, which we refer to as \algoname to increase the mini-batch data variance of each user during training, see Sec. \ref{sec:manipulation}. 
In theory, an increasing upper bound does not guarantee that the actual reconstruction error will increase. We empirically show this to be the case (see Table \ref{tab:dbscan_stats_online_1W} in Sec. \ref{sec:manipulation}).

{\bf (I2) Impact of model convergence rate.} As shown above, another key component affecting the upper bound of the attacker's reconstruction error is $|\bar g(w)|$, which is the average partial derivative over the bias and reflects the convergence of the global model: As the global model converges (\ie the training loss $\ell$ converges to zero), $|\bar g(w)|$ will also converge to zero, and hence the upper bound of the reconstruction error  will diverge to infinity. This is expected since the attacker needs user information from the gradient to reconstruct users' location.
Recall that the attacker attempts to reconstruct one user location at each mini-batch and FL round.
As the model converges faster, the reconstruction error will diverge faster and thus a smaller fraction of reconstructed user locations will be accurate, those corresponding to early reconstruction attempts, see Fig. \ref{fig:fed_sgd_initializations_multiple_rounds}. %

{\bf (I3) Impact of Averaging.}
While under FedSGD the attacker observes the gradient updates after processing each mini-batch, under FedAvg the attacker observes the gradient update at the end of the batch/round, thus this gradient update is the time average of $\frac{\partial \ell(F(x_i,w),y_i)}{\partial b^1_h}$ during each training round. 
We can apply Theorems \ref{theorem1} and \ref{theorem2} to the FedAvg case by re-defining 
$g_i(w)$ as the time average of $\frac{\partial \ell((x_i,w),y_i)}{\partial b^1_h}$ during each training round, and the impact of each parameter will be the same, as that discussed above for FedSGD. \blue{In Sec. \ref{sec:local-leakage} and \ref{sec:averaging}, we show the impact of FedAvg parameters ($B$, $E$) on the DLG attack: as $B$ decreases and/or $E$ increases, the attack is less accurate, due to faster model convergence rate caused by multiple local gradient descent steps (see  Fig. \ref{fig:minibatch_comparison} and  Fig. \ref{fig:sgd_local_epochs_comparison}).}

{\bf (I4) Impact of multiple users.} FL involves the participation of multiple users, who will jointly affect the convergence of the global model. Prior work has shown that as the data diversity across multiple users increases, \ie the dissimilarity or heterogeneity between users increases, the global model may converge slower or even diverge \cite{li2020federated, haddadpour2019convergence}. 
The global model convergence rate impacts the DLG attacker's reconstruction accuracy. Thus, when the data diversity across multiple users increases, we expect that the global model will converge slower, resulting in a more accurate reconstruction of user locations by the DLG attacker. In Sec.\ref{sec:multiusers}, we empirically show how the similarity of users affects the DLG attacker's performance; see Table \ref{tab:radiocells}.

\vspace{-5pt}
\section{Evaluation Setup\label{sec:evalsetup}}

We evaluate the success of the DLG attack for different scenarios: we specify the exact configuration and parameter tuning for the online federated learning  (Algorithm \ref{alg:FedAvg}), DLG attack (Algorithm \ref{alg:DLG}), and any defense mechanism in Section \ref{sec:results}. %
In Sec. \ref{sec:dataset}, we describe two real-world datasets that we use as input to our simulations. In Sec. \ref{sec:metrics}, we define privacy metrics that quantify the privacy loss due to the attack. %

\subsection{Datasets}\label{sec:dataset}

\textbf{\campus \cite{alimpertis2017system}.} This dataset is collected on a university campus and is the one depicted in Fig. \ref{fig:density_maps_uci}. It contains real traces from seven different devices used by student volunteers and faculty members, using two cellular providers over a period of four months. IRB review was not required as the proposed activity was deemed as non-human subjects research by the IRB of our institution. It is a relatively small dataset; it spans an entire university campus, a geographical area of approx. 3 ${km}^2$ and 25 cell towers. However, it is very dense: it consists of 169,295 measurements in total. Pseudo-ids are associated with the measurements which facilitate the simulation of user trajectories in FL.  The number of measurements per cell tower and user varies, details are deferred to Appendix \ref{sec:appendix_data}.  %
For the evaluation in Sec. \ref{sec:results}, we pick the cell tower x204 with the largest number of datapoints, and choose the user 0 with the most measurements as the target user. 
The campus is depicted in Fig. \ref{fig:all_cells_density_maps_merged}, and the locations of the target (user 0) are depicted in Fig. \ref{fig:all_cells_density_maps_user0} (all cells) and Fig. \ref{fig:1h_strongest_attack_points}  (cell x204 only). It is worth noting that we know the frequently visited locations coincide with home (%
on the right part of the picture) and work (campus buildings on the left part of the picture) locations for the user. This side information becomes useful when we assess the success of the attack. Moreover, the attacker uses the campus boundaries (an area of 3 ${km}^2$) as the defined area of the attack; if some reconstructed locations fall outside this area, they are treated as diverged.

\textbf{\radiocells \cite{radiocells}.} %
We also consider the large-scale real-world \radiocells, which comes from a community project founded in 2009 under the name \url{openmap.org}. %
It contains data from over 700 thousand cell towers around the world and is publicly available in \cite{radiocells}. %
Raw measurement data are available for in-depth analysis of communication networks or cell coverage, and can be used for our problem of signal maps. %
The measurements are contributed by real users without logging their user ids. Users log and upload their data into multiple upload files: each containing device information, version of the app, etc. that can be used for distinguishing users, as in \cite{carmela}. %
We focus on cellular data from 2017 (8.5 months) in  the area of London, UK, from the top cell tower (x455), which had the most measurements (64,302 in total) from approx. 3,500 upload files and a geographical area that spans approx. 5,167 $km^2$. Each upload file corresponds to a single device, typically containing 16 measurements per $2h$ on average. %
Since no pseudo-ids are provided that would allow us to link multiple upload files of a user, we use heuristics to create longer user trajectories so that we have enough data points per batch; see details in Sec. \ref{sec:multiusers}. The trajectories of the synthetic users are depicted in Fig. \ref{fig:radio_cell_x455_user_map_all_users}. %

{\bf For both datasets,} we partition the data into batches $D_t^k$ for each $k$ user, corresponding to different time intervals $T$ in time-increasing order ($t=1,..,R$), so as to simulate the online collection of data points as it happens. %
We mainly focus on $T$=1 week; each batch contains all datapoints in that week, which are used in one FL round. Choosing coarser $T$ results in fewer batches/rounds but it includes more datapoints per batch, which facilitates local training. In the \campus, there are 11 weeks for user 0 (the target user in the \campus) and the average batch size is 3,492 measurements. In \radiocells,  most users' batches contain fewer than 50 datapoints on average, as it is a much sparser dataset. The target user in \radiocells (user 3) contains 26 1-week  batches.  The features in each dataset are standardized by subtracting the mean and scaling to unit variance so that their values span an appropriate range. %

\subsection{Location Privacy Metrics \label{sec:metrics}}

\blue{We use the RMSE for the signal maps prediction problem as our utility metric, as described at the beginning of Sec. \ref{sec:setup}.} We also need metrics that capture how similar or dissimilar the reconstructed locations are from the real ones. %
\blue{Any FL algorithm and any defense mechanism must be evaluated w.r.t. the privacy-utility trade-off they achieve.}

\begin{figure}[t!]
		\centering
    \begin{subfigure}{0.85\linewidth}
		\centering
        \includegraphics[width=0.99\linewidth]{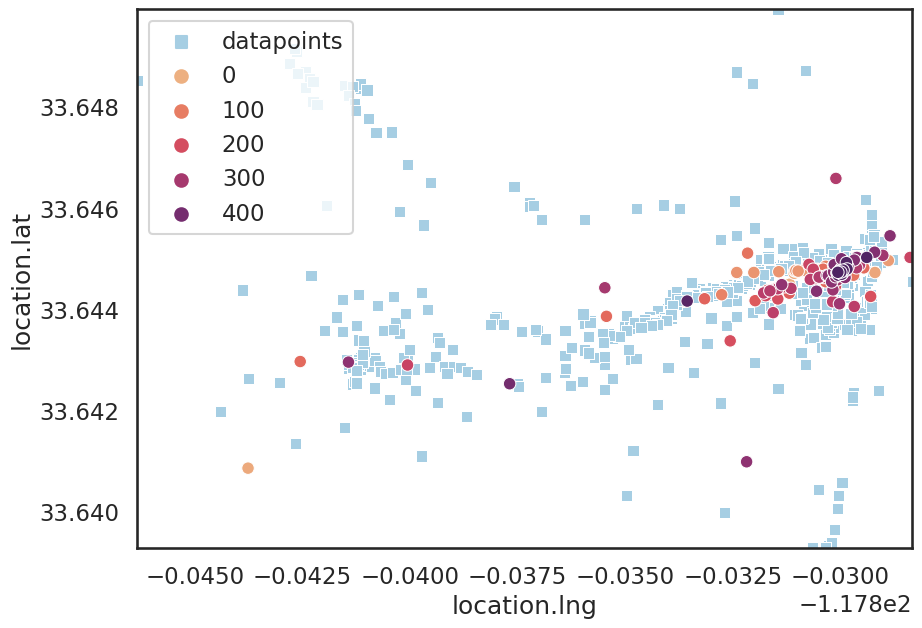}
		\caption{Reconstructed locations by the DLG attack, considering one attack per 1-hour batches and FedSGD. The distance between the real and inferred location distributions is small: EMD=5.3.}
		\label{fig:1h_strongest_attack_points}
    \end{subfigure}
    \begin{subfigure}{0.85\linewidth}
		\centering
    \includegraphics[width=0.99\linewidth]{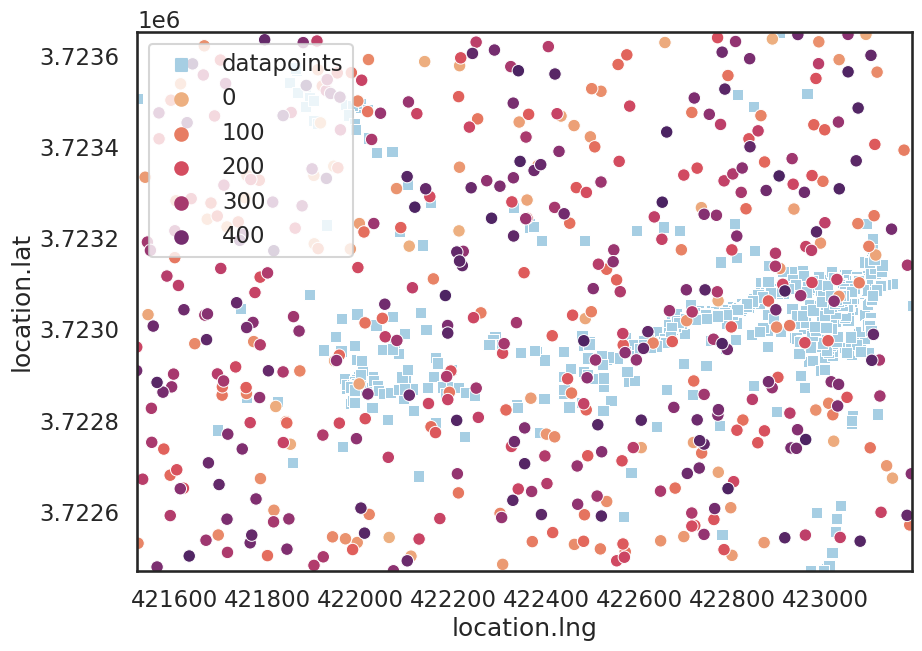}
		\caption{Locations generated uniformly at random (1 per hour) within the defined campus boundaries. The distance between the real and uniform random location distributions is high: EMD = 21.33 for 5 realizations; it provides a baseline for comparison.} %
		\label{fig:random_EMD_1h_batches_strongest}
	\end{subfigure}
\vspace{-5pt}
\caption{ \small We consider a target user and its real locations on campus, 489 in total (with $T=1$ hour), depicted in light blue. The over-sampled  area on the right corresponds to home location of that user. The other area on the left, corresponds to their work on campus. The DLG attacker processing updates from 1h rounds can successfully reconstruct the important locations of the user: the difference between the distribution of real and the inferred locations is EMD=5.2. To put that in context, if one would randomly guess the same number of locations, the EMD would be 21.33. }
\vspace{-5pt}
\label{fig:EMD_random_examples}
\end{figure}

\textbf{Visualization.} Visually comparing the reconstructed  ($x_{DLG}$) to the true locations in the batch ($D_t^k$) provides intuition. %
Fig. \ref{fig:EMD_random_examples}(a) shows the real (shown in light blue) and the reconstructed (shown in color) locations reconstructed  per 1h-long batches, for a user in the \campus. One can see that the reconstructed locations match the frequently visited locations of the user. For example, DLG seems indeed to reconstruct locations on the right side of the figure,  where we confirmed that graduate student housing is on this campus. %
 This is expected, based on our key observation O2, \ie the characteristics of human mobility. %

\textbf{Distance from the Centroid.} In order to assess how accurate the attack is within a single round $t$, we use the distance $||x_{DLG}-\bar{x_t}||$, \ie how far (in meters) is the reconstructed location  from the average location of points in that batch $B_t^k$. Based on the key observation O1, we expect this distance to be small when the attack converges. %

\textbf{Comparing location distributions.} To assess the success of the attack considering all rounds $t=1,...,R$, we need a metric that captures how similar or dissimilar are the reconstructed locations from the real ones; \eg see Fig. \ref{fig:EMD_random_examples} for an example of real (light blue) vs. reconstructed (in color) locations over multiple rounds.  We considered 
the KL-divergence and Jensen-Shannon distance, which are well-known metrics for comparing distributions. However, they only capture the difference in probability mass, not the spatial distance between real and inferred frequent locations, which is of interest in our case, as per key observation O2. %

We use the {\bf Earth Movers Distance (EMD)} \cite{bonneel2015sliced, flamary2021pot} to capture the distance between the 2D-distributions of reconstructed and real locations. It has been previously used in location privacy as a measure of t-closeness \cite{4221659}, l-diversity \cite{8367709}. %
It is defined as: $EMD(P,Q) = \inf_{\gamma \in \Pi(P,Q)}{\mathbb{E}_{(x,y)\sim\gamma}[\|x-y\|]}$, where $\Pi(P,Q)$ is the set of all joint distributions whose marginals are $P$ (true locations) and $Q$ (reconstructed locations). 
EMD takes into account the spatial correlations and returns the minimum cost required to convert one probability distribution to another by moving probability mass.\footnote{A classic interpretation of EMD is to view the two %
probability distributions as two ways to pile up an amount of dirt (``earth'') over a region and EMD as the minimum cost required to turn one pile into the other. Cost is defined as the amount of dirt moved x the  distance it was moved.} We use the Euclidean distance when calculating EMD on GPS coordinates in UTM. We compute EMD using Monte Carlo approximations for $N=1000$ projections;  which is more computationally efficient than the exact EMD calculation and it is suitable for 2D distributions. %

The range of EMD values depends on the dataset and spatial area. $EMD=0$ would mean that the distributions of real and reconstructed locations are identical. Low EMD values indicate successful location reconstruction by the DLG attack thus high privacy loss. 
To get more intuition, let's revisit Fig. \ref{fig:EMD_random_examples}. In Fig. \ref{fig:EMD_random_examples}(a), the DLG attacker reconstructed locations around the  frequently visited locations (close to home) and achieved $EMD=5.3$. In Fig. \ref{fig:EMD_random_examples}(b), we show the same number of locations chosen uniformly at random, which leads to $EMD=21.33$; this provides an upper bound in privacy (random guesses by the attackers) in this scenario. %

\textbf{\%Attack Divergence.} In our simulations, we observed that: (i) if the DLG attack converges, it converges to $\bar{x}$ regardless of the initialization;  (ii) however, the DLG attack did not always converge,
depending on the location variance of the batch, %
 the tuning of parameters of the DLG optimizer and FedAvg. Examples of attack divergence are shown in Fig. \ref{fig:fedsgd_multiple_rounds}. In Sec. \ref{sec:results}, %
 we define a rectangular geographical area of interest for the attacker  \eg the entire 3 km$^2$ campus 
 in \campus. %
 If some reconstructed locations are outside the boundaries, we declare them ``diverged'' and (i) we discard them when computing the privacy (EMD) metric, but also  (ii) we report the fraction of those  attacks that diverged. In practice, if an attack diverges outside the area of interest, the attacker can relaunch the DLG attack with a different initialization hoping until it reaches convergence to $\bar{x}$. This, however, is costly for the attacker, therefore the \% of attacks that diverged is another metric of success or failure of the DLG attack.

\section{Evaluation Results}\label{sec:results}

\begin{figure}[t!]
	\centering
	 \begin{subfigure}{0.9\linewidth}
	\centering
    \includegraphics[width=0.99\linewidth]{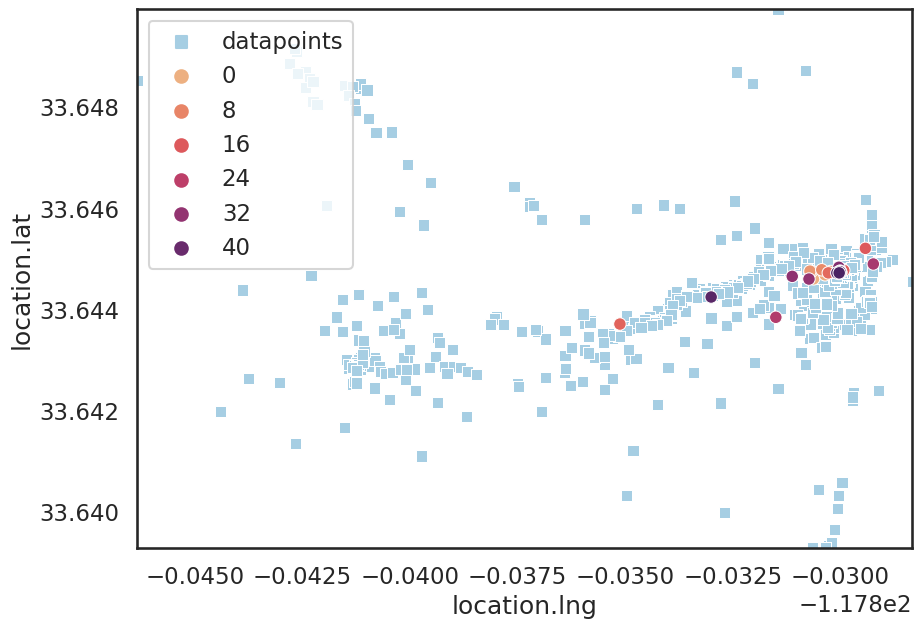}
		\caption{$T=$ 24-hour rounds: EMD=6.4.} %
		\label{fig:24h_strongest_attack_DLG_iters}
    \end{subfigure}
	
	\begin{subfigure}{0.9\linewidth}
		\centering
    \includegraphics[width=0.99\linewidth]{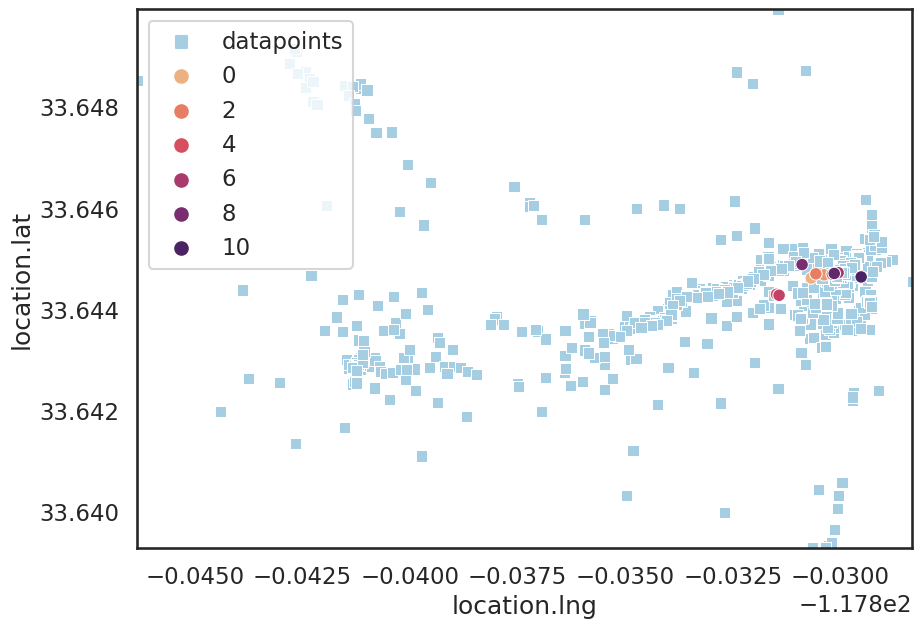}
		\caption{$T=$ 1-week rounds: EMD=7.6.} %
		\label{fig:1w_strongest_attack_points}
    \end{subfigure}
   
\caption{\blue{\small The reconstructed locations in the case of strongest attack %
for various  $T=$ 24-h vs. 1-week.} The light blue square points are the ground truth points and the circle points are the reconstructed points for each round; the darker color represents the later rounds. RMSE is 4.93, 4.91, 5.16 for 1w, 24h, 1h respectively. Reconstruction with 1-hour rounds (Fig.\ref{fig:1h_strongest_attack_points}) reveals fine-grained user trajectories. The coarser rounds (24-h, 1-week) still reveal the frequent locations of the target, \eg their home/work locations.}
\vspace{-5pt}
\label{fig:dlg_iters_cell204_strongest_attack_points}
\end{figure}

Next, we evaluate the DLG attack  in a range of scenarios. %
In Section \ref{sec:local-leakage}, we consider FedSGD, which is the most favorable scenario for  DLG -- the strongest attack. In Section \ref{sec:averaging}, we show that the averaging inherent in FedAvg provides a moderate level of protection against DLG, which also improves utility. In Section \ref{sec:manipulation}, we propose a simple defense that users can apply locally: \algoname curates local batches with high variance. In Section \ref{sec:multiusers}, we show the effect of multiple users on the success of the DLG attack.
We use a default $\eta=0.001$ (or $\eta=10^{-5}$ in case of mini-batches), details on tuning  $\eta$ are deferred to Appendix \ref{sec:additional_results}. %
W.l.o.g., we focus on a particular target, whose locations are to be reconstructed  by the DLG attacker and we report the privacy (EMD, \% diverged attacks)-utility tradeoff (RMSE).

\subsection{Location Leakage in FedSGD}
\label{sec:local-leakage}

\textbf{Strongest Attack.} FedSGD  is a special case of Algorithm \ref{alg:FedAvg} with  $B=\infty$, $E=1$: in each round $t$ the target user performs a single SGD step on their data $D_t^k$ and sends the local model parameters $w_t^k$ to server, which in turn computes the gradient (line 10 in Algorithm \ref{alg:FedAvg}). $\nabla{w_{t}^{k}}$  corresponds to the true gradient obtained on data $D_t^k$ and it is the best scenario for the attacker.

\begin{figure}[t!]
	\centering
		\centering
    \includegraphics[width=0.5\linewidth]{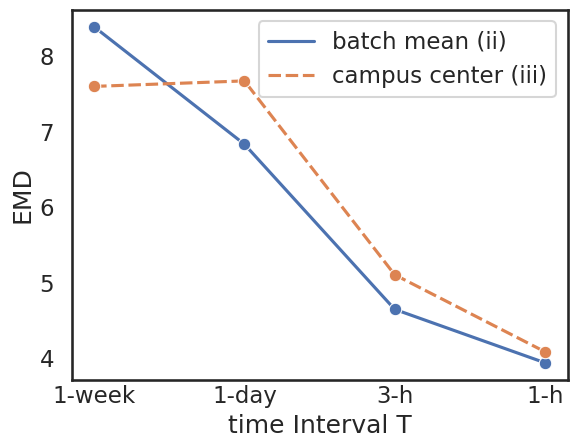}
\vspace{-5pt}
\caption{\blue{\small \textbf{Effect of Interval Duration:} the shorter the interval $T$, the less privacy (low EMD), due to lower batch variance. \textbf{Effect of Initialization Strategies:}  (ii) mean-init of the batch with Gaussian noise, (iii) initialize in the campus center, then use previous batch's $x_{DLG}$ to initialize next dummy point.  Both are strong attacks regardless of the initialization strategy: all reconstructed points converge and result in similar privacy (EMD).}}
\vspace{-5pt}
\label{fig:init_comparison_strongest_attack_1e_05}
\end{figure}

\textbf{Impact of \interval ~$T$.}  %
Fig. \ref{fig:1h_strongest_attack_points}  shows the true vs. the reconstructed locations via DLG for intervals with %
$T$=1h. Fig. \ref{fig:init_comparison_strongest_attack_1e_05} shows the %
results for the same data but divided into intervals with  duration longer than 1h, \ie 24h and 1 week.  The shorter the interval, the better the reconstruction of %
locations: we can confirm that visually and quantitatively via EMD, while the utility (RMSE) is not affected significantly. This is in agreement with Insight I1, since smaller $T$ leads to smaller batch variance in the target's trajectory.
 However, the most visited locations by the user (\ie the home and work) are successfully reconstructed for all $T$; this is due to the mobility pattern of the target, who repeats his home-work trajectory over time.

\begin{figure}[t!]
		\centering
		\centering
       \includegraphics[width=0.99\linewidth]{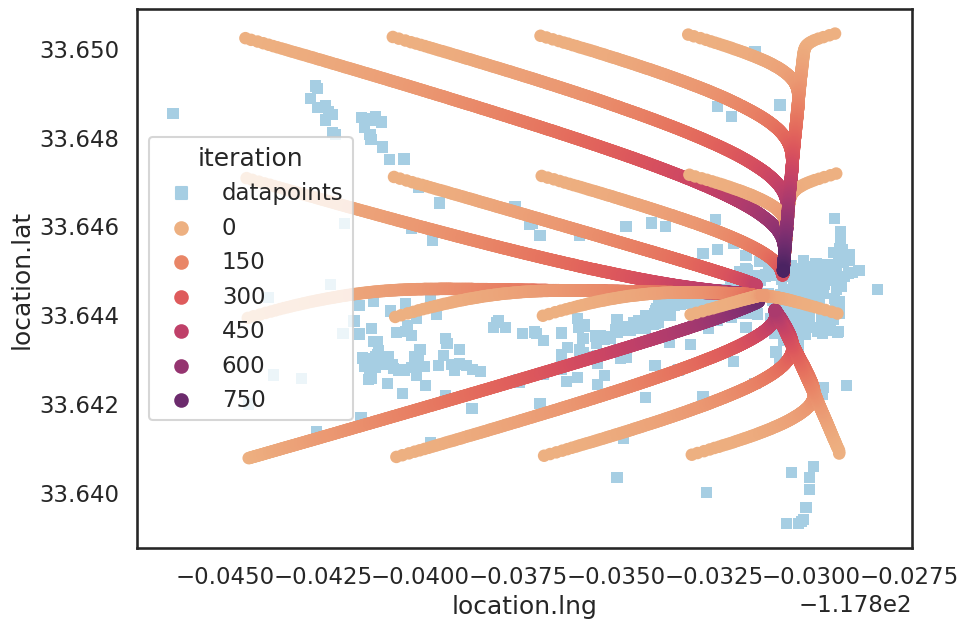}
\caption{\blue{\small \textbf{FedSGD for one Round.}} DLG converges visually and in terms of cosine loss ($\mathbb{D}<-0.9988$) to the average location $\bar{x}$ regardless of the initialization point.}%
\vspace{-5pt}
    \label{fig:fed_sgd_initializations_one_round}
	\end{figure}

\begin{figure}[t!]
		\centering
	\begin{subfigure}{0.9\linewidth}
		\centering
        \includegraphics[width=0.99\linewidth]{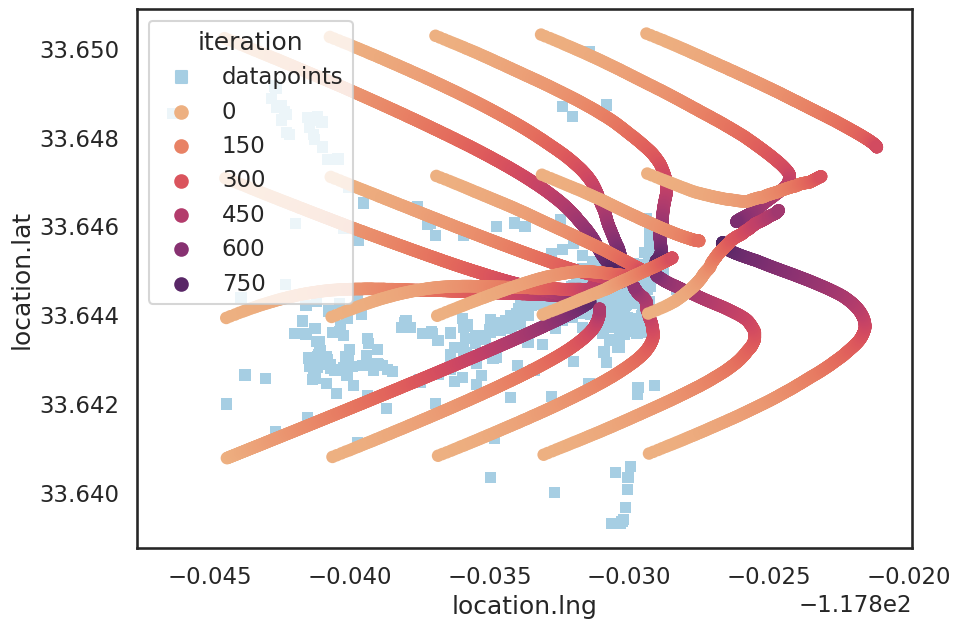}
		\caption{FedSGD over multiple FL rounds.}%
	    
		\label{fig:fedsgd_multiple_rounds}
	\end{subfigure}
	\begin{subfigure}{0.8\linewidth}
		\centering
        \includegraphics[width=0.99\linewidth]{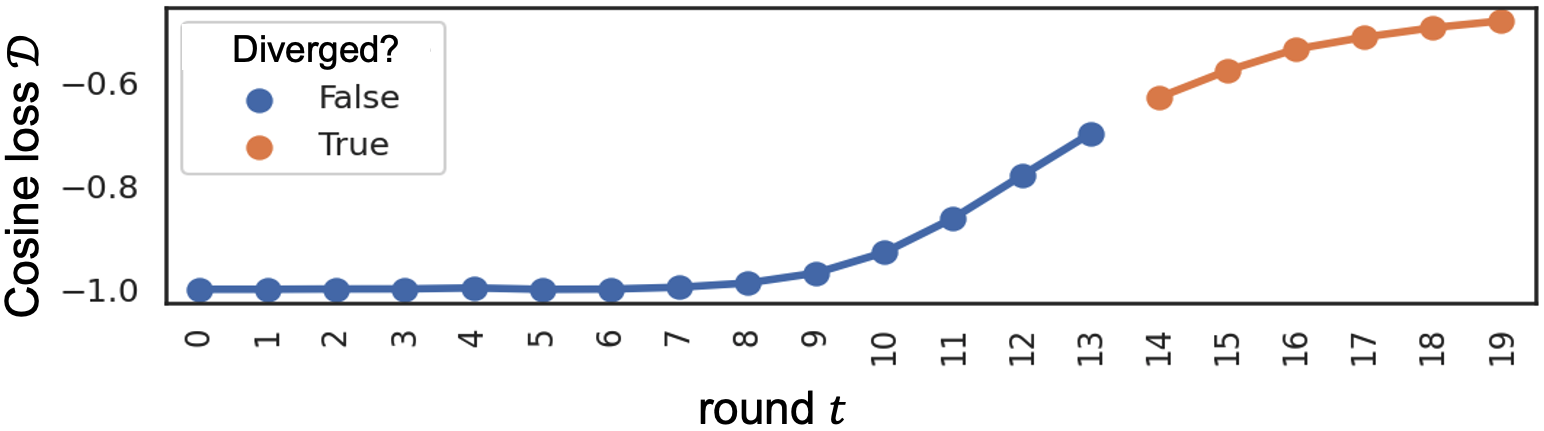}
		\caption{Cosine loss of DLG attack per round.}
		\label{fig:fedsgd_multiple_round_cosine}
	\end{subfigure}
	\begin{subfigure}{0.8\linewidth}
		\centering
        \includegraphics[width=0.99\linewidth]{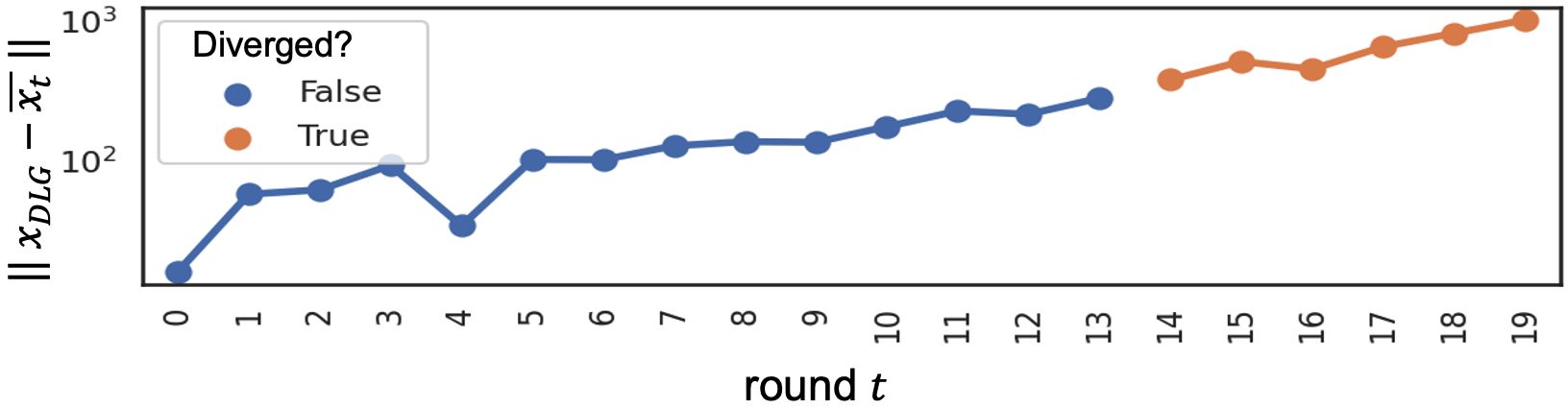}
		\caption{Distance between reconstructed $X_{DLG}$ and $\bar{x}$ in meters.}
		\label{fig:fedsgd_multiple_round_distance}
	\end{subfigure}
	\begin{subfigure}{0.8\linewidth}
		\centering
        \includegraphics[width=0.99\linewidth]{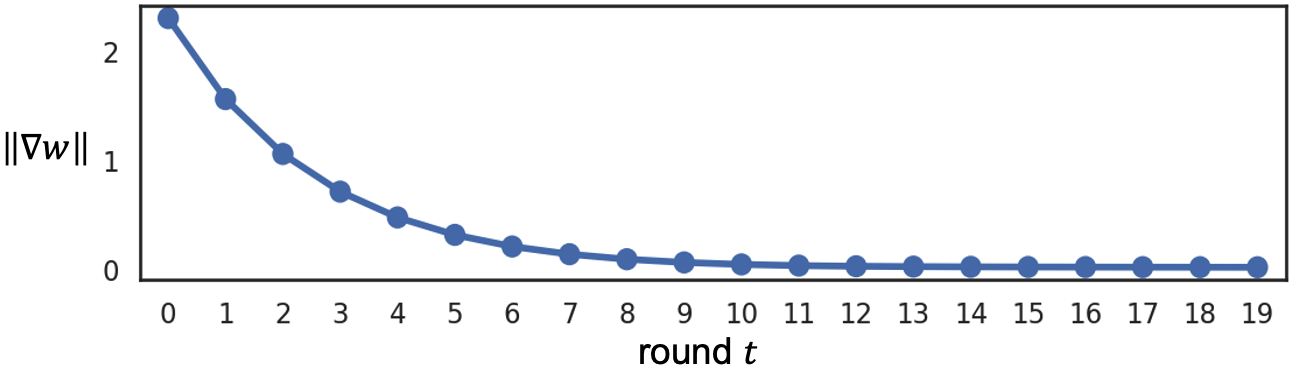}
		\caption{Norm of gradient of flattened per-layer weight matrices between two consecutive rounds.}
		\label{fig:fedsgd_multiple_round_gradient_norm}
	\end{subfigure}
\caption{\blue{\small \textbf{FedSGD over Multiple FL Rounds.}} FL is performed over multiple rounds. (a) The local data are considered the same from one round to the next, and the same as the batch used in Fig. \ref{fig:fed_sgd_initializations_one_round}. The attacker uses the same 20 random initialization as in the single round case. The effect of multiple ML rounds is that (b) the gradients are converging to zero, which results in (c) higher cosine loss for the DLG attacker and eventually (d) the reconstructed points $X_{DLG}$ end up further away from the centroid $\bar{x}$.}
\vspace{-5pt}
    \label{fig:fed_sgd_initializations_multiple_rounds}
	\end{figure}

\blue{\textbf{Impact of DLG Initialization.}} %
First, we consider a single FL round and we evaluate the effect of DLG initialization.  For example, consider LocalBatch to consist of the measurements of the target from week $t=7$ ($D_t^0$) in \campus, and perform one local SGD step to train the local model. The global model is initialized to the same random weights before local training.  The attacker splits the geographical area into a grid of 350 meters and uses the center of each grid cell as a candidate initial dummy point  in Algorithm \ref{alg:DLG}. %
Fig. \ref{fig:fed_sgd_initializations_one_round} shows the results for 20 random initialization. We observe that, for all initializations, the attack converges %
(cosine loss $<$ -0.9988) to the average location $\bar{x}$ of the local data, regardless of the distance between the initialization and the centroid of the data. %
This is explained by Insight I2 in Sec. \ref{sec:analysis}; the gradient provides information for the attack, since the model has not converged.

\textbf{Impact of Number of FL Rounds.} Second, we repeat the same experiment, but with the goal of evaluating the effect of multiple FL rounds, everything else staying the same. To that end, we consider that the LocalBatch data  is the same for all rounds $D_t^0=D_7^0, t=0,...,20$ (and the same as in the previous experiment: $D_t^0$); this removes the effect of local data changing over time in an online fashion. We also use the exact same 20 random initializations,  as above. The global model is now updated in each round and iterates with the target as in Algorithm \ref{alg:FedAvg}.  The norm of the flattened per-layer weight matrices approaches zero after round 9 (see Fig. \ref{fig:fedsgd_multiple_round_gradient_norm}) and at this point the DLG attack starts diverging; the reconstructed point is further away from the mean of the data as shown in Fig. \ref{fig:fedsgd_multiple_round_distance} and the final cosine loss $\mathbb{D}$ starts increasing (Fig. \ref{fig:fedsgd_multiple_round_cosine}). In the worst case, when the attack diverges, the reconstructed point is 1 km away from the centroid location of the batch. %
This can be explained by Insight I2 in Sec. \ref{sec:analysis}: after several rounds, the model starts converging and the gradients decrease; the average gradient goes to 0, the bound in Theorem \ref{theorem1} goes to $\infty$, and $x_{DLG}$ can go far from $\bar{x}$. Thus, even in the worse scenario of the FedSGD, without any add-on averaging or defense mechanisms, there is some protection against the DLG attack, after the initial FL rounds when the global model converges.

\textbf{DLG Initialization Strategies.}  %
There are different strategies for initializing the dummy points in each batch. (i) The attacker could pick randomly within the geographical area of interest,  as we did in Fig. \ref{fig:fed_sgd_initializations_one_round}, or the middle of the campus. (ii) The attacker could use a rough estimate of $\bar{x}$ plus Gaussian noise. (iii) The attacker could leverage the  reconstructed location from a previous round and use it to initialize the dummy point in the next round, in order to  leverage the continuity of user mobility and make an educated guess especially in the finer time intervals.
Fig. \ref{fig:init_comparison_strongest_attack_1e_05} compares  strategies (ii) and (iii): both are strong attacks, with all points converging within the area of interest, and resulting in similar EMD. If an attack diverges, (i) would be better than (ii) or (iii), to keep the dummy point within the defined boundaries. 
For the rest of the paper, we use by default strategy (ii) for faster simulation.

\subsection{Location Leakage in FedAvg} \label{sec:averaging}

FedAvg  is the general Algorithm \ref{alg:FedAvg}, of which FedSGD is a special case for  $B=\infty$, $E=1$. FedAvg \cite{original_federated} has many parameters that control the computation, communication and storage at the users and server. The learning rate $\eta$ is tuned for each dataset, see Appendix \ref{sec:additional_results}; the number of FL rounds $R$ is  discussed previously for FedSGD; the fraction $C$ of users in round is related to global averaging  in Section \ref{sec:multiusers}. Our focus here is to evaluate the effect of local averaging through the use of $B$-sized mini-batches and $E$ number of epochs, as a way to defend against DLG attacks. Intuitively, the more local SGD steps (smaller $B$ and high $E$), the more averaging over local gradients, and the less successful the DLG attack by the server based on the observed  $w_t^k - w_{t-1}^k$. Interestingly, more averaging improves both convergence and utility %
\cite{original_federated}. Throughout this section, we focus on $T=1$ week which gives 11 intervals.  %

\begin{figure}[t!]
		\centering
\begin{subfigure}{0.49\linewidth}
	\centering
    \includegraphics[width=0.90\linewidth]{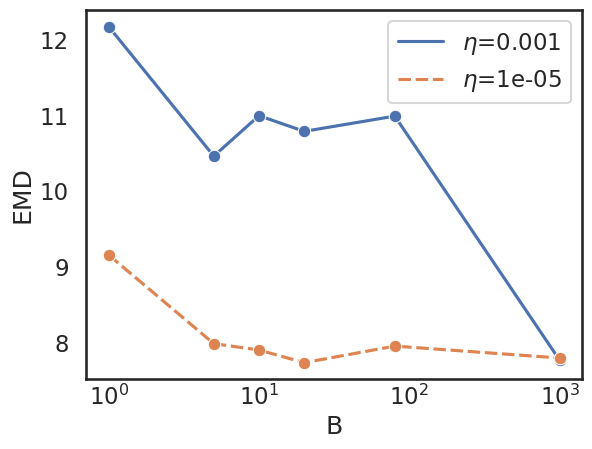}
		\caption{EMD vs. mini-batch size B.} %
	\label{fig:B_vs_EMD_comparison}
		\end{subfigure}
    \begin{subfigure}{0.49\linewidth}
		\centering
    \includegraphics[width=0.90\linewidth]{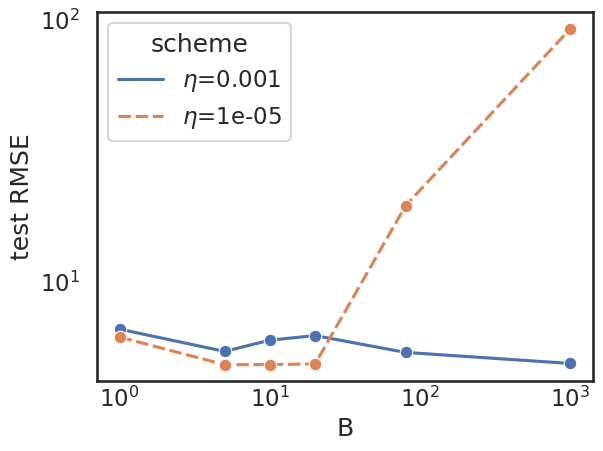}
		\caption{RMSE vs. mini-batch size B.} %
	\label{fig:B_vs_RMSE_comparison}
		\end{subfigure}
	\begin{subfigure}{0.49\linewidth}
		\centering
    \includegraphics[width=0.89\linewidth]{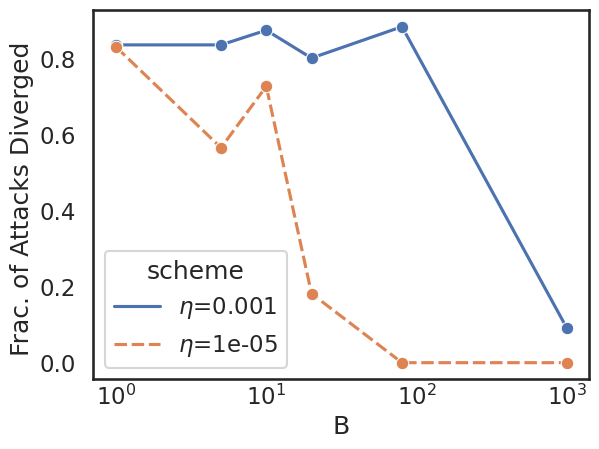}
		\caption{$B\leq20$ increases divergence.}
		\label{fig:1w_B_out_of_range}
	\end{subfigure}
	\begin{subfigure}{0.45\linewidth}
		\centering
    \includegraphics[width=0.99\linewidth]{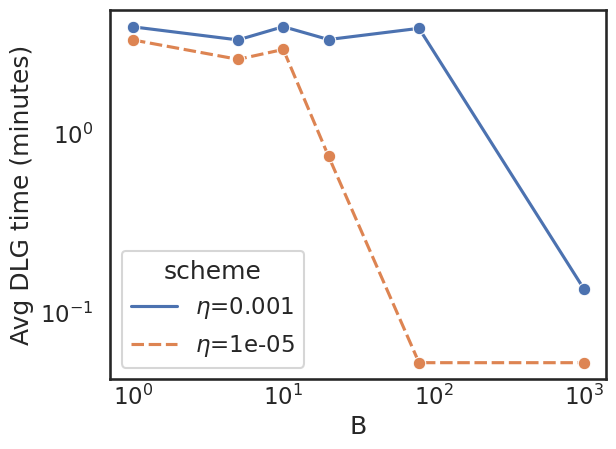}
		\caption{$B\leq20$ increases DLG time.} %
		\label{fig:B_comparison_DLG_time}
		\end{subfigure}
\vspace{-5pt}
\caption{\small \blue{\textbf{Impact of mini-batch size B in FedAvg.}
[$T$=1 week, $E=1$.] Reducing mini-batch size $B$ introduces more averaging of the gradients which increases EMD (and privacy) and makes the attack more expensive due to divergence.} The default $\eta$ seems to be less sensitive to B in terms of RMSE. Here B=1000 corresponds to $B=\infty$, thus FedSGD, %
which leads to reduced privacy but also to higher RMSE for the lower $\eta$.}
\vspace{-5pt}
    \label{fig:minibatch_comparison}
	\end{figure}

\begin{figure}[t!]
		\centering
	\begin{subfigure}{0.5\linewidth}
		\centering
    \includegraphics[width=0.86\linewidth]{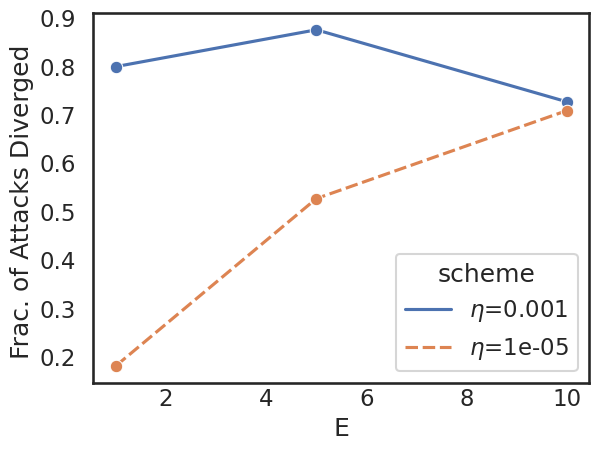}
		\caption{$E\geq5$ increases divergence.} 
		\label{fig:Epochs_priv_util_out_of_range}
	\end{subfigure}
	\begin{subfigure}{0.47\linewidth}
		\centering
    \includegraphics[width=0.99\linewidth]{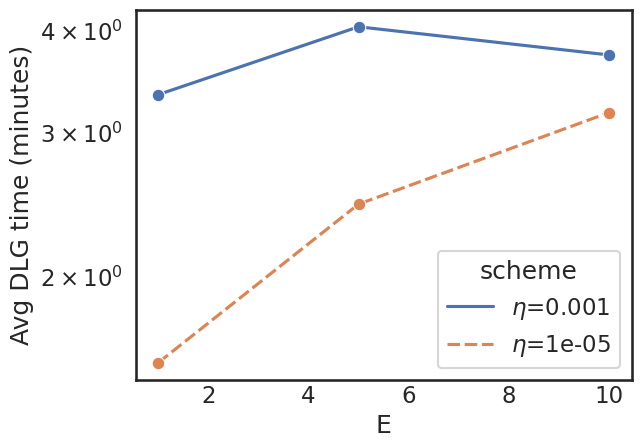}
		\caption{$E\geq5$ increases DLG time.} 
		\label{fig:Epochs_priv_util_time}
		\end{subfigure}
\caption{\small \textbf{Impact of local epochs E on 1-week rounds.} We set $B=20$ and increase local epochs, which increases the EMD and attack divergence, while utility is preserved. } %
\vspace{-5pt}
\label{fig:sgd_local_epochs_comparison}
\end{figure}

\blue{\textbf{Impact of Mini-batch Size $B$ in FedAvg.} Fig. \ref{fig:minibatch_comparison} shows the utility (RMSE) and privacy (EMD, Fraction of Attacks Diverged and Avg DLG time ) metrics when the target splits its local data into mini-batches of size $B$,} and performs one SGD step per mini-batch. %
\blue{In terms of the convergence speed and accuracy, prior works such as \cite{yin2018gradient,haddadpour2019convergence} have provided theoretical rules for the bound of mini-batch Size $B$ and proven that larger $B$ can lead to faster convergence speed.}
At one extreme $B=\infty$, the entire \LocalBatch is treated as one mini-batch, and this becomes FedSGD.  At the other extreme, $B=1$, there is one SGD step per local data point, which maximizes privacy.
Smaller $B$ values decrease RMSE, especially for lower $\eta$, but for the default $\eta=0.001$ the RMSE is not significantly affected. In addition to EMD, we show in Fig. \ref{fig:1w_B_out_of_range} the fraction of diverged DLG attacks. \blue{As $B$ decreases, the fraction of diverged attacks increases  (which makes the attack less accurate) due to increased gradient descent steps. It also makes the attack more expensive (in terms of execution time and DLG iterations).} We choose $B=20$ %
(3rd marker in Fig. \ref{fig:minibatch_comparison}) and the lower $\eta=$1e-05 in order to get some privacy protection (EMD increases slightly) and to maximize utility.

\textbf{Impact of Local Epochs $E$.} Another parameter of FedAvg that affects the number of SGD steps is the number of epochs $E$, \ie the number of local passes on the dataset, during local training. We set $B=20$, based on the previous experiment and we evaluate the impact of local epochs for two learning rates $\eta$. Fig. \ref{fig:sgd_local_epochs_comparison} shows that increasing $E$ increases  privacy both in terms of EMD and divergence. It also improved utility (not shown): can reduce RMSE from 6.25 to 4.75dbm.

\textbf{Putting it together.} We choose $E=5$ and $B=20$, which together provide improved privacy and utility. %
In summary, averaging in FedAvg provides some moderate protection against DLG attacks, which is also  consistent with Insight I3. 
However,  even with these parameters, the frequently visited locations (\ie home and work) of the target can still be revealed (\eg see Fig. \ref{fig:1w_B_20_E5}), which motivated us to design the following algorithm for further improvement.

\subsection{FedAvg with Diverse Batch \label{sec:manipulation}}
\blue{
{\bf Intuition.} So far, we have considered that every user, including the target,  processes {\em all} the local data {\em in the order they arrive} during round $t$, \ie in line 16 of Algorithm \ref{alg:FedAvg} it is \LocalBatch= $D^{target}_t$. %
The local averaging in FedAvg prevents the server from obtaining the real gradient, thus providing some %
protection. %
We can do better by exploiting the key observation (O1) supported analytically by insight (I1): %
the variance of locations in a batch affects how far the reconstructed %
$x_{DLG}$ is from the batch centroid $\bar{x}$. If the target preprocesses the data to pick a subset \LocalBatch $\subseteq D^{target}_t$, so that the selected locations have high variance, then we can force the DLG attack to have high $|x_{DLG}-\bar{x}|$ and possibly even diverge.}

{\bf \algoname Algorithm.} There are many ways to achieve the aforementioned goal. We designed  \algoname to maximize variance of locations in \LocalBatch, using DBSCAN clustering. %
In each FL round $t$, the target does the following at line  (16) of Algorithm \ref{alg:FedAvg}: 
\begin{enumerate}
\item The data $D^{target}_t$ that arrived during that round $t$ are considered candidates to include in \LocalBatch.
\item Apply DBSCAN on those points and identify the clusters.
\item Pick the center point from each cluster and include it in \LocalBatch; this intuitively increases variance. 
\item If more datapoints are needed, remove the selected points from   $D^{target}_t$ and repeat steps 1, 2  recursively on the remaining data, until the desired \LocalBatch size is reached. %
\end{enumerate}
We refer to this selection of \LocalBatch~as \emph{\algoname} and it is applied in line (16) of Algorithm \ref{alg:FedAvg}. After that, FedAvg continues as usually, potentially using mini-batches and multiple epochs. %
Appendix \ref{sec:additional_results} shows more details.%

\begin{figure*}
	\centering
	 \begin{subfigure}{0.33\linewidth}
		\centering
    \includegraphics[width=0.99\linewidth]{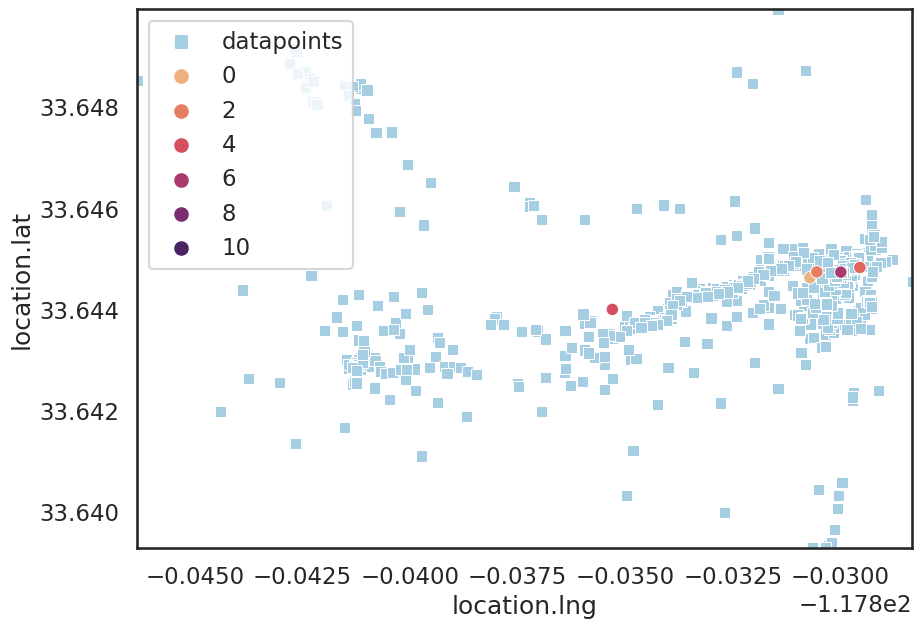}
		\caption{FedAvg: B=20, E=5: EMD=9.7,  %
		RMSE=4.83.}
	\label{fig:1w_B_20_E5}
    \end{subfigure}
	\begin{subfigure}{0.33\linewidth}
	\centering
	\includegraphics[width=0.99\linewidth]{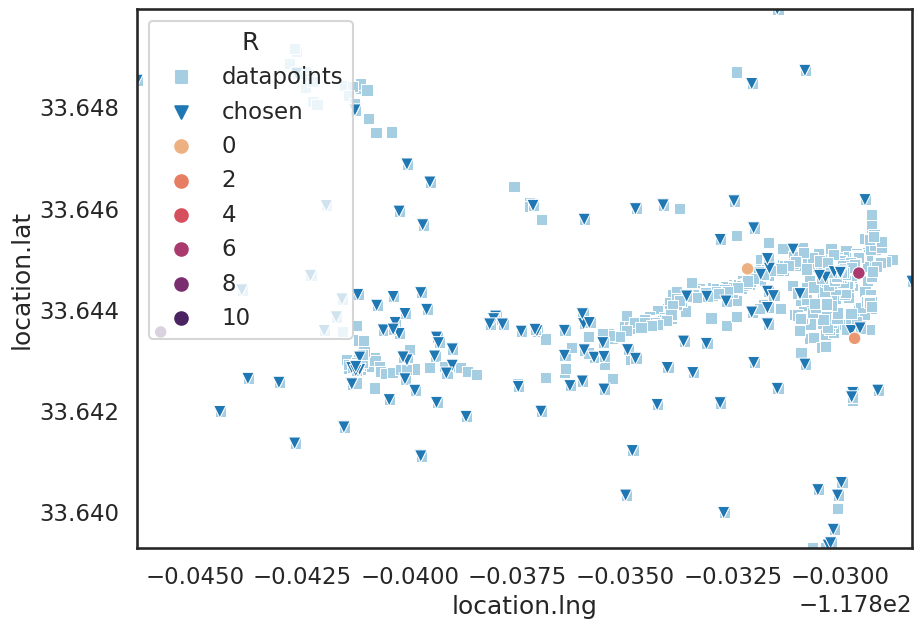}
	\caption{\blue{Diverse Batch (selects dark blue points):} EMD=15.23, %
	RMSE=4.93. }
	\label{fig:online_1w_eps_005_E_5_B_20}
  \end{subfigure}
  	 \begin{subfigure}{0.33\linewidth}
	\centering
	\includegraphics[width=0.99\linewidth]{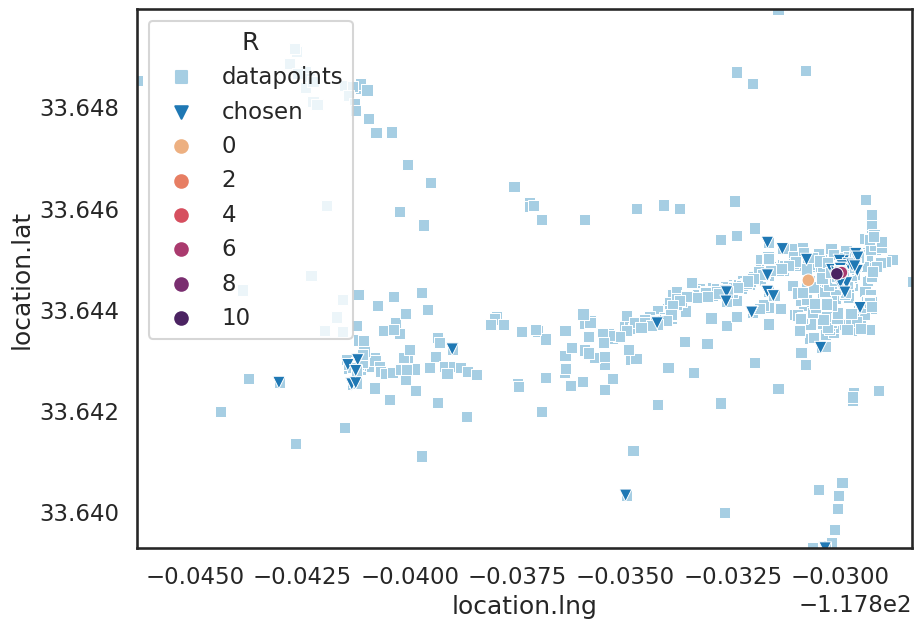}
	\caption{Random baseline to Diverse Batch: %
	 EMD=7.6.} %
	\label{fig:online_1w_eps_005_baseline}
  \end{subfigure}
\vspace{-5pt}
\caption{\small \textbf{FedAvg with \algoname.} Light blue shows the real locations of the target in \campus for $T=1$ week. \blue{Dark blue shows the points chosen  by \algoname with $eps=0.05$km.} %
The colors show the reconstructed locations by DLG.} %
 \vspace{-5pt}
\label{fig:batch_manipulation_comparison_1w}
\end{figure*}

{\bf Data Minimization.} In our implementation of \algoname, the target applies step 3 above exactly once, and skips step 4. %
As a result, \algoname uses significantly fewer points, and thus has a data minimization effect, in addition to high variance. For example, with $eps=0.05$km, \algoname achieves EMD=15.23 using 1\% of the location compared to a random sampling of 1\% of the points that leads to EMD=7.6; see Fig. \ref{fig:batch_manipulation_comparison_1w}(b) vs. \ref{fig:batch_manipulation_comparison_1w}(c), %
as well as Table \ref{tab:dbscan_stats_online_1W}. %

\begin{table*}[t]
\centering
\smallskip\noindent
\resizebox{\linewidth}{!}{%
\begin{tabular}{@{}c|cc|c|ccc|c@{}}
\toprule
\textbf{$eps$} & \textbf{Avg B} & \textbf{\% chosen} & \textbf{RMSE}  & \textbf{EMD}  & \textbf{\% diverged} & \textbf{Avg Dist (m)} & \textbf{Random}  %
\\
(km) & size & points & \textbf{FedSGD/FedAvg} & \textbf{FedSGD/FedAvg} & \textbf{FedSGD/FedAvg} & \textbf{FedSGD/FedAvg} & RMSE/EMD   
\\ \midrule
0.0001 & 239 & 6.84 & 5.24/4.88 %
& 9.61/10.59 & 18/82 & 139/265 & 4.82/7.5 %
\\
0.001 & 180 & 5.2 & 5.34/4.86 %
& 9.72/10.84 & 18/90 & 134/245 & 4.83/7.9 %
\\
0.005 & 98 & 2.8 & 5.78/4.83  %
& 10.72/14.15 & 9/57 & 170/290 & 4.82/7.9 %
\\
0.05 & 16 & 0.45 & 8.78/4.93 %
& 14.524/15.23 & 13/64 & 331/345 & 4.96/7.6 %
 \\ \bottomrule
\end{tabular}
} %
\caption{{\small {\bf  Performance of \algoname.} {\em Parameters:} $T=$1-week; DBSCAN is run once in each round; $\eta=0.001$, dropout=0.05.
When two numbers are reported (X/Y) they correspond to FedSGD and FedAvg ($B=20, E=5$), respectively.
For each value of the main parameter $eps$ of DBSCAN we report the following metrics. Since  \algoname picks \LocalBatch of different size in every round, we report the {\em average batch size}. Since it picks a subset of all data \LocalBatch $\subset D_t^k$, we report of the {\em \% of datapoints chosen}. The utility ({\em RMSE}) is not significantly affected by $eps$, but is improved by FedAvg, as expected. For privacy, we report the {\em EMD} between reconstructed and real locations, the {\em \% of diverged attacks}, %
and the {\em average distance} $|x_{DLG}-\bar{x}|$ in meters. In general, as $eps$ increases, privacy increase.  In the last column, as a baseline for comparison, we report the utility and privacy if the same number of points as in column 3 are picked uniformly at random: the EMD is approx. half.
}} %
\label{tab:dbscan_stats_online_1W}
\end{table*}

\begin{figure*}[t!]
		\centering
	\begin{subfigure}{0.24\linewidth}
		\centering
    \includegraphics[width=0.99\linewidth]{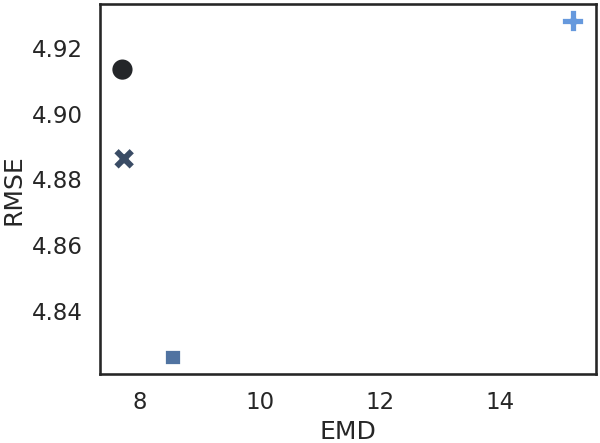}
		\caption{EMD.} 
		\label{fig:1_week_comparison_EMD}
		\end{subfigure}
	\begin{subfigure}{0.2\linewidth}
		\centering
    \includegraphics[width=0.99\linewidth]{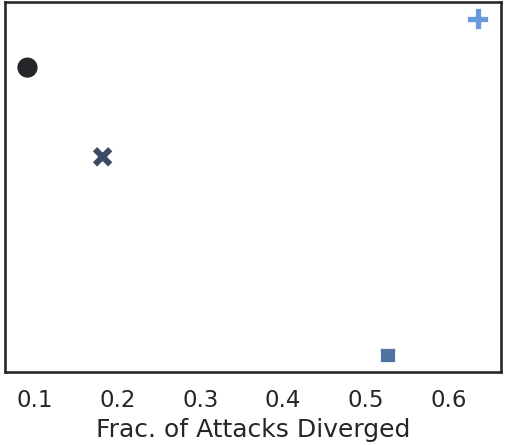}
		\caption{Diverged Attacks.} 
		\label{fig:1_week_comparison_EMD_out_of_range}
		\end{subfigure}
	\begin{subfigure}{0.2\linewidth}
		\centering
		\vspace{-1pt}
    \includegraphics[width=0.99\linewidth]{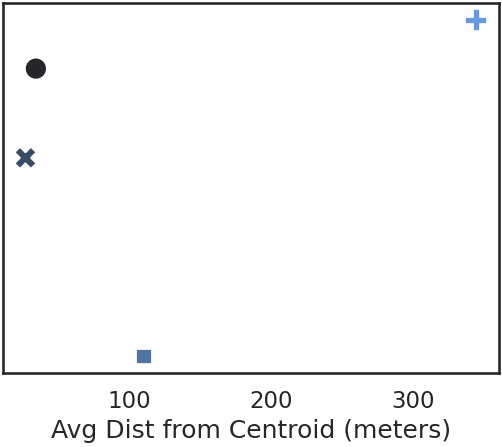}
		\caption{Distance from centroid.} 
        \vspace{-9.5pt}
		\label{fig:1_week_comparison_EMD_avg_dist}
		\end{subfigure}
	\begin{subfigure}{0.34\linewidth}
		\centering
    \includegraphics[width=0.99\linewidth]{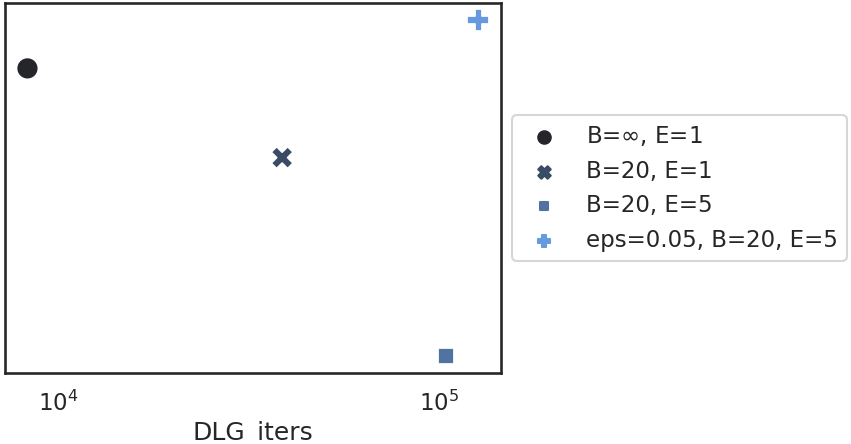}
		\caption{Avg DLG iterations.} 
		\label{fig:1_week_comparison_dlg_iters}
		\end{subfigure}
 \vspace{1pt}
\caption{\small \textbf{Privacy-utility trade-offs for all approaches.} 
Setup: \campus, 1-week intervals.
Algorithms: FedSGD ($B=\infty, E=1$), FedAvg ($B=20, E=5$), and \algoname. 
Privacy metrics: EMD, divergence, distance from centroid.  Utility: RMSE.
The DLG attack is strongest in FedSGD. FedAvg improves both privacy and utility. FedAvg with \algoname improves privacy (doubles EMD, increases divergence above 60\%, and distance from 50 to 350m), without significantly hurting RMSE,
and while using $<$ %
1\% of the data. } %
 \vspace{-5pt}
\label{fig:all_comparison_1week}
\end{figure*}

\blue{
{\bf Tuning \algoname.} First, we tune the learning rate $\eta= 0.001$ via a grid search; see Appendix \ref{sec:additional_results} for details.  An important parameter of DBSCAN is $eps$, which controls the maximum distance between points in a cluster: higher $eps$ leads to fewer clusters, \ie less datapoints chosen in \LocalBatch,  but higher variance in the coordinates of points in \LocalBatch. Table \ref{tab:dbscan_stats_online_1W} shows the performance of \algoname for various $eps$ values. We make the following observations. First, as expected, increasing $eps$ decreases the number of training points and results in smaller batches, due to fewer clusters. %
Second, w/o averaging, the RMSE increases with $eps$;  with $B=20, E=5$,  RMSE is not affected by $eps$. On the other hand, EMD increases for larger $eps$ in both cases, but with averaging EMD is higher overall. The percentage of diverged  attacks increases with $eps$, which makes the attack more expensive and less accurate. %
Third, we consider two baselines: (i) the same number of points as DBSCAN in each round, but randomly selected; %
(ii) FedAvg with $B=20$, $E=5$ but \LocalBatch=$D^{target}_t$:  the EMD remains low regardless of $eps$ since the batch variance is not controlled.}

\textbf{Performance of FedAvg with \algoname.} Fig. \ref{fig:batch_manipulation_comparison_1w} shows that adding \algoname to FedAvg ($B=20, E=5$) significantly improves privacy from $EMD=9.7$ to $EMD=15.23$. Random sampling of the same random of points per batch performs worse ($EMD=6.6$), because it preserves the spatial correlation in trajectories. This indicates that the main benefit of \algoname is controlling the batch location variance, rather than data minimization.

In Fig. \ref{fig:all_comparison_1week}, we compare all methods (in increasing privacy: FedSGD, FedAvg with mini-batches and epochs, and \algoname) w.r.t. both privacy (EMD, divergence, distance from centroid) and utility (RMSE). (a) EMD starts at 7.5  in FedSGD, increases in  FedAvg, and almost doubles (15) in \algoname. (b) In FedSGD attack, there is almost no divergence, while with \algoname more than 60\% of the attacks diverge. Divergence makes the DLG attack more expensive, since the attacker needs to relaunch the attack with other initializations. (c) We also report the distance between the reconstructed location and the centroid of the batch $|x_{DLG} - \bar{x}|$: in FedSGD, this is less than 30 meters, while with 
\algoname it increases to more than 350 meters.

\subsection{FedAvg with Farthest Batch}\label{sec:improv_DiveseBatch} 
\blue{
{\bf Intuition.} 
There are many ways to select the local batch so as to mislead the attackers, \ie achieve high $|x_{DLG}-\bar{x}|$. We already presented \algoname to maximize location variance in \LocalBatch, by picking a subset of points from different DBSCAN clusters. %
An alternative way is \improalgoname that we present here: we pick the subset of  $num$ points to include in \LocalBatch $\subseteq D^{target}_t$, from the DBSCAN cluster that is farthest away (\ie has the highest distance from the batch's true centroid $\bar{x}$). %
According to our key observation (O1), the DLG attack will infer $x_{DLG}$ close to the average location of the selected \LocalBatch,  %
 which is, by construction, far away from the true average location $\bar{x}$ in the entire batch $D^{target}_t$ in round $t$.%
}

\begin{algorithm}[h!]
	\small
    \DontPrintSemicolon
    
    \SetAlgoNoEnd
	\SetAlgoNoLine
	
	Given: $K$ users (indexed by $k$); $B$ local mini-batch size;
	$n_t$ is the total data size from all users at round t, $\eta$ is the learning rate; the server aims to reconstruct the local data of target user $k$. $w$ received from server, $eps$ the DBSCAN parameter, $num$ the number of points to select.\\
	{\bf UserUpdate($k,w,t, B,eps,num$):}\\
	\SP  Local data $D_t^k$ are collected by user $k$ during round $t$\\
	\SP  Select \LocalBatch $\subseteq D_t^k$ to use for training\\
	\SP $C_t^k$=$DBSCAN$($D_t^k, eps$) \\
	\SP $n_t^k$=$Sort\_ and\_ select$($C_t^k, num$)  \\
	\SP $B_t^k \leftarrow$ (split \LocalBatch into mini-batches of size $B$)\\
	\SP  \For{each local epoch i: 1...$E$}{%
		\SP 	\For{mini-batch $b\in B_t^k$} {
			\SP $w \leftarrow w - \eta\nabla\ell(w; b)$\\
		}
	}
	\SP  \Return{$w$ to server}
\caption{\improalgoname}
\label{alg:improDB_sec4}	
\end{algorithm}

{\bf \improalgoname.} 
In each FL round $t$, the target does the following at line  (16) of Algorithm \ref{alg:FedAvg}: 
\begin{enumerate}
\item Consider data $D^{target}_t$ that arrived during that round as candidates to include in \LocalBatch.
\item Apply DBSCAN on those points and identify the clusters, as shown at  the line (10) of Algorithm \ref{alg:improDB_sec4}.
\item Sort the clusters  in decreasing distance between the centroid of that cluster and the true centroid of entire \LocalBatch. Then, pick $num$ data points, from the farthest to the closest clusters, and include them in \LocalBatch, as shown at the line (11) of Algorithm \ref{alg:improDB_sec4}.
\end{enumerate}

\begin{figure}[t!]
	\centering
    \begin{subfigure}{0.8\linewidth}
	    \centering
        \includegraphics[width=0.99\linewidth]{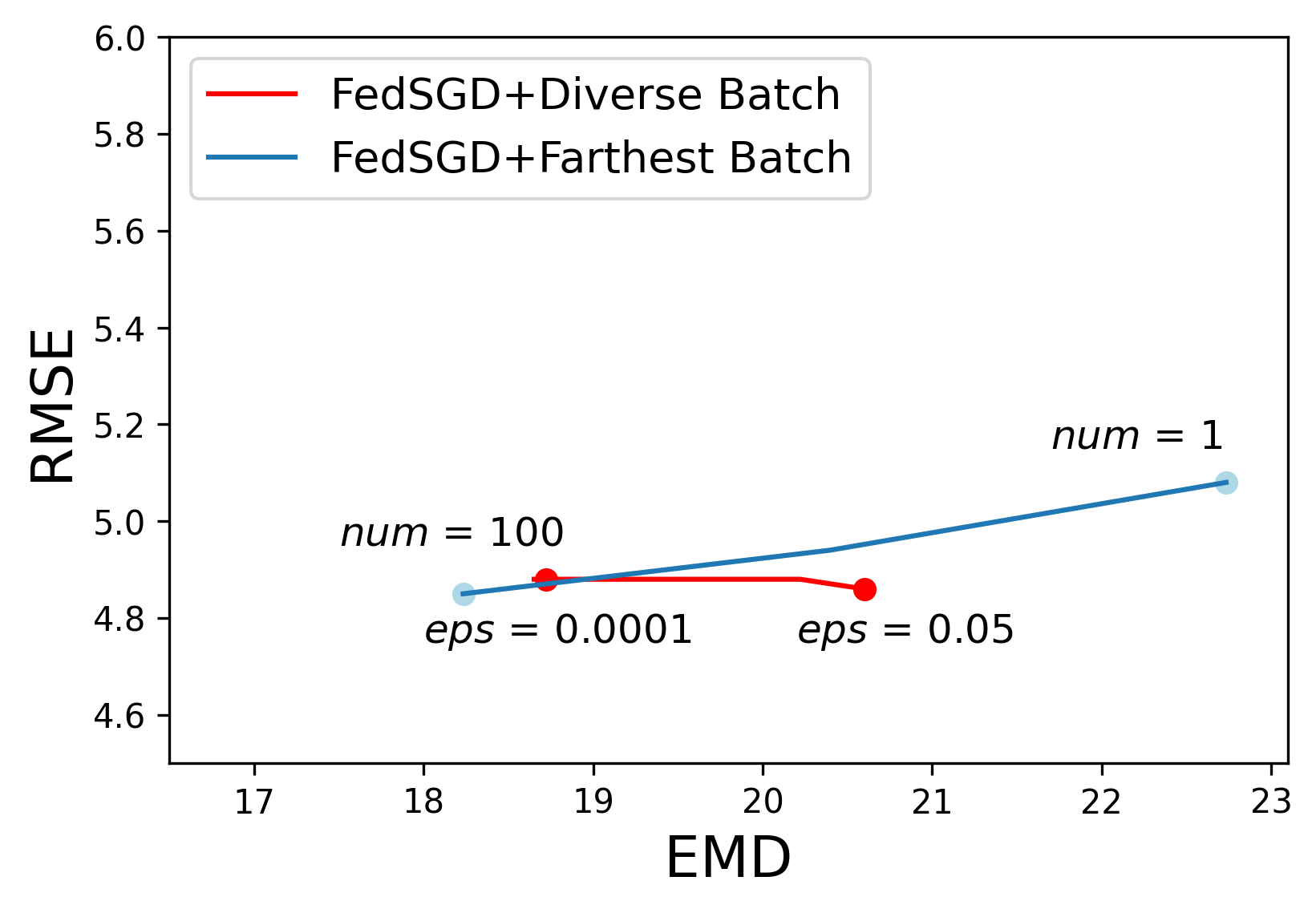}
    	\caption{\small RMSE vs. EMD.}
    	\label{fig:DBSCAN_Comparison_1daySGD_EMD}
    \end{subfigure}
         \begin{subfigure}{0.8\linewidth}
	    \centering
    	\vspace{10pt}
        \includegraphics[width=0.99\linewidth]{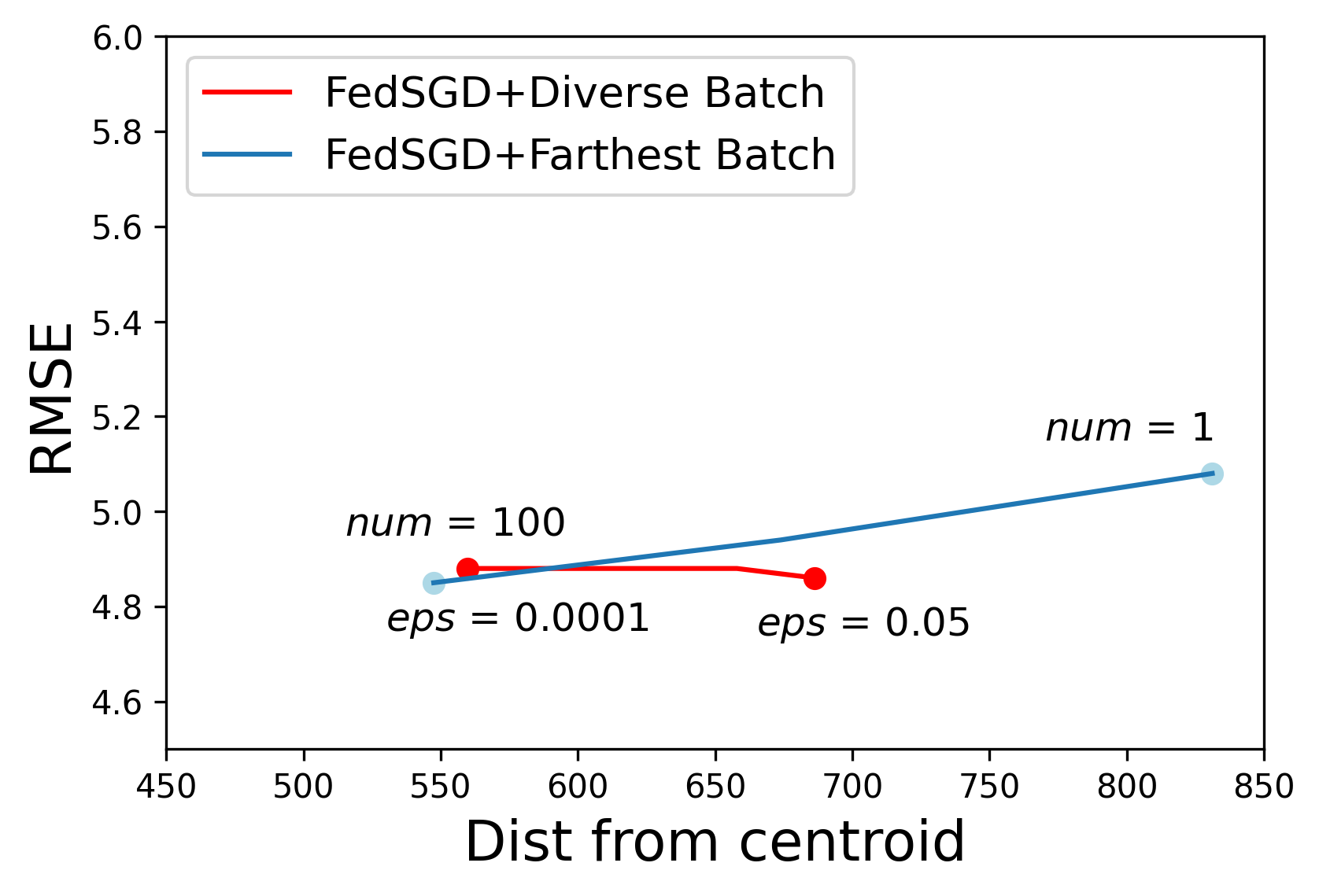}
    	\caption{\small RMSE vs. Distance from centroid.}
    	\label{fig:DBSCAN_Comparison_1daySGD_DIST}
    \end{subfigure}
    \caption{\small Comparison between \algoname and \improalgoname for 1-day interval using FedSGD.
    }
    \label{fig:DBSCAN_Comparison_1day_SGD}
\end{figure}

\begin{figure}[t!]
	\centering
    \begin{subfigure}{0.8\linewidth}
	    \centering
        \vspace{-1pt}
        \includegraphics[width=0.99\linewidth]{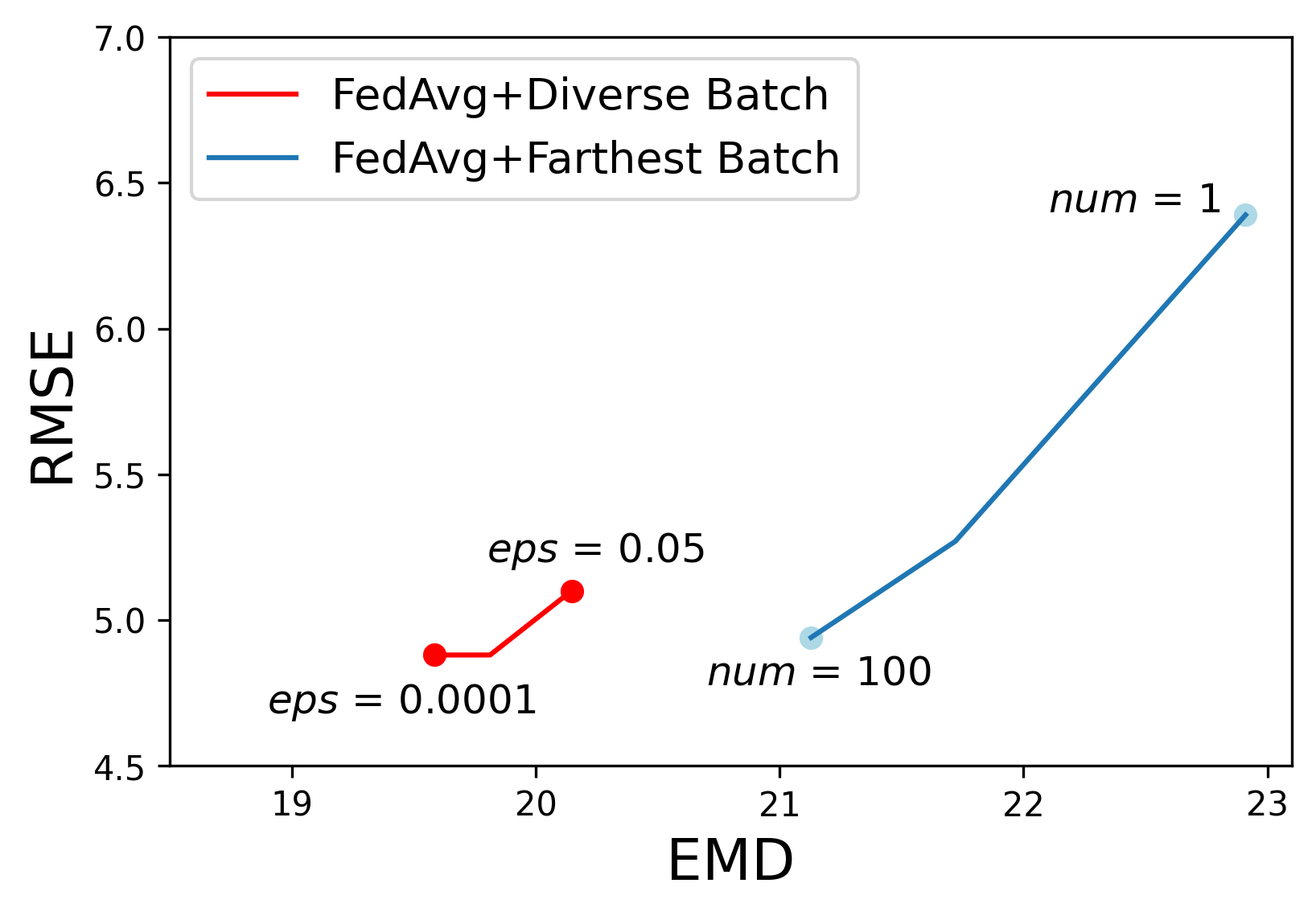}
    	\caption{\small RMSE vs. EMD.}
    	\label{fig:DBSCAN_Comparison_1dayAvg_EMD}
    \end{subfigure}
	\begin{subfigure}{0.8\linewidth}
	    \centering
    	\vspace{8pt}
        \includegraphics[width=0.99\linewidth]{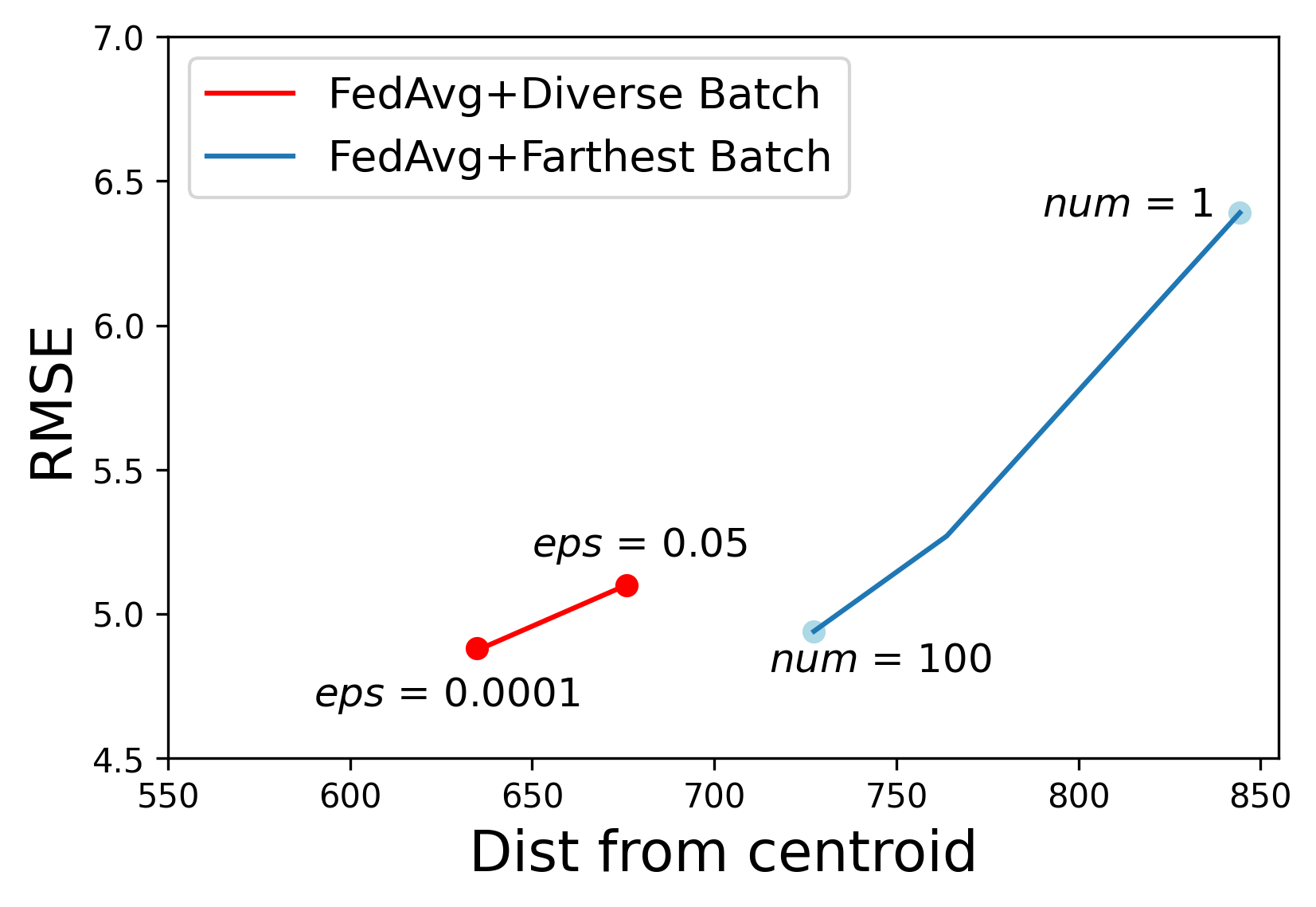}
    	\caption{\small RMSE vs. Distance from centroid.}
    	\label{fig:DBSCAN_Comparison_1dayAvg_DIST}
    \end{subfigure}
    \caption{\small Comparison between \algoname and \improalgoname for 1-day interval using FedAvg.
    }
    \label{fig:DBSCAN_Comparison_1day_Avg}
\end{figure}

\blue{
{\bf Tuning \improalgoname.} For learning rate $\eta$, we adopt the same values $\eta = 0.001$ as in \algoname. Regarding the DBSCAN parameter $eps$, we select $eps = 0.05 $, which shows the best performance in Table \ref{tab:dbscan_stats_online_1W}. An important parameter of \improalgoname is $num$, which controls the number of measurements selected in each \LocalBatch. When $num = 1$, it means we only pick one measurement from the farthest cluster and include it into LocalBatch. Also, by choosing $num = 1$, it will provide the best privacy protection, and the worst utility. By increasing $num$, there are more measurements selected from farthest to closest clusters, which improves utility and degrades privacy.

\textbf{Performance of FedAvg with \improalgoname.}
Fig. \ref{fig:DBSCAN_Comparison_1day_SGD} and \ref{fig:DBSCAN_Comparison_1day_Avg} show that, when compared to \algoname, \improalgoname can enhance the privacy from $EMD = 20.147$ to $EMD = 22.91$ and from $Dist = 675.9$ to $Dist=844.35$. in FedAvg. Although utility degrades when privacy increases, the loss of utility is still in a reasonable and acceptable range. This motivates us to use \improalgoname for better privacy protection in this setting.}

\blue{\textbf{Discussion: Local Batch Selection.}} \blue{In vanilla FL, local batches are typically randomly selected. To the best of our knowledge, we are the first to notice} that there can be many ways to pick a subset \LocalBatch of all local measurements, collected and processed in round $t$, $D^{target}_t$, so as to construct $\bar{x}$ that is far from the true centroid, thus misleading the DLG attacker as per key observation O1 \blue{-- which is specific to our setting. Which local batch selection algorithm performs better depending on the characteristics of the mobility patterns (e.g. how many important locations/clusters there are and how fast they change), as well as the time scale $T$ over which new data arrive and are processed in a batch.} In this paper, we proposed two intuitive such algorithms for local batch selection (\algoname and \improalgoname) and we showed that they achieve good privacy protection, without significant degradation in prediction, with \improalgoname outperforming \algoname in our evaluation. This good privacy-utility tradeoff was achieved while operating strictly within the "native" FL framework, and {\em without} orthogonal add-ons, such as DP that is discussed next.

\subsection{\blue{Baselines:  DP, GeoInd, GAN Obfuscation}\label{sec:DP}}
\begin{figure}[t!]
	\centering
   \begin{subfigure}{0.8\linewidth}
	    \centering
        \includegraphics[width=0.99\linewidth]{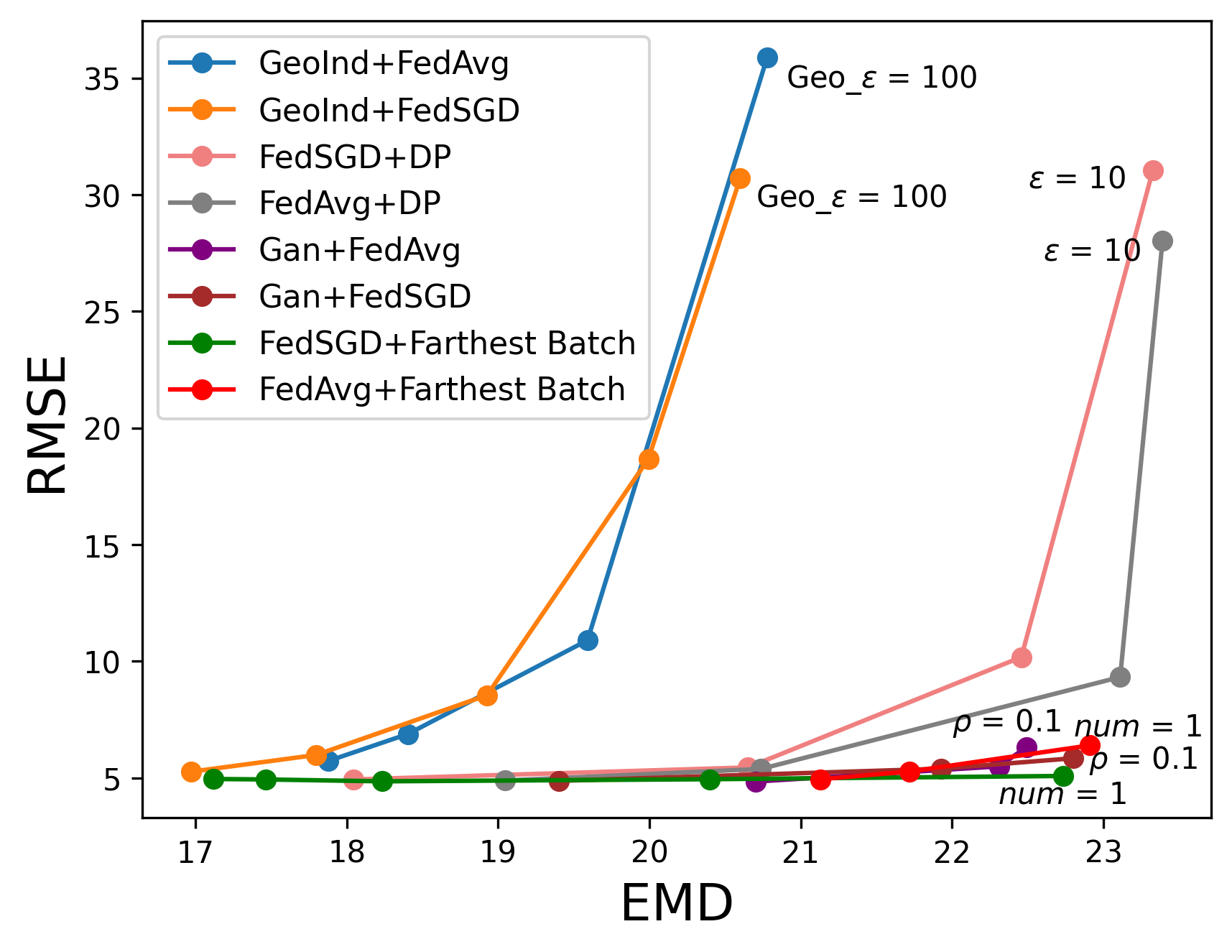}
    	\caption{\small RMSE vs. EMD.}
    	\label{fig:All_Comparison_1daydp_EMD}
    \end{subfigure}
    \begin{subfigure}{0.8\linewidth}
	    \centering
    	\vspace{10pt}
        \includegraphics[width=0.99\linewidth]{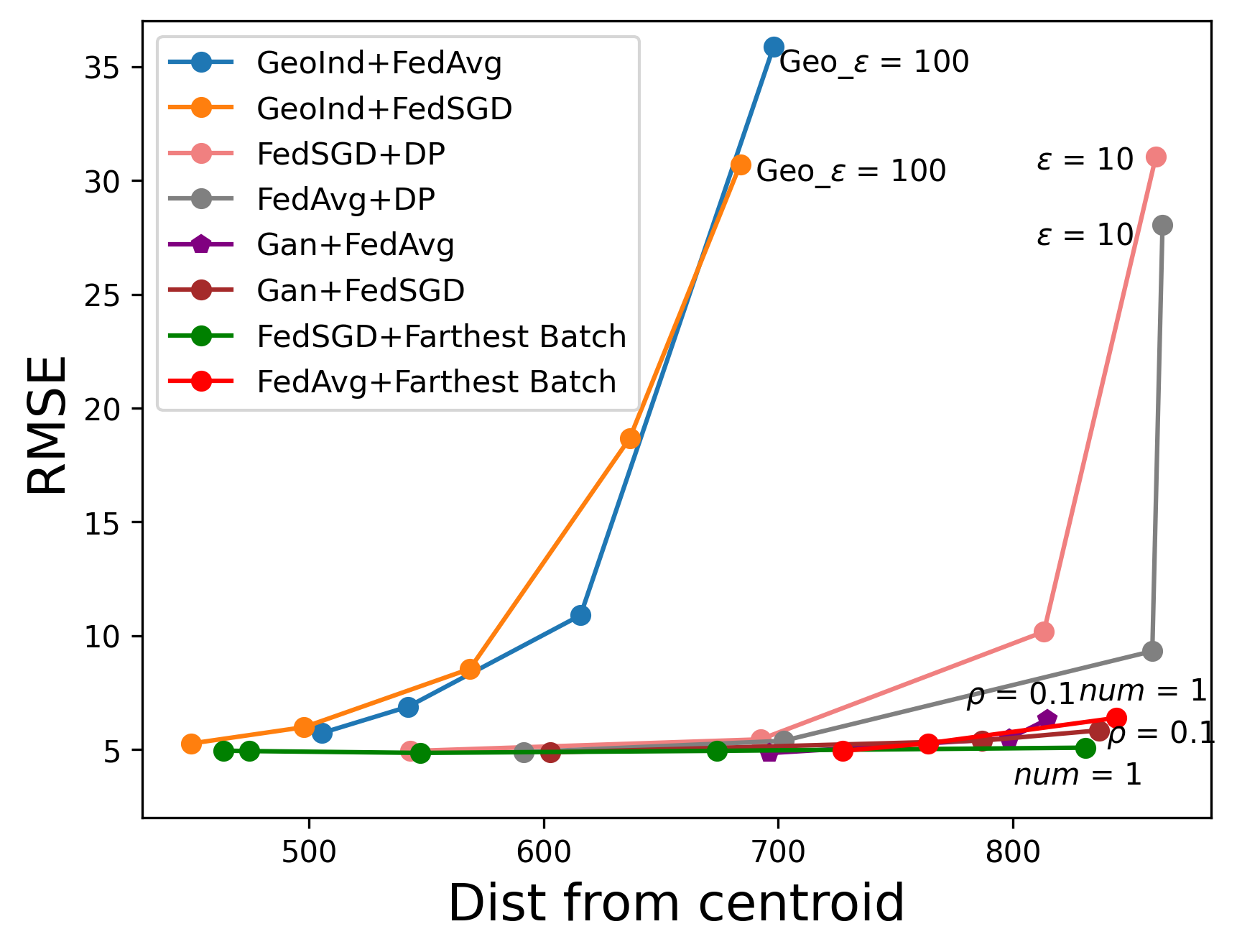}
    	\caption{\small RMSE vs. Distance from centroid.}
    	\label{fig:All_Comparison_1daydp_DIST}
    \end{subfigure}
    \caption{\blue{\small Comparison between Baselines (DP, GeoInd, Gan) and \improalgoname, for rounds of 1-day, in terms of the privacy-utility tradeoff they achieve. Privacy is captured by two metrics: EMD and Dist from centroid. Model performance is captured by the prediction error (RMSE).
    }}
    \label{fig:All_Comparison_1day_dp}
\end{figure}

\blue{
So far, we enhanced privacy through native FL mechanisms, \ie tuning the parameters and local batch selection. In this section, we explore how much benefit we get by adding state-of-the-art (but orthogonal) defenses, such as Differential Privacy (DP), Geo-Indistinguishability (GeoInd) \cite{andres2013geo} and Gan Obfuscation (Gan) \cite{zhang2022privacy}, with both FedSGD and FedAvg and we compare it to \improalgoname alone. We use $RMSE$ as the utility metric; $EMD$ and $Distance$ $from$ $centroid$ as privacy metrics separately. 

\textbf{DP.} In the FL setting, differentially private (DP) noise is typically added to the gradients of each \LocalBatch, as in \cite{wei2020federated,truex2020ldp}. %
Before transferring gradients from each user to the server, the gradients will first be clipped and then DP noise will be added. Clipping can bound the maximum influence of each user and we are using the fixed clipping method with parameter $\mathbb{C}$ in our experiments. We use ($\epsilon$, $\delta$)-DP bound for the Gaussian mechanism with noise N(0, $\sigma^{2}$), where $\sigma$ = $\sqrt{2\log\frac{1.25}{\delta}}\cdot\frac{\mathbb{C}}{\epsilon}$, $\delta = \frac{1}{\Vert LocalBatch\Vert}$. We choose $\mathbb{C}$ = 1 and $\delta$ =1e-5.

{\bf GeoInd.} The mechanism of Geo-Indistinguishability (GeoInd) \cite{andres2013geo} is a variant of DP and is designed to provide strong privacy guarantees, specifically for location-based applications, by adding spatially controlled  local noise to the user’s location data. We use $Geo\_\epsilon$ to represent the noise parameter of GeoInd in our evaluations. 

{\bf GAN.} Generative adversarial networks (GANs) have been applied before to crowdsourced signal maps, stored on a server, to protect the privacy of  users \cite{zhang2022privacy}. Here, we  apply  Gan Obfuscation to obfuscate users’ private data  before the data leaves the mobile device. The goal is to increase privacy such that it is difficult to recover sensitive features from the obfuscated data (\eg user ids and user whereabouts), while still allowing network providers to obtain accurate signal maps to improve their network services. We use $\rho$ to represent the obfuscation level of Gan.

 {\bf Comparison.} By applying DP, GeoInd and Gan locally, each user can protect their private training data from DLG attacks. %
Figure \ref{fig:All_Comparison_1day_dp} compares the local batch selection approach (using the best of our two algorithms \improalgoname) against DP, GeoInd and Gan for 1-day intervals, in terms of the privacy-utility tradeoff they achieve.  
 We observe that as $\epsilon$, $Geo\_\epsilon$ and $\rho$ decrease, the privacy (in terms of $EMD$ and $Distance$ $from$ $centroid$) improves, while the model performance captured by RMSE)  deteriorates.
We also observe that for the same privacy level, our \improalgoname can provide more utility compared to differential privacy, GeoInd and Gan. Especially, when $EMD$ is higher than 21 and $Distance$ $from$ $centroid$ is larger than 680, the utility loss in \improalgoname is much smaller. Although Gan could provide similar utility as \improalgoname for the same privacy level, Gan requires more computation resources since two additional neural networks need to be trained  (i.e. a generator and an adversary) and it also requires hundreds of training rounds for these two neural networks to converge. Compared with Gan, our local batch selection approach is more efficient and lightweight. In summary, optimizing the batch selection to mislead the DLG attack (via \improalgoname) achieves a better privacy-utility tradeoff compared to all three baselines. Please see Appendix \ref{sec:Differential privacy Comparison} for additional evaluation results between \algoname, \improalgoname, DP, GeoInd and Gan.

\textbf{Discussion.}  
The focus of this paper has been the privacy-enhancing design of FL-native local mechanisms for privacy, such as the tuning of local parameters and the local batch selection (\eg \algoname and \improalgoname.) These native mechanisms are necessary ingredients of FL itself and orthogonal to add-ons such as DP, GeoInd, Gan or SecAgg. The comparisons to DP, GeoInd and Gan are provided as a baseline for comparison against state-of-the-art (local) defense mechanisms. In the future, these add-ons (DP, GeoInd and Gan) can co-exist with and be combined with our local batch selection approach (\algoname and \improalgoname)  further improve performance.
}

\subsection{Multiple Users}\label{sec:multiusers}

\begin{figure}[t!]
	\centering
    \begin{subfigure}{0.85\linewidth}
		\centering
    \includegraphics[width=0.99\linewidth]{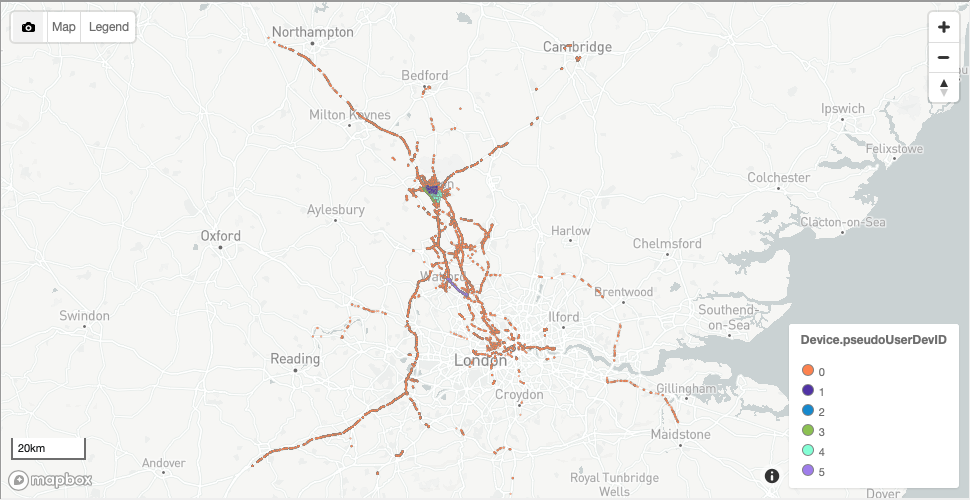}
		\caption{\small The entire London 2017 \radiocells.}
		\label{fig:radio_cell_x455_user_map_all_users}
    \end{subfigure}
 
    \begin{subfigure}{0.85\linewidth}
		\centering
    \includegraphics[width=0.99\linewidth]{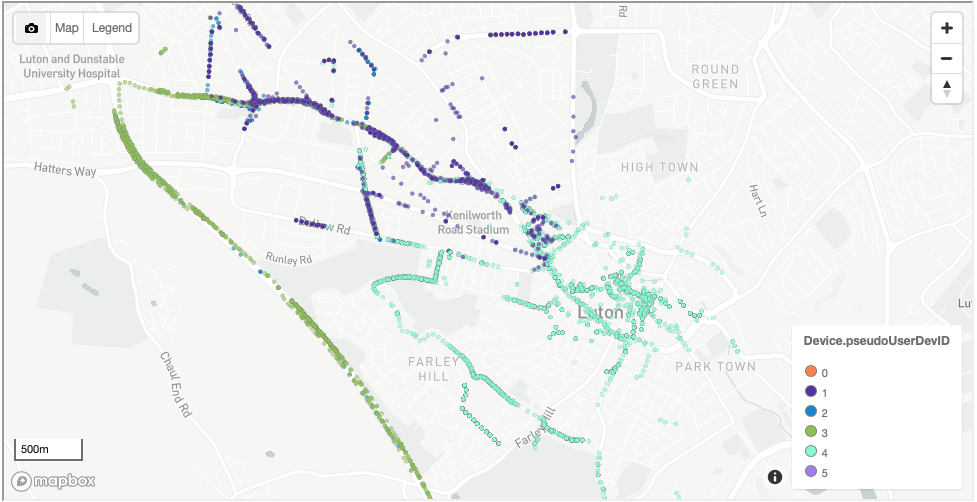}
		\caption{\small Zooming into the Luton area. %
		}
		\label{fig:radio_cell_x455_user_map_target_users}
    \end{subfigure}
\vspace{-5pt}
\caption{\small The London 2017 \radiocells, partitioned into ``synthetic users'' 0-5, for cell x455. Background (``user 0'', shown in orange,) contains the vast majority (1059 our of 1351) of upload files across the entire London region which are contributed by a large number of real users. Synthetic users 1-5 are also shown. %
Target is user 3 is shown in green.}
\vspace{-5pt}
\label{fig:radio_cell_x455_user_stats}
\end{figure}

Throughout this paper, we focus on the $target$ user exchanging local ($w_t^{target}$) and global ($w_t$ ) model parameter updates with the server, for $t=1,\dots,R$. When there are more users participating in FL, 
they contribute their updates and there is also global averaging of the gradients across users to produce $w_t$ (line 12 in Algorithm \ref{alg:FedAvg}).
The data used for updates across rounds may be different, depending on how similar the users' trajectories are.
 As discussed in (I4) in Sec. \ref{sec:analysis}, when the diversity of data (across rounds) increases, convergence slows down and the gradient magnitude $|\bar{g}|$ remains large for more rounds, which makes the DLG attack more successful.
Here, we evaluate the performance of all previous methods (FedSGD, FedAvg, FedAvg with \algoname) under DLG attack, still from the perspective of a single target user, but considering the presence of multiple users updating the global model.

\textbf{Creating Multiple Synthetic Users.} We use the London 2017 part of the \radiocells, described in Sec. \ref{sec:dataset}. %
Each upload file contains a sequence of measurements from a  single device, %
but without specifying user pseudo-ids. In order to create realistic trajectories for the simulation of FL,%
we use heuristics to concatenate upload files, and we refer to the result as ``synthetic'' users.\footnote{
We start by grouping upload files with identical cell and device information (\ie device
manufacturer, device model, software id and version, etc.). The device information  %
was also used in \cite{carmela} to create ``users'', but we also consider the cell id because we train a DNN model per cell. 
In addition, we merge files with the same device information  as follows:
if the start and end locations of two upload files are within a certain distance $Z$, then we assign them to the same ``synthetic user''. The intuition is that users typically repeat their trajectory across days, \ie  a user starts/ends logging data from/to the same places, such as home or work. By setting $Z=1$ mile, we obtain 933 potential synthetic users. We select five of them (users 1-5), each with a sufficiently high number of measurements per week %
 and spanning several weeks. We merge {\em all} remaining upload files of the dataset (1059 out of 1351) into one ``background'' user 0.} 
The entire dataset (partitioned into synthetic users 0-5) is depicted in Fig.\ref{fig:radio_cell_x455_user_stats}.
We confirmed visually and via pair-wise similarity (via EMD) that: users 1-4 are highly similar to each other, and dissimilar to users 0 and 5. %
We pick user 3, with the most measurements, as the target.

\begin{table}
\centering
\begin{tabular}{@{}ccccc@{}}
\toprule
\textbf{Scheme} & \textbf{User(s)} & \textbf{RMSE} & \textbf{EMD} & \textbf{\% diverged} \\ \midrule
FedSGD  & user 3      & 6.1  & 17.08 & 65 \\
FedAvg & user 3      & 5.43 & 24.13 & 91 \\
FedAvg, $eps=0.005$ & user 3 & 5.42 & 25.2  & 97 \\
FedAvg, $eps=0.01$ & user 3 & 5.44 & 26.9  & 97 \\
\hline
FedAvg & user 3, 0  & 5.43 & 22.51 & 90 \\
FedAvg, $eps=0.01$  & user 3, 0 & 5.41 & 23.01  & 90 \\
\hline
FedAvg & user  3, 1 & 5.47 & 29.16  & 95 \\
FedAvg & user 3, 2 & 5.43 & 26.15  & 95 \\
FedAvg & user 3, 4 & 5.47 & 29.30  & 95 \\
\hline
FedSGD & user 3, 5 & 5.82 & 18.02  & 59 \\
FedAvg & user 3, 5 & 5.42 & 23.02  & 92 \\
FedAvg, $eps=0.01$  & user 3, 5 & 5.41 & 31.33  & 96 \\
\bottomrule
\end{tabular}
\caption{{\small The effect of multiple users on the DLG attack. Parameters: \radiocells, $\eta=0.001$; results are averaged over multiple runs. We compare FedSGD, FedAvg (with $E=5, B=10$) and FedAvg with \algoname (with $eps$ in km). Users 1,2,3,4 are similar to each other and dissimilar from users 0 (background) and 5; user 3 is the target.}}
\vspace{-10pt}
\label{tab:radiocells}
\end{table}

\textbf{Impact of Additional Users.} We simulate DLG on FL with the target and additional synthetic users participating. We report the results in Table \ref{tab:radiocells}.  
First, we consider the target (3) alone: the DLG attack achieves: EMD 17.08, under FedSGD; 24.13, under FedAvg;  up to 26.9, for FedAvg with \algoname and $eps=0.01$; the RMSE is roughly the same across all FedAvg and better than FedSGD. %

Next, we consider that the background user 0 joins and updates the global model after locally training on its data. %
This is a realistic scenario of DLG on target 3, where the effect of most other real users in London is captured by this massive "background'' user, who updates the global model following the timestamps indicated in the individual upload files. This results in a slight decrease of EMD for FedAvg from 24.13 to 22.51, while attack divergence remains practically the same. 

Finally, we consider the other synthetic users (1,2,4,5), joining the target (3) one at a time, and updating the global model via FedAvg.
When the most dissimilar user (5) participates, privacy loss is maximum (EMD is 18.02 with FedSGD and 23.02 with FedAvg). When similar users (1,2,4)  participate,  privacy loss is smaller (EMD $\in[26.1,29.3]$). This was expected from Insight I4 in Sec.\ref{sec:analysis}:  dissimilar users lead to slower convergence and more successful DLG attack.

\textbf{Adding \algoname} to FedAvg offers significant protection to the target: for $eps=0.005$, one third of the runs resulted in 100\% divergence. %
Increasing $eps$ to 0.01 resulted in two-thirds of runs with 100\% divergence, and the remaining had EMD=26.9 and RMSE=5.44. In the multi-users case, it is sufficient that the target applies  \algoname locally, to get privacy protection, \eg EMD=31.33 and 96\% divergence %
in the case of (3,5). This is amplified when users are dissimilar, like users (3,5), rather than the case of background (3,0). %

\section{Related Work}\label{sec:related_work}

\textbf{Signal Maps Prediction Framework.} There has been significant interest in signal maps prediction %
based on a limited number of spatiotemporal cellular measurements \cite{alimpertis2022unified}. These include  propagation models~\cite{raytracing:15, winnerreport} as well as data-driven approaches~ZipWeave \cite{fidaZipWeave:17}, SpecSense \cite{specsense:17}, BCCS \cite{heBCS:18} and combinations thereof~\cite{phillips:12}. Increasingly sophisticated machine learning models are being developed to capture various spatial, temporal and other characteristics of signal strength~\cite{ray:16,raik:18,alimpertis2019city} and throughput~\cite{infocomLSTM:17,milanMobihocDeepnets:18}.  %
The problem has been considered so far only in a centralized, not distributed setting. To the best of our knowledge, this paper is the first to consider signal maps prediction (i) in the FL framework (ii) considering online learning in the case  %
of streaming data and (iii) a DLG inference attack on location. %

\textbf{Online federated learning} is an emerging area  \cite{chen2020asynchronous, liu2020fedvision, damaskinos2020fleet}. To the best of our knowledge, existing work such as ASO-Fed \cite{chen2020asynchronous}, Fedvision \cite{liu2020fedvision}, Fleet \cite{damaskinos2020fleet}, in the online setting does not consider privacy leakage and focuses on convergence and device heterogeneity. \blue{RoF \cite{zhang2021optimizing} and \cite{zhang2021deep} focus on achieving efficient federated learning in industrial IoT but do not consider privacy leakage.} Different communities \cite{LAHA2018298} consider the case of online location data, although not in a FL way and with different goals (\eg location prediction, where the utility lies in the location itself).  %

\textbf{Location Privacy.} Numerous works evaluated location privacy or trajectory data, \eg  \cite{isaacman2011identifying, de2013unique, pyrgelis2017knock, inferring_activities,  becker2013human, bagdasaryan2021towards}, Federated RFF KDE \cite{zong2023privacy}, where the utility of the dataset lies in the location itself. In this work, we focus on location privacy in mobile crowdsourcing systems (MCS), similarly to \cite{carmela, pyrgelis2017knock}. As \cite{carmela} pointed out,  an important difference is that the utility in MCS %
does not lie in the location itself, but in the measurement associated with that location. In our case, the measurement is RSRP, and location is only a feature in an ML %
model. However, location is {\em the} primary feature needed to predict signal strength: additional features other than location and time (such as frequency, device information, environment \etc) bring only incremental benefit \cite{alimpertis2019city}.  %

\textbf{Reconstruction attacks and defenses based on gradients.}
It has been shown that observing gradients DLG \cite{dlg}, iDLG \cite{idlg} gradients (as in FedSGD) or model parameters \cite{geiping2020inverting}, ROG \cite{yue2022gradient} (as in FedAvg) %
can enable reconstruction of the local training data.  
There has been a significant amount of prior work in this area,  \cite{dlg, idlg, yue2022gradient, fedpacket, melis2019exploiting, geiping2020inverting, hitaj2017deep, chu2022securing, bonawitz2022federated}
to mention just a few representatives.
Since gradient/model updates are the core of federated learning, FL is inherently vulnerable to such inference attacks based on observing and inverting the gradients.

Such attacks have been mostly applied to reconstruct image or text training data, with a few exceptions such as FedPacket \cite{fedpacket} that inferred users' browsing history. %
``Deep leakage from gradients'' (DLG) \cite{dlg},  reconstructed training data (images and text) and their corresponding labels, from observing a  single gradient during training of DNN image classifiers,  without the need for additional models (\eg GANs \cite{hitaj2017deep}) or side information. %
DLG \cite{dlg} discussed potential defenses, including tuning training parameters such as mini-batch size, which we incorporate in our setting.
We consider an attack that builds on "inverting the gradients" in \cite{geiping2020inverting}:  using a cosine-based instead of the Euclidean distance, %
and also evaluating against FedAvg and the impact of averaged gradients due to local epochs.
In our setting, the attack reconstructs only one (the average) location  from a gradient update computed on N locations, as opposed to all N data points in a batch.

The work in \cite{elkordy2022much} points out that information about a user's dataset can still leak through the aggregated model at the server, and provides a first analysis of the formal privacy guarantees for federated learning with secure aggregation. The work in \cite{bagdasaryan2021towards} designs a scalable algorithm combining distributed differential privacy and secure aggregation to privately generate location heatmaps over decentralized data; however, the communication and computation costs brought by secure aggregation should be considered for mobile users.
In contrast, in our work (1) we only utilize  FL-"native" mechanisms  to achieve location privacy, without added DP or SecAggr; and (2) we consider an online FL setting, where users collect data in an online fashion, and process them in intervals of duration $T$. 

\blue{In Section \ref{sec:DP}, we  considered two more privacy-enhancing techniques, beyond just DP. Geo-Indistinguishability (GeoInd) is a state-of-the-art, privacy-preserving method, inspired by DP and modified specifically to protect location data \cite{andres2013geo}. The GAN obfuscation technique we adopted from \cite{zhang2022privacy}, applied generative adversarial networks to signal map datasets, stored at the server. We adapted the above techniques as well as DP, to be applied locally at the mobile.}

{\bf Our prior work.} In our prior work, we developed tools for collecting crowdsourced mobile data and uploading to a server \cite{alimpertis2017system}; the \campus collected therein is one we use for evaluation here as well. \cite{alimpertis2019city}, we formulated a framework for centralized signal maps prediction using random forests. In \cite{alimpertis2022unified},  we extended the framework to provide several knobs to operators and allow them to optimize prediction when training on data of unequal important and/or for different tasks. In \cite{zhang2022privacy}, we proposed a centralized GAN obfuscation technique to provide privacy for such tasks.

\section{Conclusion}\label{sec:conclusion}

{\bf Summary.} \blue{In this paper, we make three contributions. 
First, we design a lightweight online federated learning framework, specifically for the signal strength prediction {\em problem} in a crowdsourced setting. Second, we introduce a   {\em privacy attack}, specifically for this framework: an honest-but-curious server employs gradient inversion to infer the location of users participating in the federated signal map framework. This DLG  attack is specifically designed to reconstruct the average location in each round; this is in contrast to state-of-the-art DLG attacks on images or text, which aim at fully reconstructing all training data points. Third, we propose a {\em defense} approach that selects local batches so that the inferred location is far from the true average location, thus misleading the DLG attacker. Evaluation results show that our defense mechanisms achieve better privacy-performance trade-off compared to state-of-the-art baselines.} Ultimately, the success of the attack depends not only on the FL algorithm and defenses used, but also on the characteristics of the underlying user trajectory data.%

{\bf Future Directions.} 
First, in terms of applications:
\blue{signal maps prediction is a representative special case of predicting properties of interest based on crowdsourced spatiotemporal data. The same methodology to protect location privacy against DLG attacks can be applied to other learning tasks that rely on such crowdsourced measurements.}
Second, in terms of methodology, there are several directions for extension. 
Due to the characteristics of human trajectories, there are more dependencies and opportunities to explore in the design of DLG attacks and defenses.
Another direction is  exploiting similarities and differences in users' trajectories, to further  optimize aggregation schemes. Finally, this paper focused on designing and optimizing  local FL-"native" privacy-preserving algorithms that can provide inherent privacy protection; these can be combined with other state-of-the-art but orthogonal defenses, such as DP or Gan. %

\bibliographystyle{plain}
\bibliography{allrefs_sorted.bib}

\begin{thebibliography}{10}

\bibitem{radiocells}
{Radiocells Dataset}.
\newblock \url{https://radiocells.org}.

\bibitem{tutela}
{Tutela}.
\newblock \url{https://www.tutela.com}.

\bibitem{nytimesarticle}
{Your Apps Know Where You Were Last Night, and They’re Not Keeping It Secret}.
\newblock \url{https://nyti.ms/3a5VCbp}.

\bibitem{alimpertis2019city}
E.~Alimpertis, A.~Markopoulou, C.~Butts, and K.~Psounis.
\newblock City-wide signal strength maps: Prediction with random forests.
\newblock In {\em WWW}, 2019.

\bibitem{alimpertis2017system}
Emmanouil Alimpertis and Athina Markopoulou.
\newblock A system for crowdsourcing passive mobile network measurements.
\newblock {\em 14th USENIX NSDI}, 17, 2017.

\bibitem{alimpertis2022unified}
Emmanouil Alimpertis, Athina Markopoulou, Carter~T Butts, Evita Bakopoulou, and Konstantinos Psounis.
\newblock A unified prediction framework for signal maps: Not all measurements are created equal.
\newblock {\em IEEE Transactions on Mobile Computing}, 2022.

\bibitem{bagdasaryan2021towards}
Eugene Bagdasaryan, Peter Kairouz, Stefan Mellem, Adri{\`a} Gasc{\'o}n, Kallista Bonawitz, Deborah Estrin, and Marco Gruteser.
\newblock Towards sparse federated analytics: Location heatmaps under distributed differential privacy with secure aggregation.
\newblock {\em arXiv preprint arXiv:2111.02356}, 2021.

\bibitem{fedpacket}
E.~Bakopoulou, B.~Tillman, and A.~Markopoulou.
\newblock Fedpacket: A federated learning approach to mobile packet classification.
\newblock {\em IEEE Transactions on Mobile Computing}, 2021.

\bibitem{becker2013human}
R.~Becker, R.~C{\'a}ceres, K.~Hanson, S.~Isaacman, J.~M. Loh, M.~Martonosi, J.~Rowland, S.~Urbanek, A.~Varshavsky, and C.~Volinsky.
\newblock Human mobility characterization from cellular network data.
\newblock {\em Communications of the ACM}, 2013.

\bibitem{andres2013geo}
\blue{Andr{\'e}s, Miguel E and Bordenabe, Nicol{\'a}s E and Chatzikokolakis, Konstantinos and Palamidessi, Catuscia}.
\newblock \blue{Geo-indistinguishability: Differential privacy for location-based systems}.
\newblock In {\em \blue{Proceedings of the 2013 ACM SIGSAC conference on Computer \& communications security}}, pages \blue{901--914}, \blue{2013}.

\bibitem{french1999catastrophic}
\blue{French, Robert M}.
\newblock \blue{Catastrophic forgetting in connectionist networks}.
\newblock {\em \blue{Trends in cognitive sciences}}, \blue{3}(\blue{4}):\blue{128--135}, \blue{1999}.

\bibitem{haddadpour2019convergence}
\blue{Haddadpour, F. and Mahdavi, M.}
\newblock \blue{On the convergence of local descent methods in federated learning}.
\newblock {\em \blue{arXiv preprint arXiv:1910.14425}}, \blue{2019}.

\bibitem{kemker2018measuring}
\blue{Kemker, Ronald and McClure, Marc and Abitino, Angelina and Hayes, Tyler and Kanan, Christopher}.
\newblock \blue {Measuring catastrophic forgetting in neural networks}.
\newblock In {\em \blue{Proceedings of the AAAI conference on artificial intelligence}}, volume \blue{32}, \blue{2018}.

\bibitem{yin2018gradient}
\blue{Yin, Dong and Pananjady, Ashwin and Lam, Max and Papailiopoulos, Dimitris and Ramchandran, Kannan and Bartlett, Peter}.
\newblock \blue{Gradient diversity: a key ingredient for scalable distributed learning}.
\newblock In {\em \blue{International Conference on Artificial Intelligence and Statistics}}, pages \blue{1998--2007}. \blue{PMLR}, \blue{2018}.

\bibitem{zhang2022privacy}
\blue{Zhang, Jiang and Clark, Lillian and Clark, Matthew and Psounis, Konstantinos and Kairouz, Peter}.
\newblock \blue{Privacy-utility trades in crowdsourced signal map obfuscation}.
\newblock {\em \blue{Computer Networks}}, \blue{215}:\blue{109187}, \blue{2022}.

\bibitem{zhang2021deep}
\blue{Zhang, Weiting and Yang, Dong and Peng, Haixia and Wu, Wen and Quan, Wei and Zhang, Hongke and Shen, Xuemin}.
\newblock \blue{Deep reinforcement learning based resource management for DNN inference in industrial IoT}.
\newblock {\em \blue{IEEE Transactions on Vehicular Technology}}, \blue{70}(\blue{8}):\blue{7605--7618}, \blue{2021}.

\bibitem{zhang2021optimizing}
\blue{Zhang, Weiting and Yang, Dong and Wu, Wen and Peng, Haixia and Zhang, Ning and Zhang, Hongke and Shen, Xuemin}.
\newblock \blue{Optimizing federated learning in distributed industrial IoT: A multi-agent approach}.
\newblock {\em \blue{IEEE Journal on Selected Areas in Communications}}, \blue{39}(\blue{12}):\blue{3688--3703}, \blue{2021}.

\bibitem{bonawitz2022federated}
Kallista Bonawitz, Peter Kairouz, Brendan Mcmahan, and Daniel Ramage.
\newblock Federated learning and privacy.
\newblock {\em Communications of the ACM}, 65(4):90--97, 2022.

\bibitem{fed_sec_aggregation}
Keith Bonawitz, Vladimir Ivanov, Ben Kreuter, Antonio Marcedone, H~Brendan McMahan, Sarvar Patel, Daniel Ramage, Aaron Segal, and Karn Seth.
\newblock Practical secure aggregation for privacy-preserving machine learning.
\newblock In {\em proceedings of the 2017 ACM SIGSAC Conference on Computer and Communications Security}, pages 1175--1191, 2017.

\bibitem{bonneel2015sliced}
N.~Bonneel, J.~Rabin, G.~Peyr{\'e}, and H.~Pfister.
\newblock Sliced and radon wasserstein barycenters of measures.
\newblock {\em Journal of Mathematical Imaging and Vision}, 51(1):22--45, 2015.

\bibitem{bottou2004large}
L.~Bottou and Y.~LeCun.
\newblock Large scale online learning.
\newblock {\em Advances in neural information processing systems}, 2004.

\bibitem{carmela}
S.~Boukoros, M.~Humbert, S.~Katzenbeisser, and C.~Troncoso.
\newblock On (the lack of) location privacy in crowdsourcing applications.
\newblock In {\em USENIX Security Symposium}, 2019.

\bibitem{winnerreport}
Y.~J. Bultitude and T.~Rautiainen.
\newblock {IST-4-027756 WINNER II D1. 1.2 V1. 2 WINNER II Channel Models}.
\newblock {\em EBITG, TUI, UOULU, CU/CRC, NOKIA, Tech. Rep.}, 2007.

\bibitem{specsense:17}
Ayon Chakraborty, Md~Shaifur Rahman, Himanshu Gupta, and Samir~R Das.
\newblock Specsense: Crowdsensing for efficient querying of spectrum occupancy.
\newblock In {\em IEEE INFOCOM 2017-IEEE Conference on Computer Communications}, pages 1--9. IEEE, 2017.

\bibitem{chen2020asynchronous}
Y.~Chen, Y.~Ning, M.~Slawski, and H.~Rangwala.
\newblock Asynchronous online federated learning for edge devices with non-iid data.
\newblock In {\em IEEE Big Data}, 2020.

\bibitem{chu2022securing}
Tianyue Chu, Alvaro Garcia-Recuero, Costas Iordanou, Georgios Smaragdakis, and Nikolaos Laoutaris.
\newblock Securing federated sensitive topic classification against poisoning attacks.
\newblock {\em arXiv preprint arXiv:2201.13086}, 2022.

\bibitem{damaskinos2020fleet}
G.~Damaskinos, R.~Guerraoui, A.-M. Kermarrec, V.~Nitu, R.~Patra, and F.~Ta{\"\i}ani.
\newblock Fleet: Online federated learning via staleness awareness and performance prediction.
\newblock In {\em International Middleware Conference}, 2020.

\bibitem{de2013unique}
Y.-A. De~Montjoye, C.~A Hidalgo, M.~Verleysen, and V.~D Blondel.
\newblock Unique in the crowd: The privacy bounds of human mobility.
\newblock {\em Scientific reports}, 3:1376, 2013.

\bibitem{inferring_activities}
M.~Diao, Y.~Zhu, J.~Ferreira, and C.~Ratti.
\newblock Inferring individual daily activities from mobile phone traces: A boston example.
\newblock {\em Environment and Planning B: Planning and Design}, 2015.

\bibitem{dwork2011differential}
C.~Dwork.
\newblock Differential privacy.
\newblock {\em Encyclopedia of Cryptography and Security}, pages 338--340, 2011.

\bibitem{elkordy2022much}
Ahmed~Roushdy Elkordy, Jiang Zhang, Yahya~H Ezzeldin, Konstantinos Psounis, and Salman Avestimehr.
\newblock How much privacy does federated learning with secure aggregation guarantee?
\newblock {\em arXiv preprint arXiv:2208.02304}, 2022.

\bibitem{raik:18}
R.~Enami, D.~Rajan, and J.~Camp.
\newblock {RAIK: Regional analysis with geodata and crowdsourcing to infer key performance indicators}.
\newblock In {\em Proc. of the IEEE WCNC}, April 2018.

\bibitem{fidaZipWeave:17}
M.~R. Fida, A.~Lutu, M.~K. Marina, and O.~Alay.
\newblock {ZipWeave: Towards efficient and reliable measurement based mobile coverage maps}.
\newblock In {\em IEEE INFOCOM '17}, 2017.

\bibitem{flamary2021pot}
R.~Flamary, N.~Courty, A.~Gramfort, M.~Z. Alaya, A.~Boisbunon, et~al.
\newblock Pot: Python optimal transport.
\newblock {\em JMLR}, 2021.

\bibitem{5gonapkpis}
J.~{Garcia-Reinoso} et~al.
\newblock The 5g eve multi-site experimental architecture and experimentation workflow.
\newblock In {\em IEEE 5G World Forum}, 2019.

\bibitem{geiping2020inverting}
J.~Geiping, H.~Bauermeister, H.~Dr\"{o}ge, and M.~Moeller.
\newblock Inverting gradients - how easy is it to break privacy in federated learning?
\newblock In {\em NeurIPS}, 2020.

\bibitem{heBCS:18}
S.~He and K.~G. Shin.
\newblock Steering crowdsourced signal map construction via bayesian compressive sensing.
\newblock In {\em IEEE INFOCOM}, 2018.

\bibitem{hitaj2017deep}
B.~Hitaj, G.~Ateniese, and F.~Perez-Cruz.
\newblock Deep models under the gan: information leakage from collaborative deep learning.
\newblock In {\em ACM SIGSAC}, 2017.

\bibitem{imran:14}
A.~Imran, A.~Zoha, and A.~Abu-Dayya.
\newblock {Challenges in 5G: {H}ow to empower SON with big data for enabling 5G}.
\newblock {\em IEEE network}, 2014.

\bibitem{isaacman2011identifying}
R.~Isaacman, S.and~Becker, R.~C{\'a}ceres, S.~Kobourov, M.~Martonosi, J.~Rowland, and A.~Varshavsky.
\newblock Identifying important places in people’s lives from cellular network data.
\newblock In {\em International Conference on Pervasive Computing}, 2011.

\bibitem{fl_survey}
P.~Kairouz, H~B. McMahan, B.~Avent, A.~Bellet, M.~Bennis, A.~N. Bhagoji, et~al.
\newblock Advances and open problems in federated learning.
\newblock {\em arXiv preprint arXiv:1912.04977}, 2019.

\bibitem{Krijestorac2021SpatialSS}
E.~Krijestorac, S.~S. Hanna, and D.~Cabric.
\newblock Spatial signal strength prediction using 3d maps and deep learning.
\newblock {\em IEEE International Conference on Communications}, 2021.

\bibitem{LAHA2018298}
A.~K. Laha and S.~Putatunda.
\newblock Real time location prediction with taxi-gps data streams.
\newblock {\em Transportation Research Part C: Emerging Technologies}, 2018.

\bibitem{hyperband_tuner}
L.~Li, K.~Jamieson, G.~DeSalvo, A.~Rostamizadeh, and A.~Talwalkar.
\newblock Hyperband: A novel bandit-based approach to hyperparameter optimization.
\newblock {\em JMLR}, 2018.

\bibitem{li2014efficient}
M.~Li, T.~Zhang, Y.~Chen, and A.~J Smola.
\newblock Efficient mini-batch training for stochastic optimization.
\newblock In {\em ACM SIGKDD}, 2014.

\bibitem{4221659}
N.~Li, T.~Li, and S.~Venkatasubramanian.
\newblock t-closeness: Privacy beyond k-anonymity and l-diversity.
\newblock In {\em IEEE 23rd International Conference on Data Engineering}, 2007.

\bibitem{li2020federated}
T.~Li, A.~K. Sahu, M.~Zaheer, M.~Sanjabi, A.~Talwalkar, and V.~Smith.
\newblock Federated optimization in heterogeneous networks.
\newblock {\em Proceedings of Machine Learning and Systems}, 2020.

\bibitem{liu2020loss}
C.~Liu, M.~Salzmann, T.~Lin, R.~Tomioka, and S.~S{\"u}sstrunk.
\newblock On the loss landscape of adversarial training: Identifying challenges and how to overcome them.
\newblock {\em arXiv preprint arXiv:2006.08403}, 2020.

\bibitem{liu2020fedvision}
Y.~Liu, A.~Huang, Y.~Luo, Y.~Huang, H.and~Liu, Y.~Chen, L.~Feng, T.~Chen, H.~Yu, and Q.~Yang.
\newblock Fedvision: An online visual object detection platform powered by federated learning.
\newblock In {\em AAAI Conference on Artificial Intelligence}, 2020.

\bibitem{8367709}
M.~Ma.
\newblock Enhancing privacy using location semantics in location based services.
\newblock In {\em IEEE ICBDA}, 2018.

\bibitem{original_federated}
H.~B. McMahan, E.~Moore, D.~Ramage, S.~Hampson, and B.~Aguera y~Arcas.
\newblock Communication-efficient learning of deep networks from decentralized data.
\newblock In {\em International Conference on Artificial Intelligence and Statistics}, 2017.

\bibitem{melis2019exploiting}
L.~Melis, C.~Song, E.~De~Cristofaro, and V.~Shmatikov.
\newblock Exploiting unintended feature leakage in collaborative learning.
\newblock In {\em IEEE Symposium on Security and Privacy (SP)}, 2019.

\bibitem{naseri2020toward}
M.~Naseri, J.~Hayes, and E.~De~Cristofaro.
\newblock Toward robustness and privacy in federated learning: Experimenting with local and central differential privacy.
\newblock {\em arXiv preprint arXiv:2009.03561}, 2020.

\bibitem{opensignal:11}
{Open Signal Inc.}
\newblock {Mobile Analytics and Insights}, June 2011.

\bibitem{phillips:12}
C.~Phillips, M.~Ton, D.~Sicker, and D.~Grunwald.
\newblock {Practical radio environment mapping with geostatistics}.
\newblock {\em Proc. of the IEEE DYSPAN '12}, pages 422--433, October 2012.

\bibitem{pyrgelis2017knock}
A.~Pyrgelis, C.~Troncoso, and E.~De Cristofaro.
\newblock Knock knock, who's there? membership inference on aggregate location data.
\newblock In {\em NDSS}, 2018.

\bibitem{ray:16}
A.~Ray, S.~Deb, and P.~Monogioudis.
\newblock Localization of lte measurement records with missing information.
\newblock In {\em IEEE INFOCOM}, 2016.

\bibitem{dropout_paper}
N.~Srivastava, G.~Hinton, A.~Krizhevsky, I.~Sutskever, and R.~Salakhutdinov.
\newblock Dropout: A simple way to prevent neural networks from overfitting.
\newblock {\em JMLR}, 2014.

\bibitem{truex2020ldp}
Stacey Truex, Ling Liu, Ka-Ho Chow, Mehmet~Emre Gursoy, and Wenqi Wei.
\newblock Ldp-fed: Federated learning with local differential privacy.
\newblock In {\em Proceedings of the Third ACM International Workshop on Edge Systems, Analytics and Networking}, pages 61--66, 2020.

\bibitem{infocomLSTM:17}
J.~{Wang}, J.~{Tang}, Z.~{Xu}, Y.~{Wang}, G.~{Xue}, X.~{Zhang}, and D.~{Yang}.
\newblock Spatiotemporal modeling and prediction in cellular networks: A big data enabled deep learning approach.
\newblock In {\em IEEE INFOCOM '17}, pages 1--9, May 2017.

\bibitem{wei2020federated}
Kang Wei, Jun Li, Ming Ding, Chuan Ma, Howard~H Yang, Farhad Farokhi, Shi Jin, Tony~QS Quek, and H~Vincent Poor.
\newblock Federated learning with differential privacy: Algorithms and performance analysis.
\newblock {\em IEEE Transactions on Information Forensics and Security}, 15:3454--3469, 2020.

\bibitem{yang:10}
J.~Yang, A.~Varshavsky, H.~Liu, Y.~Chen, and M.~Gruteser.
\newblock Accuracy characterization of cell tower localization.
\newblock In {\em Proc. of the ACM UbiComp '10}, pages 223--226, 2010.

\bibitem{yue2022gradient}
Kai Yue, Richeng Jin, Chau-Wai Wong, Dror Baron, and Huaiyu Dai.
\newblock Gradient obfuscation gives a false sense of security in federated learning.
\newblock {\em arXiv preprint arXiv:2206.04055}, 2022.

\bibitem{raytracing:15}
Z.~Yun and M.~F. Iskander.
\newblock Ray tracing for radio propagation modeling: Principles and applications.
\newblock {\em IEEE Access}, 2015.

\bibitem{milanMobihocDeepnets:18}
C.~Zhang and P.~Patras.
\newblock Long-term mobile traffic forecasting using deep spatio-temporal neural networks.
\newblock In {\em ACM MobiHoc}, 2018.

\bibitem{idlg}
B.~Zhao, K.~R. Mopuri, and H.~Bilen.
\newblock idlg: Improved deep leakage from gradients.
\newblock {\em arXiv preprint:2001.02610}, 2020.

\bibitem{dlg}
L.~Zhu, Z.~Liu, and S.~Han.
\newblock Deep leakage from gradients.
\newblock In {\em NeurIPS}, 2019.

\bibitem{zong2023privacy}
Zixiao Zong, Mengwei Yang, Justin Ley, Carter~T Butts, and Athina Markopoulou.
\newblock Privacy by projection: Federated population density estimation by projecting on random features.
\newblock {\em Proceedings on Privacy Enhancing Technologies}, 1:309--324, 2023.

\end{thebibliography}

\vfill
\newpage

\clearpage
\begin{appendix}
\section{}\label{sec:appendix}

\subsection{Proofs}
In this part of the appendix, we provide the proofs of the statements made in Section \ref{sec:analysis}. %

\subsubsection{Proof of Lemma 1}\label{sec:appendix_proofs}
\label{proof-lemma1}

\begin{restatable}{assumption}{assumptionone}
The service model $y_i=F(x_i, w)$ starts with a biased fully-connected layer, which is defined as $y_i^{1}=\sigma(w^1x_i+b^1),$ where $y_i^{1}\in \mathbb{R}^H$ is the output of the first layer with dimension $H$, and $w^1\in \mathbb{R}^{H\times 2}$ and $b^1\in \mathbb{R}^{H}$ represent the weight matrix and bias vector in the first layer. 
\label{assumption1}
\end{restatable}

\begin{restatable}{assumption}{assumptiontwo}
Given a data mini-batch of size $B$: $\{(x_i,y_i)\}_{i=1}^{i=B}$, $\exists~h\in\{1,...,H\}$, $\frac{1}{B}\sum_{i=1}^{i=B}\frac{\partial \ell(F(x_i,w),y_i)}{\partial b^1_h}\neq 0,$ where $b^1_h$ is the $h$-th element in the bias vector of the first layer (\ie $b^1=[b^1_1,...,b^1_H]^T$). 
\label{assumption2}
\end{restatable}

The first assumption states that the DNN model $y_i=F(x_i, w)$ starts with a biased fully-connected layer, which is defined as $y_i^{1}=\sigma(w^1x_i+b^1),$ where $y_i^{1}\in \mathbb{R}^H$ is the output of the first layer with dimension $H$, and $w^1\in \mathbb{R}^{H\times 2}$ and $b^1\in \mathbb{R}^{H}$ represent the weight matrix and bias vector in the first layer. 
While the first layer in our DNN is a biased fully-connected layer with $H=224$, there is a dropout layer which nullifies some elements (5\% of them) thus breaks the fully-connected assumption. However, in the absence of a dropout layer the DLG attacker would perform better, and thus we are being conservative by analyzing a slightly stronger attacker. 

The second assumption ensures that the gradient w.r.t the bias vector in the first layer is a non-zero vector. This typically holds during training when the model parameters have not converged. Formally,
Given a data mini-batch of size $B$: $\{(x_i,y_i)\}_{i=1}^{i=B}$, $\exists~h\in\{1,...,H\}$, $\frac{1}{B}\sum_{i=1}^{i=B}\frac{\partial \ell(F(x_i,w),y_i)}{\partial b^1_h}\neq 0,$ where $b^1_h$ is the $h$-th element in the bias vector of the first layer (\ie $b^1=[b^1_1,...,b^1_H]^T$). 

We are now ready to prove Lemma \ref{lemma1} in Sec. \ref{sec:analysis} which holds under the two assumptions stated above.

\noindent \textbf{\textit{Proof of Lemma \ref{lemma1}:}}
Under Assumption \ref{assumption1}, the first layer of the service model starts with a biased fully-connected layer, defined as $y_i^{1}=\sigma(w^1x_i+b^1)$. Based on the chain rule, we can calculate the following partial derivatives:
\begin{equation}
    \frac{\partial \ell(F(x_i,w),y_i)}{\partial w^1_k}=\frac{\partial \ell(F(x_i,w),y_i)}{\partial y^1_{i,k}}\cdot\sigma'(w^1_kx_i+b^1_h)\cdot x_i^T,
    \label{proof-eq1}
\end{equation}
\begin{equation}
    \frac{\partial \ell(F(x_i,w),y_i)}{\partial b^1_h}=\frac{\partial \ell(F(x_i,w),y_i)}{\partial y^1_{i,k}}\cdot\sigma'(w^1_kx_i+b^1_h),
    \label{proof-eq2}
\end{equation}
where $w^1_k$ is the $k$-th row of weight matrix $w^1$ (note that $w^1\in\mathbb{R}^{K\times2}$).

\noindent Combing Eq. (\ref{proof-eq1}) and Eq. (\ref{proof-eq2}), we can derive:
\begin{equation}
    \frac{\partial \ell(F(x_i,w),y_i)}{\partial w^1_k}=\frac{\partial \ell(F(x_i,w),y_i)}{\partial b^1_h}\cdot x_i^T.
    \label{proof-eq3}
\end{equation}
Note the above steps are borrowed from the proof of Proposition D3.1 in \cite{geiping2020inverting}. The following steps are novel and specific to our application:

Based on Eq. (\ref{proof-eq3}) and the definition of DLG attack, the reconstructed user location of DLG attacker $x_{DLG}$ satisfies:
\begin{equation}
\frac{1}{B}\sum_{i=1}^{i=B}\frac{\partial \ell(F(x_i,w),y_i)}{\partial w^1_k}=\frac{1}{B}\sum_{i=1}^{i=B}\frac{\partial \ell(F(x_i,w),y_i)}{\partial b^1_h}\cdot x_{DLG}^T.
\label{proof-eq4}
\end{equation}
By combining Eq. (\ref{proof-eq3}) and Eq. (\ref{proof-eq4}), we can further derive:
\begin{equation}
\frac{1}{B}\sum_{i=1}^{i=B}g_i(w)\cdot x_i^T=\frac{1}{B}\sum_{i=1}^{i=B}g_i(w)\cdot x_{DLG}^T,
\label{proof-eq5}
\end{equation}
where $g_i(w)=\frac{\partial \ell(F(x_i,w),y_i)}{\partial b^1_h}$. Therefore, under Assumption \ref{assumption2}, we can get:
\begin{equation}
x_{DLG}=\frac{1}{B}\sum_{i=1}^{i=B}\frac{g_i(w)}{\bar g(w)}\cdot x_i,
\label{proof-eq6}
\end{equation}
where $\bar g(w) = \frac{1}{B}\sum_{i=1}^{i=B}g_i(w).$ $\square$

\subsubsection{Proof of Theorem 1}
\label{proof-theorem1}
Subject to Assumption \ref{assumption1} and \ref{assumption2}, we are ready to prove Theorem \ref{theorem1}  (see Sec. \ref{sec:analysis}).

\noindent \textbf{\textit{Proof of Theorem \ref{theorem1}:}}
The distance between the reconstructed user location and the center of user locations in the mini-batch can be bounded as follows:
\begin{equation}
\begin{split}
    &||x_{DLG}- \bar x||_2 =||x_{DLG}- \frac{1}{B}\sum_{i=1}^{i=B}\frac{g_i(w)}{\bar g(w)}\cdot x_i||_2 \\
    & = ||\frac{1}{B}\sum_{i=1}^{i=B}\frac{g_i(w) - \bar g(w)}{\bar g(w)}\cdot x_{i}||_2\\
    & = \frac{1}{B|\bar{g}(w)|}||\sum_{i=1}^{i=B}\big(g_i(w) - \bar g(w)\big)\cdot\big(x_{i}-\bar x\big)||_2\\
    &\leq \frac{1}{B|\bar{g}(w)|}\sum_{i=1}^{i=B}||\big(g_i(w) - \bar g(w)\big)\cdot\big(x_{i}-\bar x\big)||_2 \\
    &\leq \frac{1}{2B|\bar{g}(w)|}\sum_{i=1}^{i=B}\Big(\big(g_i(w) - \bar g(w)\big)^2 + ||x_{i}-\bar x||_2^2\Big).\quad
\end{split}
\label{proof-eq7}
\end{equation}

\subsubsection{Proof of Theorem 2}
\label{proof-theorem2}

\begin{restatable}{assumption}{assumptionthree}
The gradient function $\ell(F(x_i,w),y_i)$ is Lipschitz continuous with respect to $(x_i, y_i)$, that is:
\begin{equation}
\begin{split}
    &||\nabla \ell(F(x_i,w),y_i)-\nabla \ell(F(x_i,w),y_i)||_2\\
    &\leq L\sqrt{||x_i-x_j||_2^2 + ||y_i-y_j||_2^2}, (L>0).
\end{split}
\end{equation}
\label{assumption3}
\end{restatable}
Note that while the ReLU activation function violates this assumption, any smooth approximations would satisfy it \cite{liu2020loss}. Subject to Assumption \ref{assumption1}, \ref{assumption2} and \ref{assumption3}, we are ready to prove Theorem \ref{theorem2} (see Sec. \ref{sec:analysis}).

\noindent \textbf{\textit{Proof of Theorem \ref{theorem2}:}} 
Based on Assumption \ref{assumption3}, we can derive that $\forall (x_i,y_i), (x_j,y_j)$, the following inequality holds:
\begin{equation}
\begin{split}
   \big(g_i(w)-g_j(w)\big)^2 \leq L^2\big(||x_i-x_j||_2^2 + ||y_i-y_j||_2^2\big).
\end{split}
\label{proof-eq8}
\end{equation}
Since:
\begin{equation}
\begin{split}
    &\big(g_i(w)-\bar g(w)\big)^2=\frac{1}{B^2}\Big(\sum_{j=1}^{j=B}\big(g_i(w)- g_j(w)\big)\Big)^2\\
    &\leq\frac{1}{B}\sum_{j=1}^{j=B}\big(g_i(w)-g_j(w)\big)^2 \\ 
    &\leq\frac{L^2}{B}\sum_{j=1}^{j=B}\big(||x_i-x_j||_2^2 + ||y_i-y_j||_2^2\big) \\
    &=L^2\big(||x_i||_2^2+y_i^2-2x_i^T\bar x-2y_i\bar y\big)\\
    &+\frac{L^2}{B}\sum_{j=1}^{j=B}\big(||x_j||_2^2 + y_j^2\big),
\end{split}
\label{proof-eq9}
\end{equation}
we can prove that:
\begin{equation}
\begin{split}
    &\frac{1}{B}\sum_{i=1}^{i=B}\big(g_i(w)-\bar g(w)\big)^2\\
    &\leq\frac{L^2}{B}\sum_{i=1}^{i=B}\big(||x_i||_2^2+y_i^2-2x_i^T\bar x-2y_i\bar y\big)\\
    &+\frac{L^2}{B}\sum_{j=1}^{j=B}\big(||x_j||_2^2 + y_j^2\big)\\
    &=\frac{2L^2}{B}\sum_{i=1}^{i=B}\big(||x_i||_2^2+y_i^2-x_i^T\bar x-y_i\bar y\big)\\
    &=\frac{2L^2}{B}\sum_{i=1}^{i=B}\big(||x_{i}-\bar x||^2+||y_{i}-\bar y||^2\big)
\end{split}
\label{proof-eq10}
\end{equation}
Finally, combining Eq. (\ref{proof-eq8}) and Eq. (\ref{proof-eq10}), we can derive that:
\begin{equation}
\begin{split}
    &||x_{DLG}-\bar x_{i}||_2 \\
    &\leq \frac{L^2}{B|\bar{g}(w)|}\sum_{i=1}^{i=B}\big(\frac{2L^2+1}{2L^2}||x_{i}-\bar x||^2+||y_{i}-\bar y||^2\big).
\end{split}
\label{proof-eq11}
\end{equation}

\begin{figure*}[!ht]
	\centering
	\begin{subfigure}[t]{0.3\textwidth}
		\centering
		\includegraphics[width=0.99\textwidth]{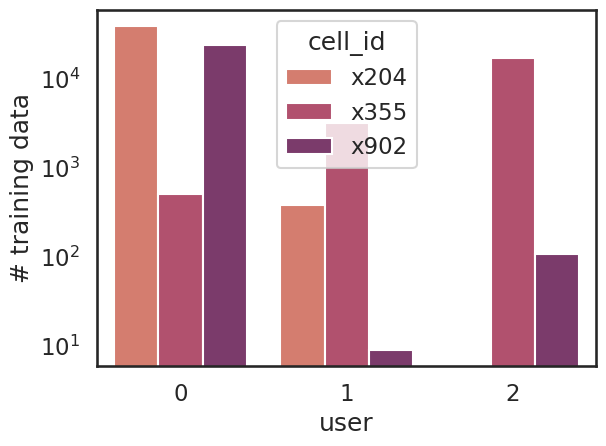}
		\caption{Number of datapoints per user, for the top-3 cell towers.}
		\label{fig:distribution_statistics_per_cell}
	\end{subfigure}
	\begin{subfigure}[t]{0.3\textwidth}
		\centering
		\includegraphics[width=0.99\textwidth]{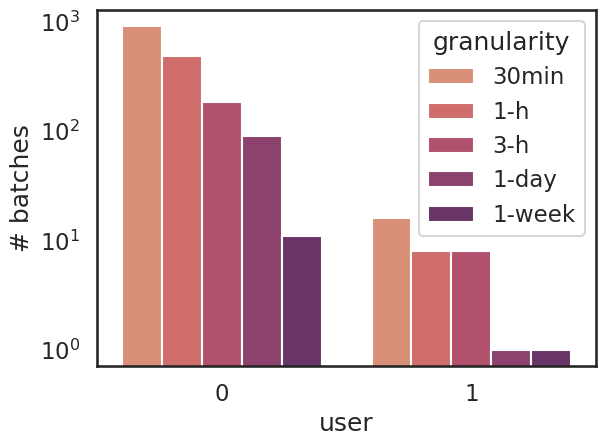}
		\caption{Total rounds  per user: x204 cell.}
		\label{fig:batches_per_interval_204}
	\end{subfigure}
	\begin{subfigure}[t]{0.3\textwidth}
		\centering
		\includegraphics[width=0.99\textwidth]{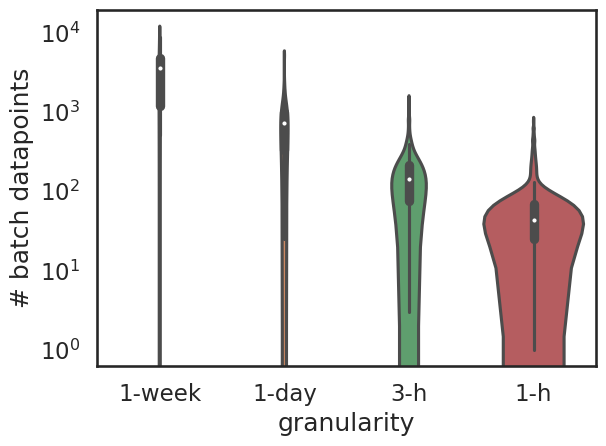}
		\caption{\blue{Total datapoints per round:} x204 cell, user 0.}
		\label{fig:datapoints_per_batch_per_interval_204}
	\end{subfigure}
	\caption{\blue{More statistics for \campus. The data is split into intervals/rounds of duration $T$; the longer $T$ is, the fewer rounds/batches are obtained but each round has more datapoints.}}
	\label{fig:uci_stats_top_cells}
\end{figure*}

\subsection{Datasets}\label{sec:appendix_data}
This section of the appendix extends Sec. \ref{sec:dataset}.

\begin{figure*}[t!]
	\centering
	\begin{subfigure}{0.3\linewidth}
		\centering
    \includegraphics[width=0.99\linewidth]{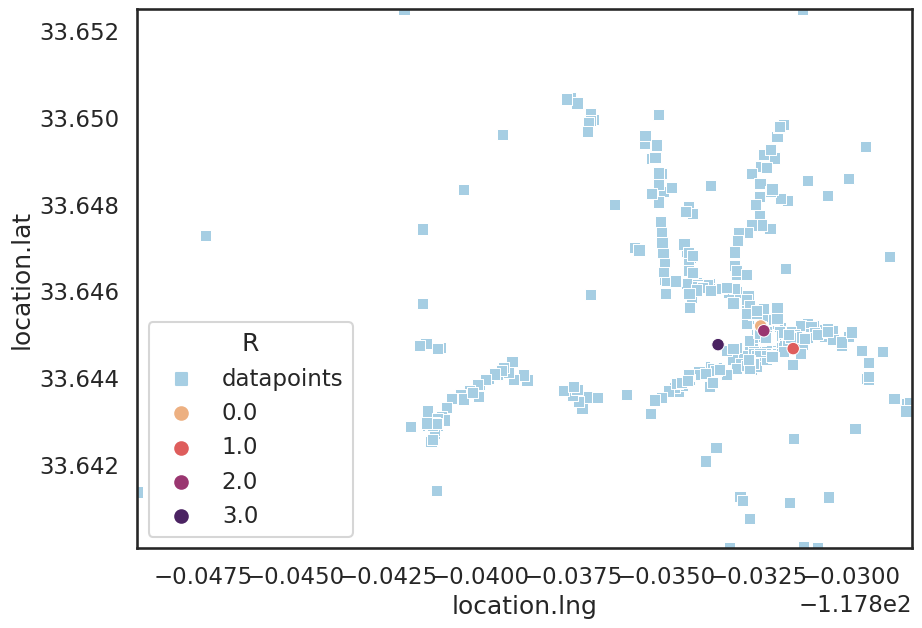}
		\caption{$T=$ 1-week: EMD=8.1.}
		\label{fig:1w_strongest_attack_points_355}
    \end{subfigure}
    	\begin{subfigure}{0.3\linewidth}
		\centering
    \includegraphics[width=0.99\linewidth]{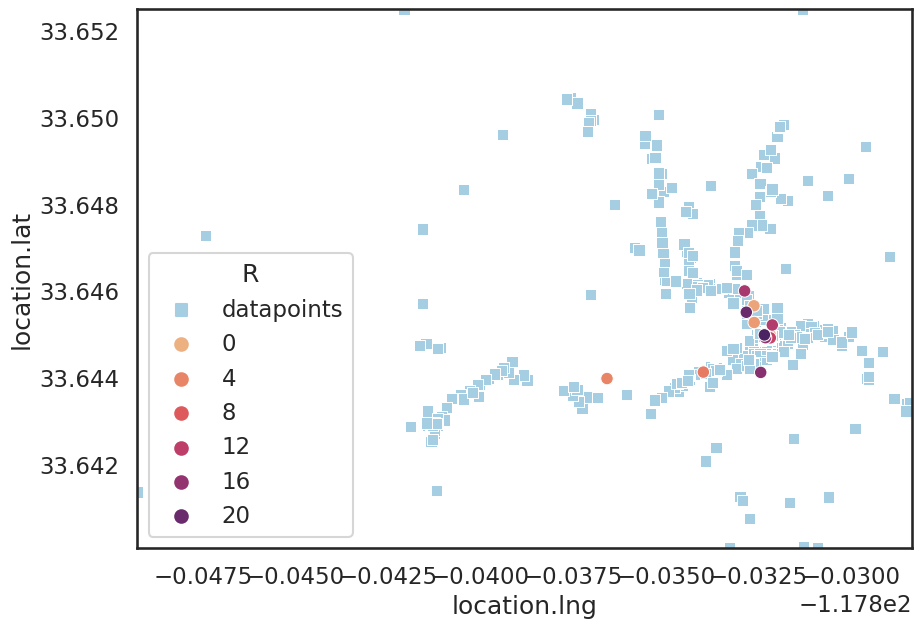}
		\caption{$T=$ 24-hour: EMD=7.4.}
		\label{fig:24h_strongest_attack_DLG_iters_355}
    \end{subfigure}
\begin{subfigure}{0.3\linewidth}
		\centering
    \includegraphics[width=0.99\linewidth]{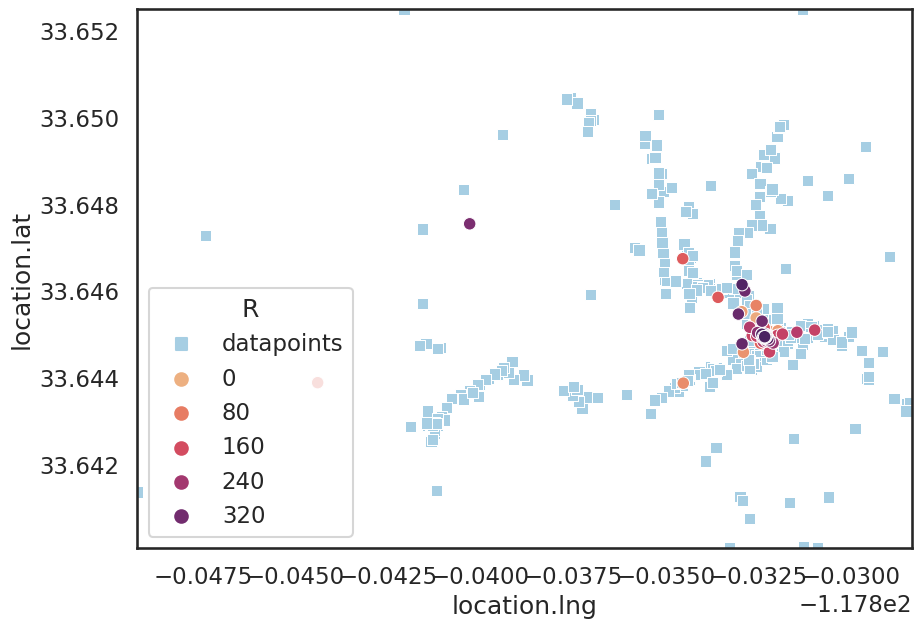}
		\caption{$T=$ 1-hour: EMD=6.9.}
		\label{fig:1h_strongest_attack_points_355}
    \end{subfigure}
\caption{\small \campus cell x355: The real vs. reconstructed locations with FedSGD with $\eta=0.001$ for various $T$ intervals.} %
\label{fig:dlg_iters_cell355_strongest_attack_points}
\end{figure*}

\begin{figure*}[t!]
	\centering
	\begin{subfigure}{0.3\linewidth}
		\centering
    \includegraphics[width=0.99\linewidth]{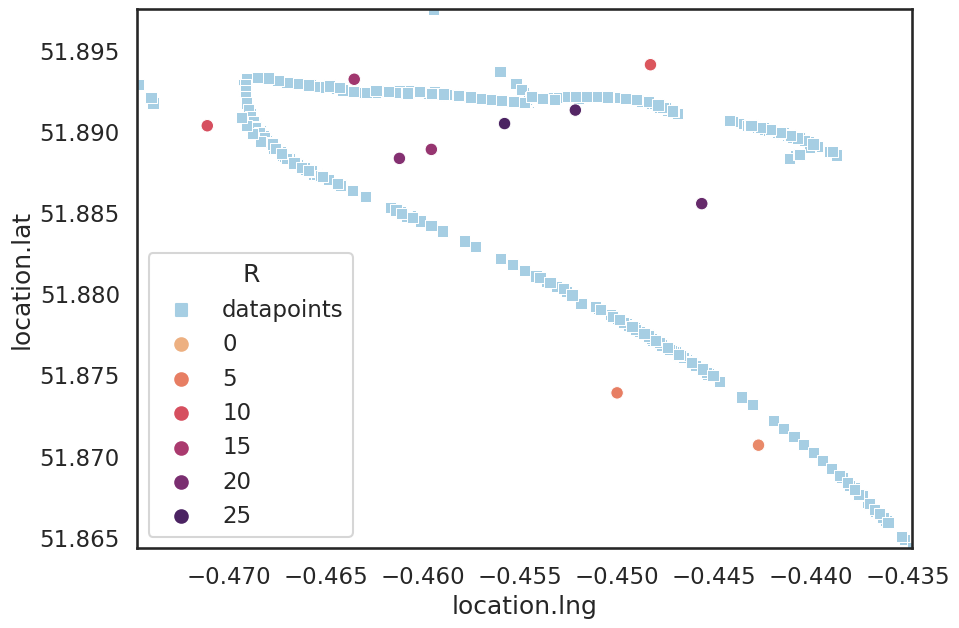}
		\caption{\small B=inf, E=1: EMD=13.47.}
		\label{fig:radio_target3_strongest}
    \end{subfigure}
    	\begin{subfigure}{0.3\linewidth}
		\centering
    \includegraphics[width=0.99\linewidth]{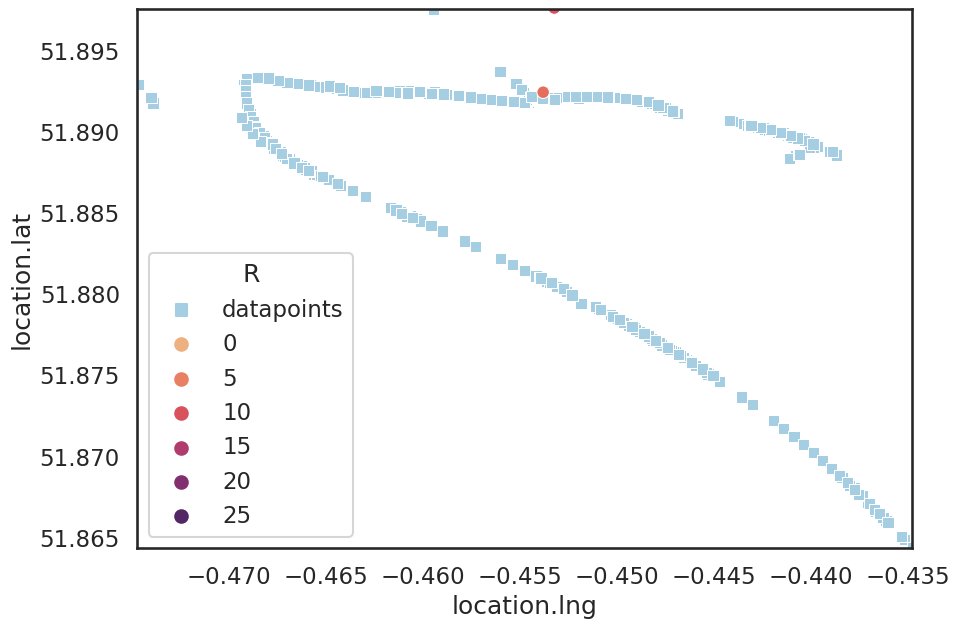}
		\caption{\small B=10, E=5: EMD=21.03.}
		\label{fig:radio_target3_averaging}
    \end{subfigure}
\begin{subfigure}{0.3\linewidth}
		\centering
    \includegraphics[width=0.99\linewidth]{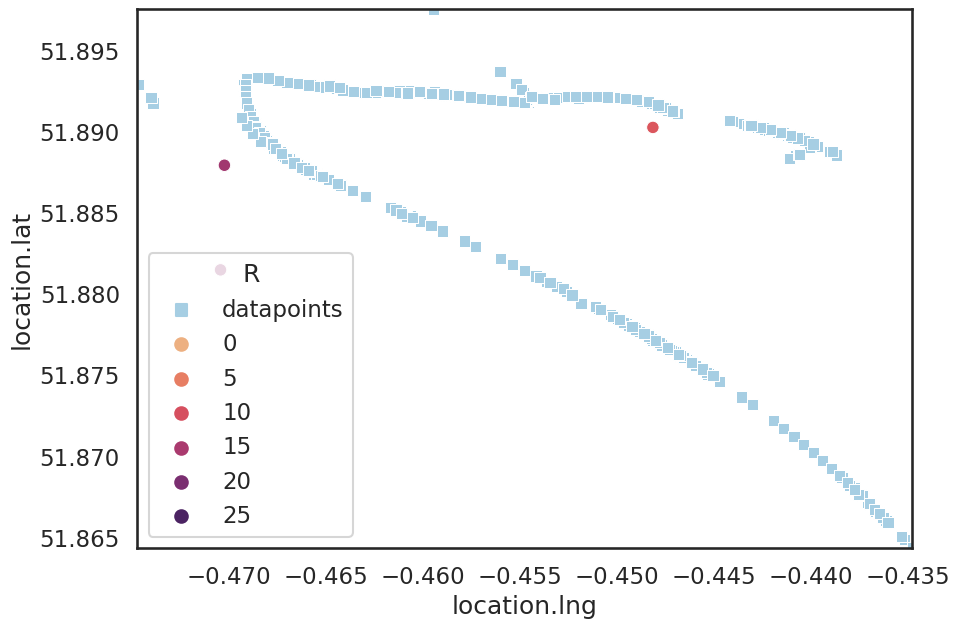}
		\caption{\small multi-user B=10, E=5: EMD=22.515.}
		\label{fig:radio_target3_multiuser_w_user1}
    \end{subfigure}
\caption{\small \radiocells for target user 3 for $T=$1-week. %
The RMSE is evaluated on global test set when multiple users. The avg random EMD is 34 (0.5). Multi-users correspond to when background ``user'' 0 (orange trajectory) participates in global updates.}
\label{fig:radio_cell_x455_target_3}
\end{figure*}

\begin{figure}[!h]
	\centering
    \includegraphics[width=0.6\linewidth]{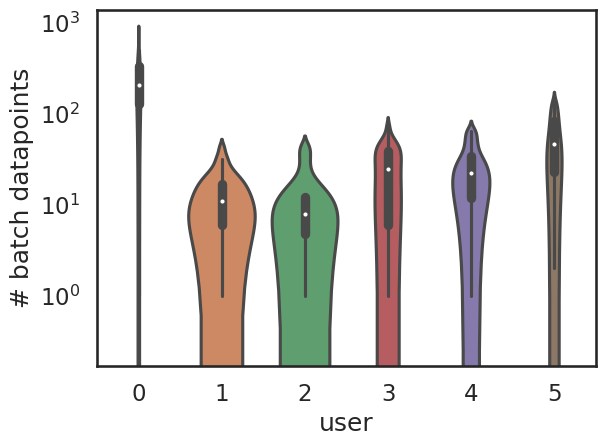}
	\caption{\radiocells: datapoints per batch per user. }%
	\label{fig:datapoints_per_batch_per_interval_x455_radiocells}
	
\end{figure}

\begin{figure}[!h]
    \centering
\includegraphics[width=0.6\linewidth]{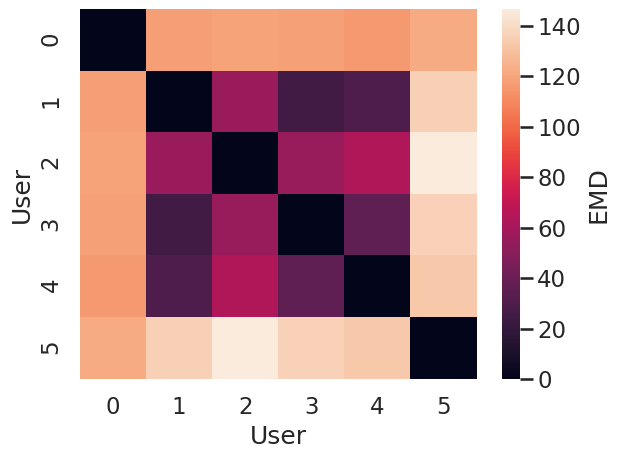}
    \caption{\small \textbf{User similarity via EMD;} darker color corresponds to higher similarity (low EMD), lighter means dissimilar (high EMD). }
    \label{fig:radio_cell_x455_heatmap}
    
\end{figure}

\textbf{\campus.}  \blue{Fig. \ref{fig:distribution_statistics_per_cell} shows the number of measurements for the top three cell towers, per user in each cell tower. Fig. \ref{fig:batches_per_interval_204} and Fig. \ref{fig:datapoints_per_batch_per_interval_204} show the number of batches per user and the distribution of number of datapoints per round for each granularity of duration $T$ for the x204 cell, respectively. For longer $T$, we obtain fewer batches but they contain significantly more datapoints, \eg for cell x204 and user 0, the average batch size for 1-week batches is 3492, for 1-day is 817, for 3-hour is 205 and for 1-hour batches is 79. }

\textbf{\radiocells.} Fig. \ref{fig:datapoints_per_batch_per_interval_x455_radiocells} shows the distribution of datapoints per batch for each user for $T=$ 1-week. It it significantly sparser than the \campus; the average batch size per user is 50. Fig. \ref{fig:radio_cell_x455_heatmap} shows the pair-wise user similarity based on the EMD value between the distribution of locations between each pair of users. %

\subsection{Considering All Available Data\label{appendix:alldata}}

\blue{Throughout the main paper, we define and evaluate our "online FL" framework, 
where the user trains locally on the data that becomes available in the current round and does not use local data from previous rounds neither training nor testing.
In this section, we provide quantitative evidence, that catastrophic forgetting is indeed not a problem in our setting. To that end, we consider all past data for training (which we refer to as ``cumulative online FL'') and then we consider past data for testing. The consistency in performance across all evaluation scenarios indicates that there is indeed no catastrophic forgetting. The intuition is that because of the predictable and repeated patterns in human mobility, similar data arrive in an online fashion over days and weeks. }

\begin{figure*}[!h]
	\centering
	\begin{subfigure}{0.33\linewidth}
		\centering
    \includegraphics[width=0.99\linewidth]{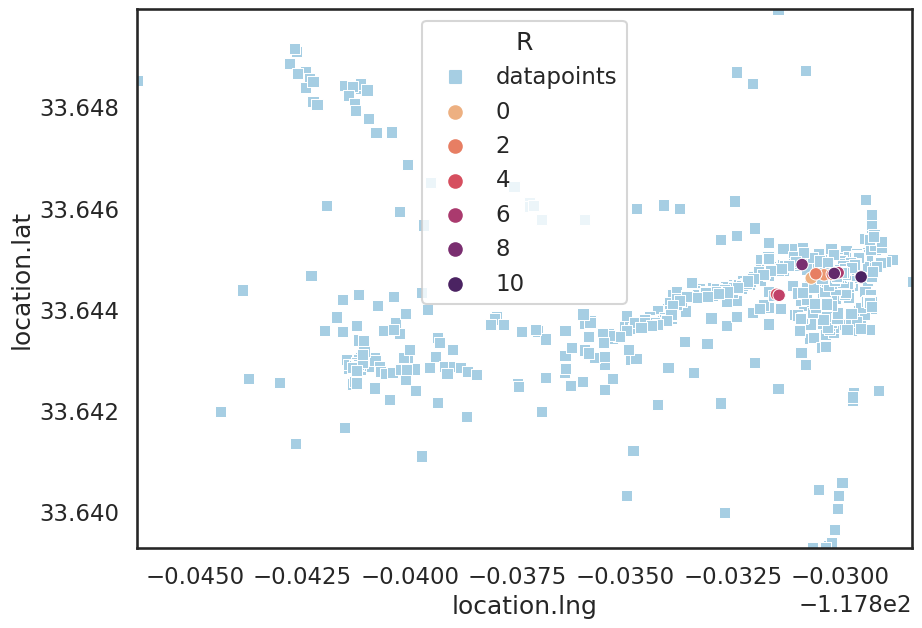}
		\caption{{Online FL, without averaging: B=inf, E=1. Privacy Loss: EMD=7.6 (0.17). Model Performance: RMSE=4.93.}}
		\label{fig:cumulativeFL1}
    \end{subfigure}
    \begin{subfigure}{0.33\linewidth}
		\centering
    \includegraphics[width=0.99\linewidth]{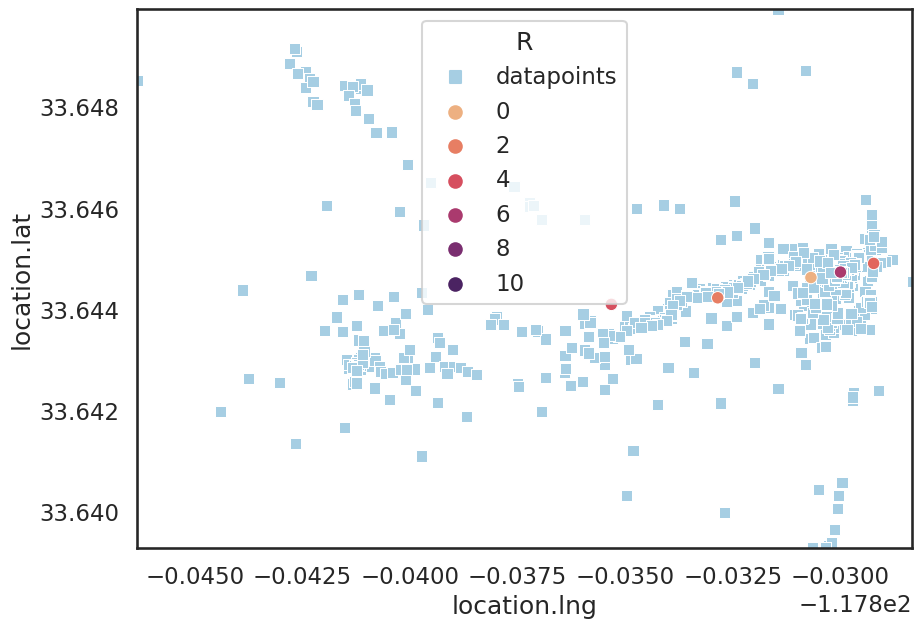}
		\caption{{Online FL with averaging: B=20, E=5. Privacy Loss: EMD=9.7 (0.215). Model Performance: RMSE=4.83.}}
	\label{fig:cumulativeFL2}
    \end{subfigure}
\begin{subfigure}{0.33\linewidth}
		\centering
    \includegraphics[width=0.99\linewidth]{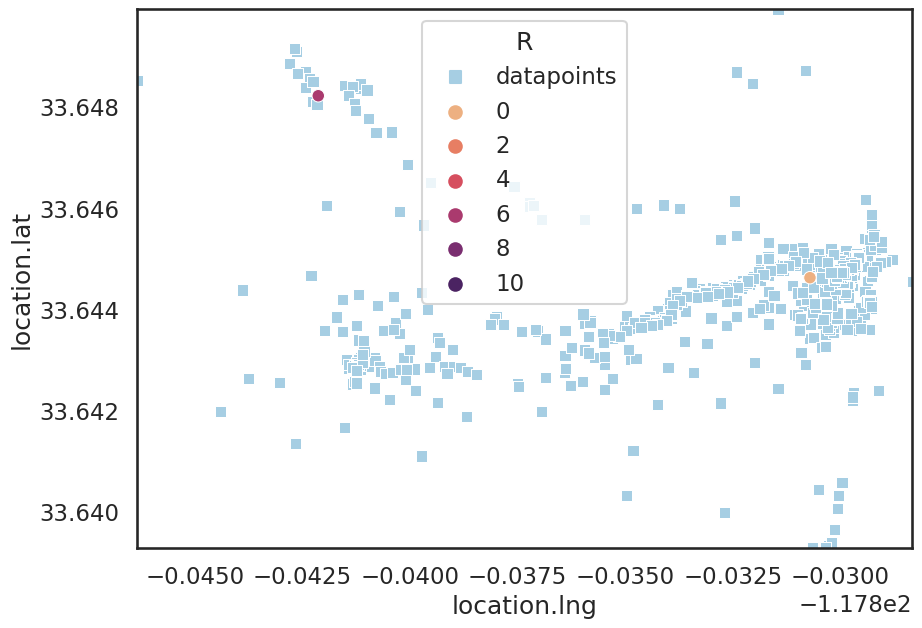}
		\caption{{Cumulative online FL with B=20, E=5. Privacy Loss: EMD=19.2 (0.425). Model PErformance: RMSE=4.85.}}
	\label{fig:cumulativeFL3}
    \end{subfigure}
\caption{\blue{\small {\bf The effect of discarding training data.} Online FL with 1-week rounds and  two sets of parameters (a) w/o averaging, \ie B=inf, E=1 and (b) with averaging, \ie B=20, E=5; compared to Cumulative Online FL (which does not discard any past training data). Without any averaging, even for coarser intervals the attacker can reconstruct accurately the target's most frequent locations which correspond to home/work locations. Adding averaging results in more divergence (90\%) and the converged locations are farther from the true locations. Cumulative online FL has increased EMD, although it still recovers the oversampled location (home/office) in the first round, it also requires storage of all local data from all rounds and increased local training time.}}
\label{fig:cumulativeFL}
\vspace{-10pt}
\end{figure*}

\blue{\textbf{Cumulative Online FL.} In Sec. \ref{sec:results}, we consider online FL where the user trains locally on the data that becomes available in the current round and discarding the local data from previous rounds. %
Here, we consider the other extreme, namely ``Cumulative Online FL'' where the previously seen data are not discarded but they are re-used in addition to the newly acquired batch of data in each round. Fig. \ref{fig:cumulativeFL} show the reconstructed points in the case of the strongest attack, with no averaging (B=inf, E=1), averaging (B=20, E=5) and cumulative online FL, As shown from evaluation results, the cumulative online FL (Figure \ref{fig:cumulativeFL3}) increases EMD to 19.2 without negatively affecting the RMSE compared to “online FL” (Figure \ref{fig:cumulativeFL1} and Figure \ref{fig:cumulativeFL2}). Training on all available data provides some level of privacy protection from location leakage (although the important locations of a user are still revealed), at the cost of increased training time and storage on the device. Ultimately, how much memory to keep in training data depends on those practical constraints and the data itself (in our case, human mobility patterns lead to data repeat over days and weeks, so it is acceptable to forget.).
In the \campus, the user mobility patterns are very similar during different time intervals. For the above reasons, we adopt online FL in our paper.}

\begin{figure}[!h]
\begin{center}
    \begin{subfigure}{.37\textwidth} %
        \centering
        \includegraphics[width=.99\linewidth]{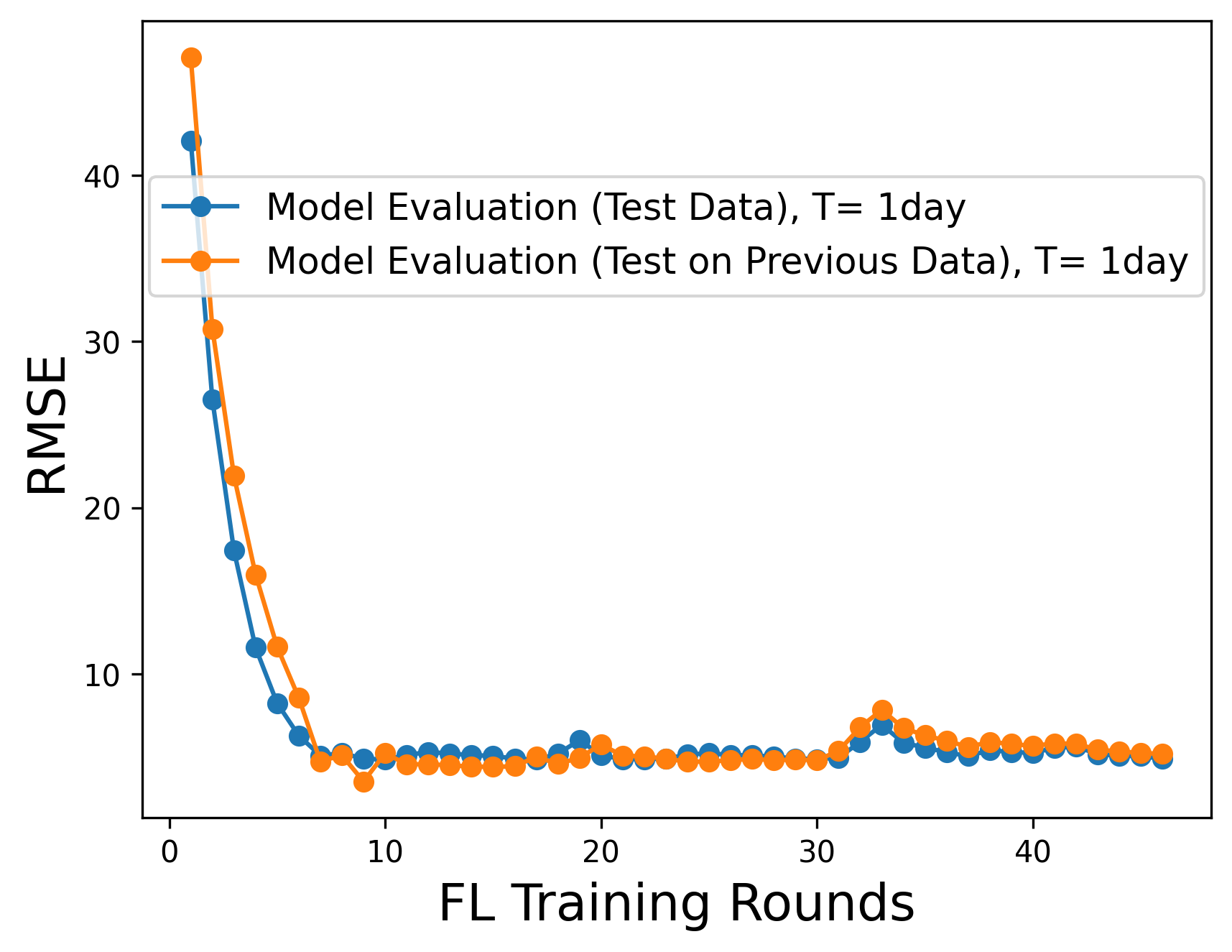}
        \caption{{Model evaluation with 1 day interval.}}
        \label{fig:catastrophic_forgetting1}
    \end{subfigure}
    \begin{subfigure}{.37\textwidth} %
        \centering
        \includegraphics[width=.99\linewidth]{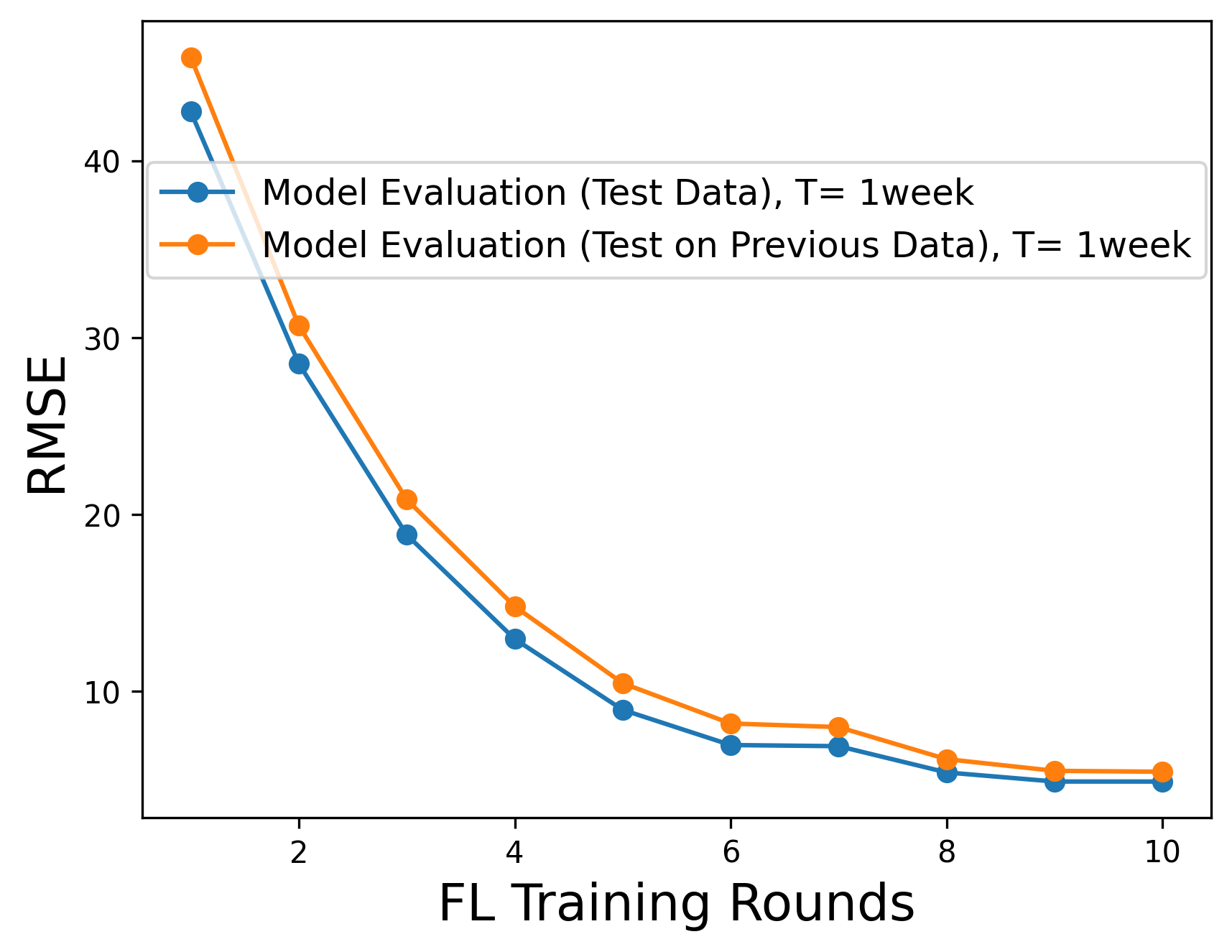}
        \caption{Model evaluation with 1 week interval.}
        \label{fig:catastrophic_forgetting2}
    \end{subfigure}
    \caption{\blue{{\bf Assessing (the potential of) Catastrophic Forgetting.} We evaluate the model performance considering (i) Test Data as in the main paper vs. Test on Previous Data (\ie from all previous rounds) and (ii) the effect of testing window, considering rounds with duration $T$, 1 day and 1 week. The fact that there is no difference in the model performance indicates that there is no catastrophic forgetting. The underlying reason is that human mobility patterns repeat themselves over days and weeks.}}
\label{fig:catastrophic_forgetting}
\vspace{-10pt}
\end{center}
\end{figure}

\blue{\textbf{Catastrophic Forgetting.} To address the potential concern of catastrophic forgetting, we have evaluated our model using the previous data as a test set and investigated the effect of the testing window. The results of this evaluation are presented in Figure \ref{fig:catastrophic_forgetting}, where we compare Model Evaluation (Test Data) and Model Evaluation (Test on Previous Data) for both time intervals, 1 day and 1 week. Model Evaluation (Test Data) means we evaluate the training model using test dataset as in the main paper. Model Evaluation (Test on Previous Data) means we evaluate the training model using the data from all previous rounds. According to our evaluation results, it is evident that Model Evaluation (Test Data) and Model Evaluation (Test on Previous Data) exhibit very similar performance and curve trends. This consistency in performance across the two evaluation scenarios strongly indicates that our current training model does not suffer from catastrophic forgetting.}

\subsection{Tuning Parameters}\label{sec:additional_results}
This section of the appendix extends Sec. \ref{sec:results} by providing additional results that could not be included due to lack of space: 

\blue{\textbf{Tuning the DNN.} In Sec. \ref{sec:results}, the tuned DNN architecture is  optimized for the datasets. %
For example, when tuned for \campus cell x204, the DNN consists of two hidden layers with 224 and 640 units and ReLU and sigmoid activation functions, respectively, and dropout layers between the hidden layers with 5\% of units being dropped randomly. The dropout layers are necessary to prevent overfitting in over-parameterized DNN architectures \cite{dropout_paper}.} %

\textbf{\campus cell x355.} Fig. \ref{fig:dlg_iters_cell355_strongest_attack_points} demonstrates the DLG attack for another cell tower (x355). %

\begin{figure}[t!]
	\centering
        \includegraphics[width=0.5\linewidth]{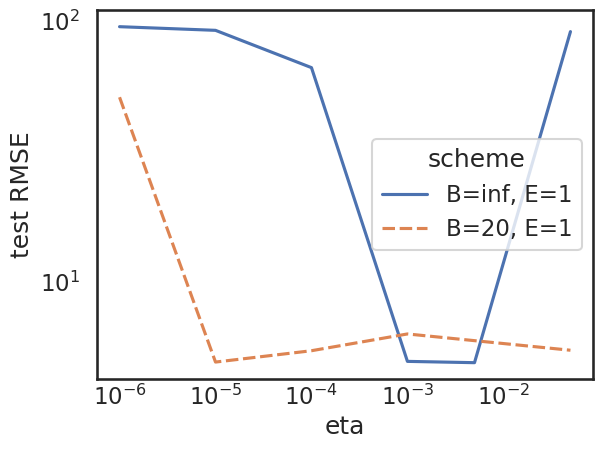}
\caption{\small \textbf{Tuning learning rate.} We use \campus with $T=$1-week to tune $\eta$ for B=inf and B=20 and show how utility (RMSE) and privacy (EMD) is affected. We choose $\eta=0.001$ as our default $\eta$ (unless stated otherwise) and $\eta=10^{-5}$ for B=20, since both minimize RMSE and EMD (not shown).}
 \vspace{-5pt}
\label{fig:eta_tuning_1_week}
\end{figure}

\textbf{Tuning learning rates $\eta$ for FedSGD.}
We performed a grid search with $\eta=10^{-6}, 10^{-5}, 10^{-4}, 10^{-3}, 0.001, 0.1$ for two scenarios: FedSGD and FedAvg with E=1, B=20.  To minimize RMSE, we choose $\eta=0.001$ as our default learning rate. In case of mini-batches, we observe a lower $\eta=10^{-5}$ can slightly lower RMSE. Fig. \ref{fig:eta_tuning_1_week} shows the corresponding $\eta$ values in the case of 1-week rounds and FedSGD and in the case where mini-batch size is $B=20$.
In  Sec. \ref{sec:local-leakage}, we chose $\eta=0.001$ for FedSGD and we set $\eta=10^{-5}$ when $B=20$ to minimize the RMSE and then measure the privacy leakage.

\begin{figure}[t!]
	\centering
	\includegraphics[width=0.6\linewidth]{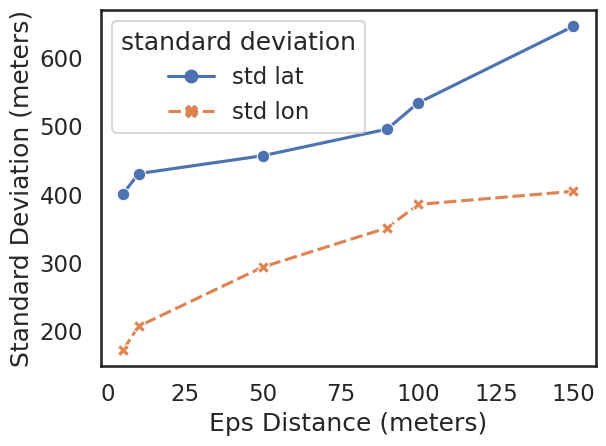}
	\caption{{\small Average standard deviation (in meters) per batch for each coordinate increases based on the $eps$ parameter of DBSCAN which controls the maximum distance between points in a cluster. This motivate us towards the \algoname for creating batches with high variance.} }
	 \vspace{-5pt}
	\label{fig:uci_std_vs_eps}

\end{figure}

\textbf{Tuning of DBSCAN parameter in \algoname.} Fig. \ref{fig:uci_std_vs_eps} shows the standard deviation for each coordinate and how it increases when we increase the $eps$ parameter of DBSCAN. Thus, higher $eps$ corresponds to fewer clusters obtained which also increases the location variance of the batch.

\begin{figure}[t!]
		\centering
	\begin{subfigure}{0.82\linewidth}
		\centering
    \includegraphics[width=0.7\linewidth]{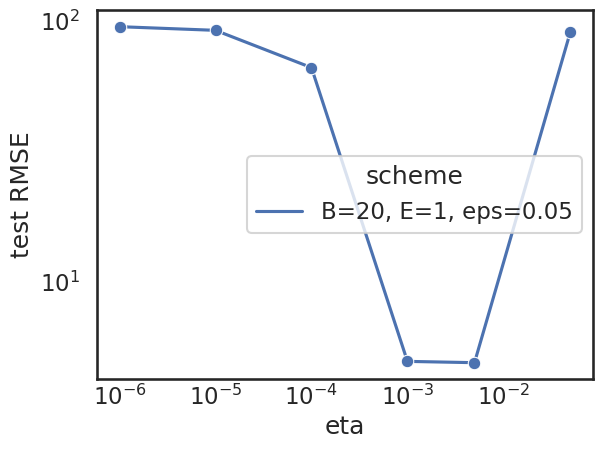}
		\caption{RMSE vs. various $\eta$.} 
		\label{fig:batch_manipeta_comparison_EMD}
		\end{subfigure}
	\begin{subfigure}{0.82\linewidth}
		\centering
    \includegraphics[width=0.99\linewidth]{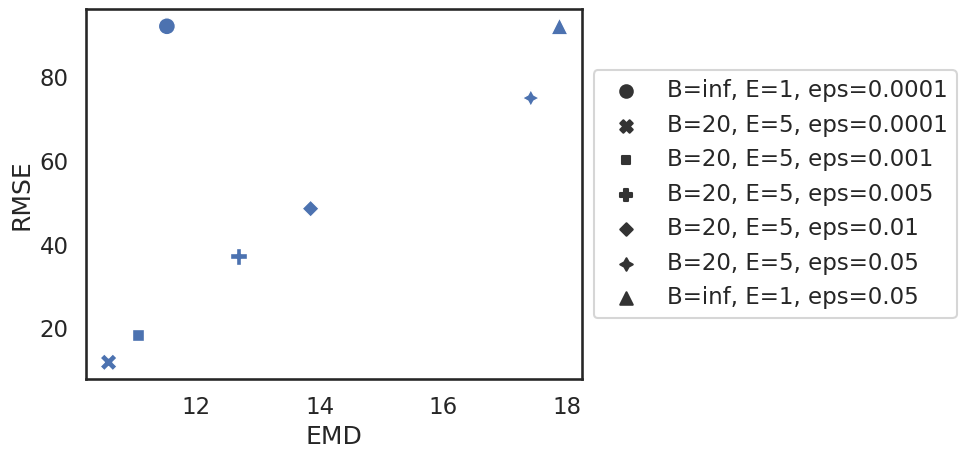}
		\caption{RMSE vs. EMD for various eps when $\eta=10^{-5}$. }
		\label{fig:batch_manip_1e_5_comparison_EMD}
		\end{subfigure}
\caption{\small \textbf{\algoname with various $\eta$.} For lower eta, the EMD behaves similarly but RMSE is higher. We choose $\eta=0.001$.}
\vspace{-15pt}
\label{fig:1week_batch_manip_etas_vs_1e_05}
\end{figure}

\textbf{Tuning $\eta$ in \algoname.} Fig. \ref{fig:batch_manipeta_comparison_EMD} shows how the selected $\eta$ minimizes RMSE in case of \algoname. Fig \ref{fig:batch_manip_1e_5_comparison_EMD} shows the results for  $\eta=$1e-05, which increases RMSE but the privacy patterns are similar to the tuned $\eta$.

\blue{\textbf{Analysis of DLG Label Initialization.} As mentioned in Sec. \ref{sec:setup}, the attacker may not have access to all measurements and may only be able to collect a limited number of measurements around each cell tower. Thus, the attacker may obtain different RSRP than the mean RSRP from the training data. To explore the impact of different RSRP initializations, we conduct an analysis where we randomly select different percentages of measurements around the cell tower. This analysis allows us to investigate how variations in the mean RSRP affect the attack's performance.} 

\blue{In particular, we perform four different RSRP initializations for the attacker at cell ID x204, including mean RSRP of all training data from target user, mean RSRP of all data from cell x204, mean RSRP of randomly selected data (random 1\% of all data from cell x204), mean RSRP of randomly selected data (random 10\% of all data from cell x204). The initial RSRP values are -95.96, -95.98, -96.048, -96.0685 (dB), respectively. According to our evaluation results, the attack performance for all above four RSRP initializations are the same: RMSE: 4.88, EMD :11.33, DIST: 230.57. Our attack results indicate that, even with randomly selected percentages of measurements, the mean RSRP values remain quite close to the mean RSRP of training data and the  attack converges to the same reconstructed location.
These results suggest that the attack's performance is robust to variations in the mean RSRP introduced by the attacker's limited knowledge.}

\end{appendix}

\section{Comparison to Baselines}\label{sec:Differential privacy Comparison}

\blue{This appendix extends Section \ref{sec:DP}, which compares our local batch selection algorithms (\algoname and \improalgoname) to three state-of-the-art baselines, \ie DP,  GeoInd and Gan, applied as a local privacy-preserving defense against DLG attacks. Figure \ref{fig:All_Comparison_1day_dp} compared our best defense \improalgoname with DP, GeoInd and Gan on top of two federated learning strategies (FedSGD and FedAvg). We report $RMSE$ as the utility metric; and $EMD$ in (a) and $Distance$ $from$ $centroid$ in (b), as privacy metrics separately. This appendix provides additional results }

\begin{figure*}%
    \vspace{-20pt}
	\centering
    \begin{subfigure}{0.4\linewidth}
	    \centering
        \includegraphics[width=0.95\linewidth]{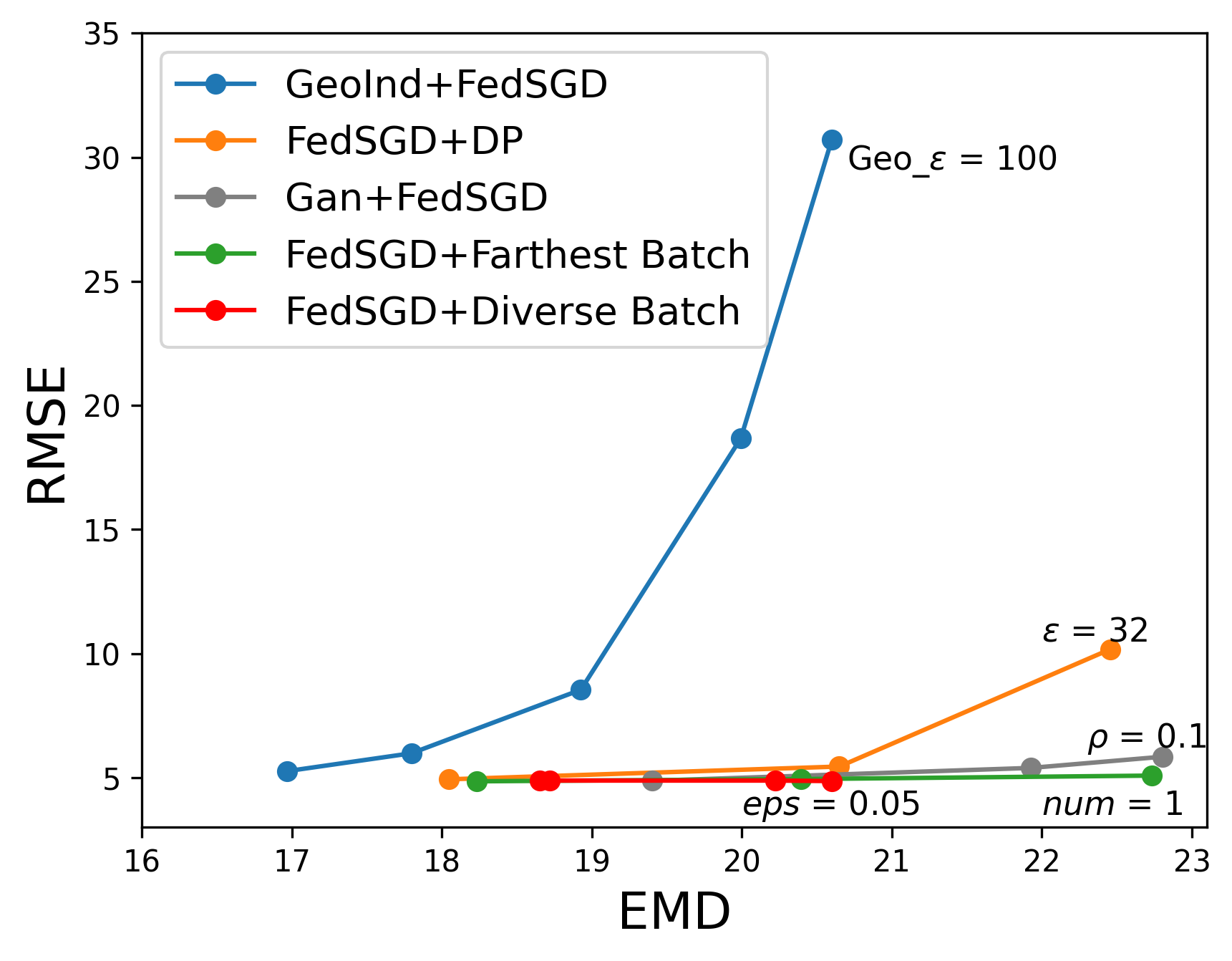}
    	\caption{\blue{\small RMSE vs EMD.}}
    	\label{fig:appen_DBSCAN_Comparison_1day_EMD_sgd}
    	\vspace{-5pt}
    \end{subfigure}
	\begin{subfigure}{0.4\linewidth}
	    \centering
        \includegraphics[width=0.95\linewidth]{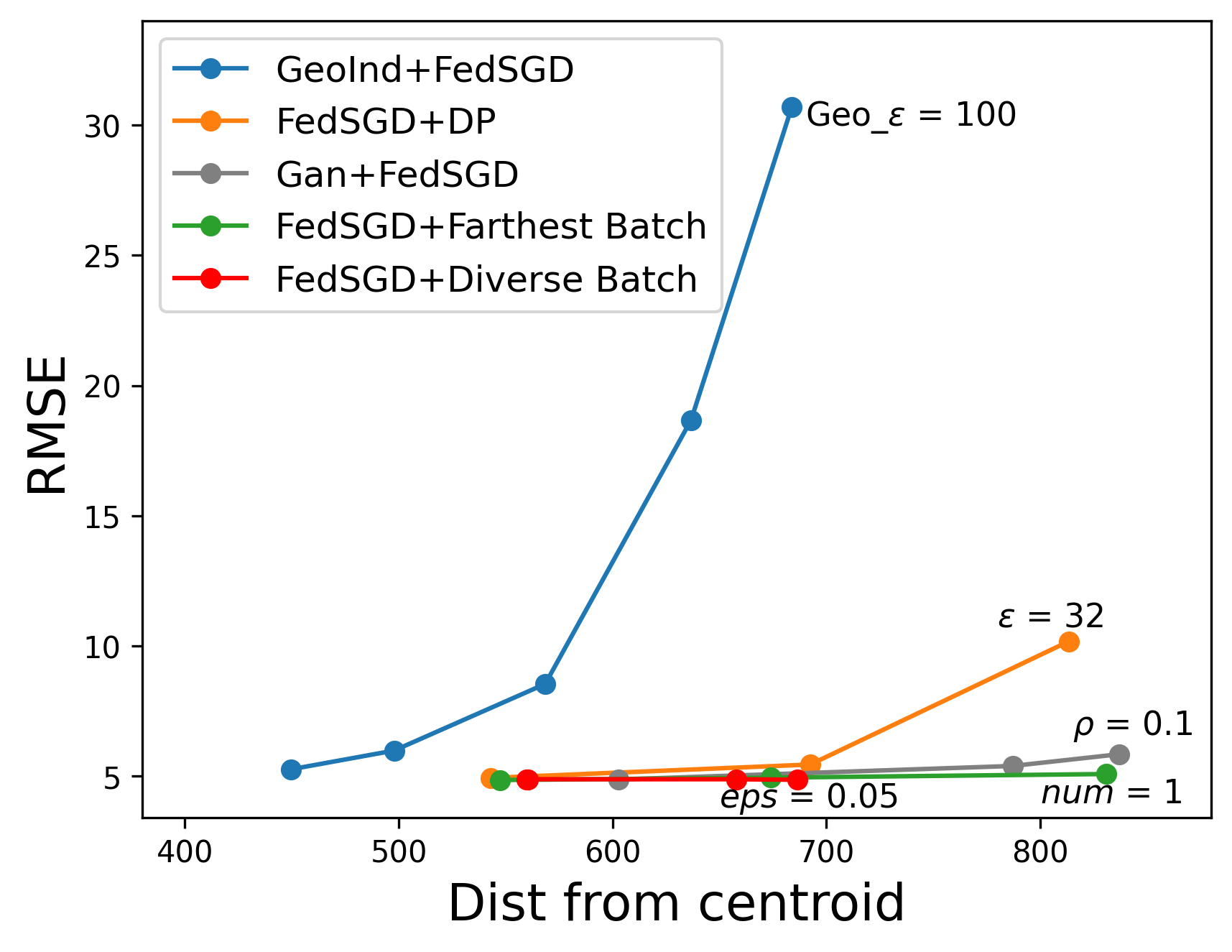}
    	\caption{\blue{\small RMSE vs Distance from centroid}}
    	\label{fig:appen_DBSCAN_Comparison_1day_DIST_sgd}
    	\vspace{-5pt}
    \end{subfigure}
    \caption{\blue{\small Comparison between \algoname, \improalgoname, GeoInd, Gan and DP for 1-day interval using FedSGD.}}
    \vspace{-20pt}
    \label{fig:appendix_Comparison_all_1day_sgd}
\end{figure*}

\begin{figure*}%
	\centering
    \begin{subfigure}{0.4\linewidth}
	    \centering
        \includegraphics[width=0.95\linewidth]{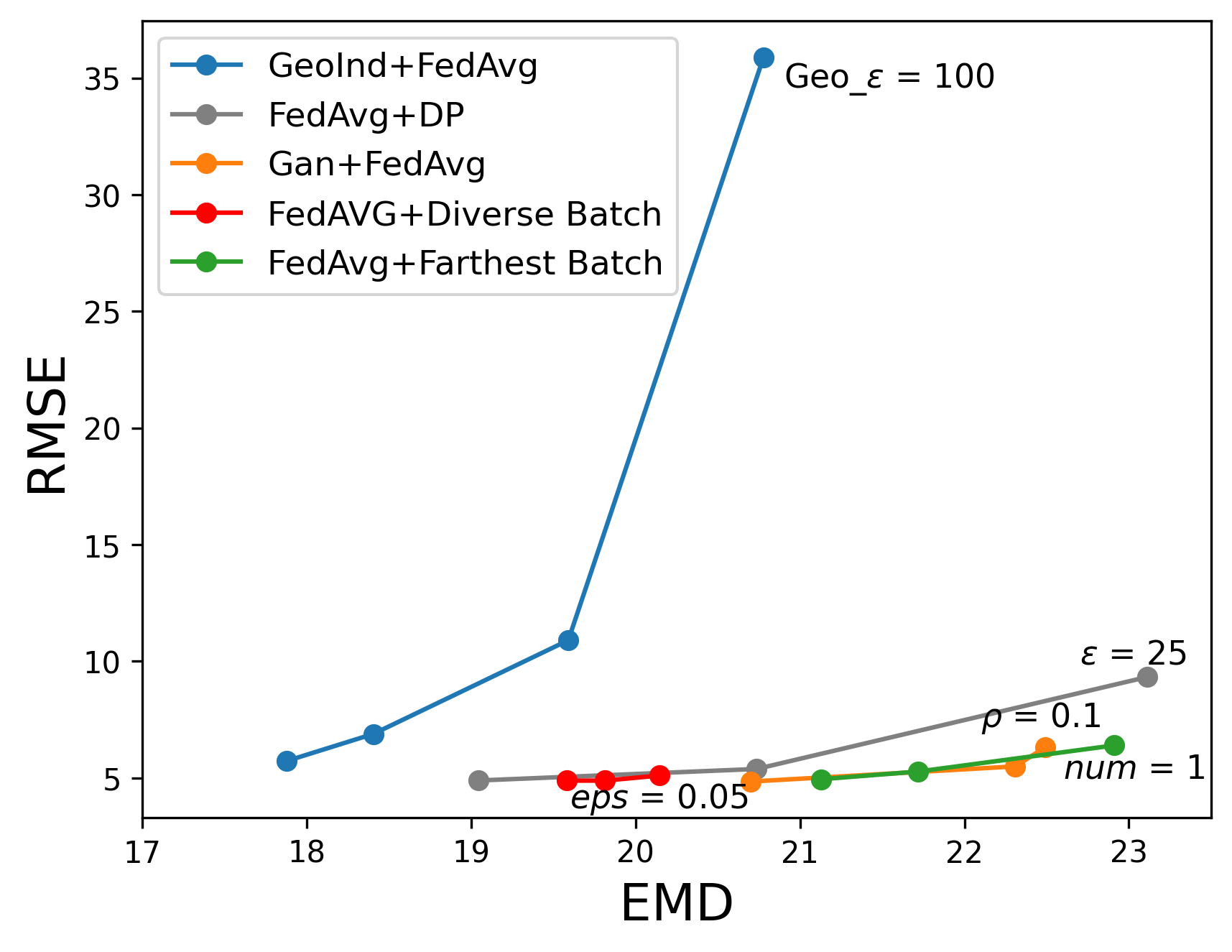}
    	\caption{\blue{\small RMSE vs EMD.}}
    	\label{fig:appen_All_Comparison_1day_EMD_Avg}
    	\vspace{-5pt}
    \end{subfigure}
	\begin{subfigure}{0.4\linewidth}
	    \centering
        \includegraphics[width=0.95\linewidth]{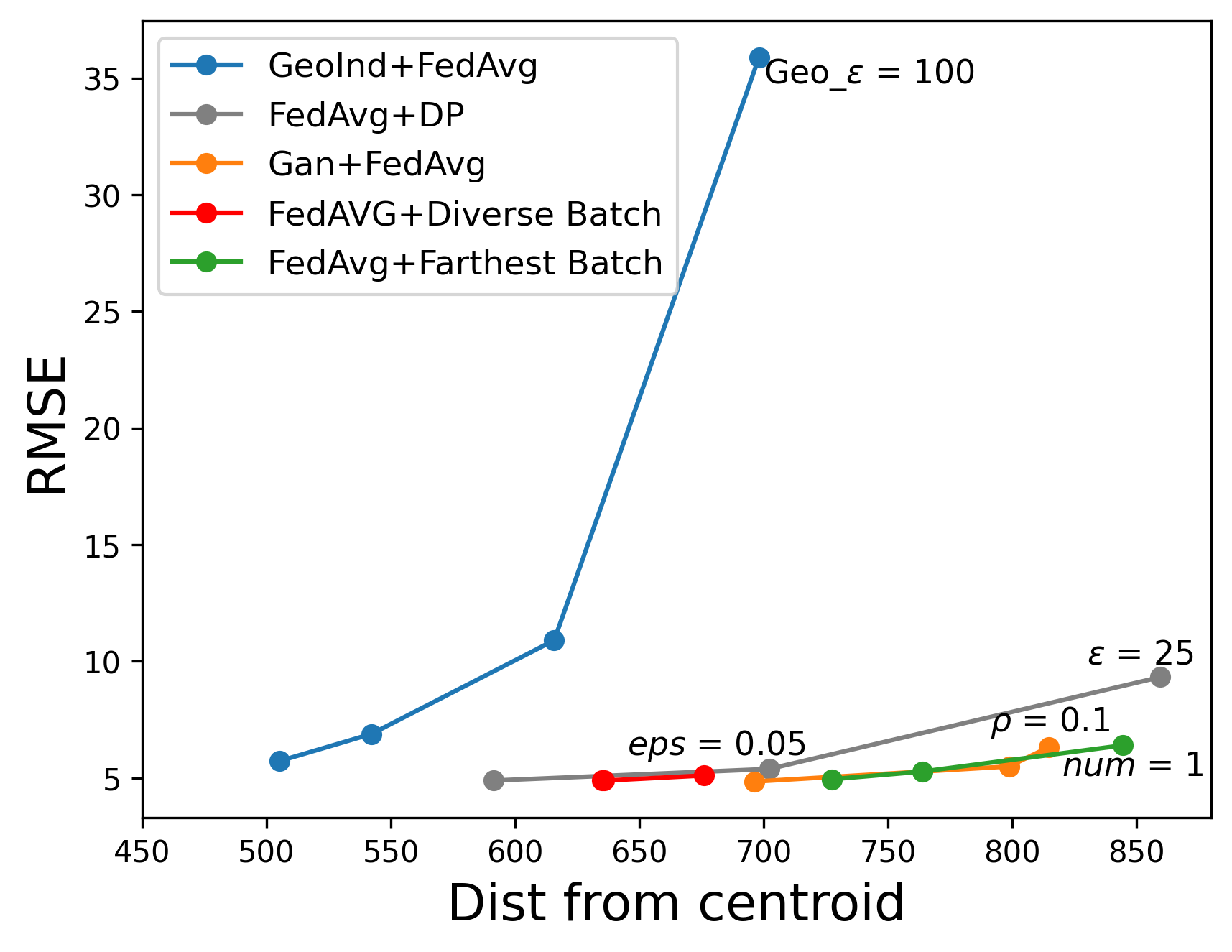}
    	\caption{\blue{\small RMSE vs. Distance from centroid.}}
    	\label{fig:appen_All_Comparison_1day_DIST_Avg}
    	\vspace{-5pt}
    \end{subfigure}
    \caption{\blue{\small Comparison between \algoname, \improalgoname, GeoInd, Gan and DP for 1-day interval using FedAvg.}}
    \vspace{-30pt}
    \label{fig:appendix_Comparison_all_1day_Avg}
\end{figure*}

\begin{figure*}%
	\centering
    \begin{subfigure}{0.4\linewidth}
	    \centering
        \includegraphics[width=0.95\linewidth]{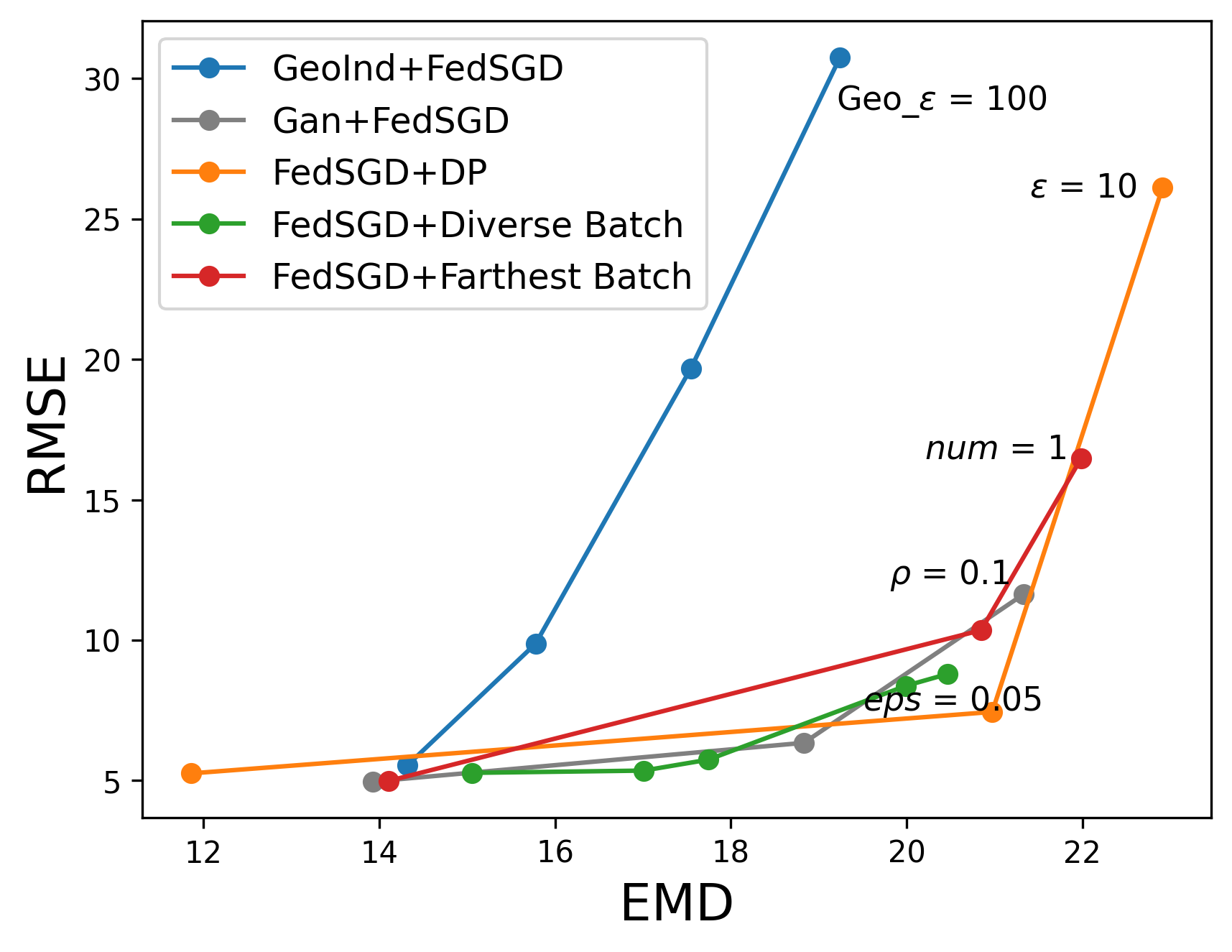}
    	\caption{\blue{\small RMSE vs. EMD.}}
    	\label{fig:appen_All_Comparison_1week_EMD_SGD}
    	\vspace{-5pt}
    \end{subfigure}
	\begin{subfigure}{0.4\linewidth}
	    \centering
        \includegraphics[width=0.95\linewidth]{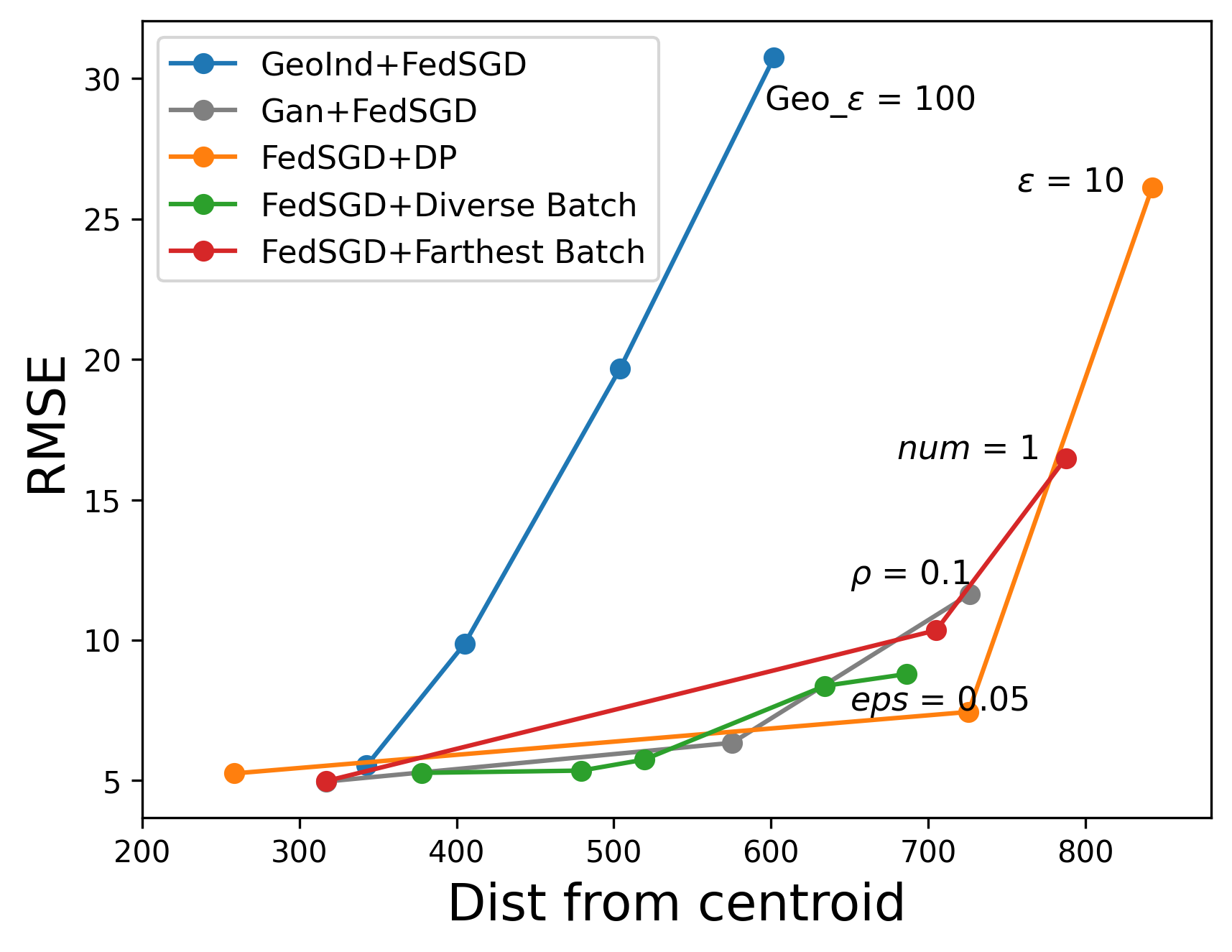}
    	\caption{\blue{\small RMSE vs. Distance from centroid.}}
    	\label{fig:appen_All_Comparison_1week_DIST_SGD}
    	\vspace{-5pt}
    \end{subfigure}
    \caption{\blue{\small Comparison between \algoname, \improalgoname, GeoInd, Gan and DP for 1-week interval using FedSGD.}}
    \vspace{-30pt}
    \label{fig:appendix_Comparison_all_1week_SGD}
\end{figure*}

\begin{figure*}%
	\centering
    \begin{subfigure}{0.4\linewidth}
	    \centering
        \includegraphics[width=0.95\linewidth]{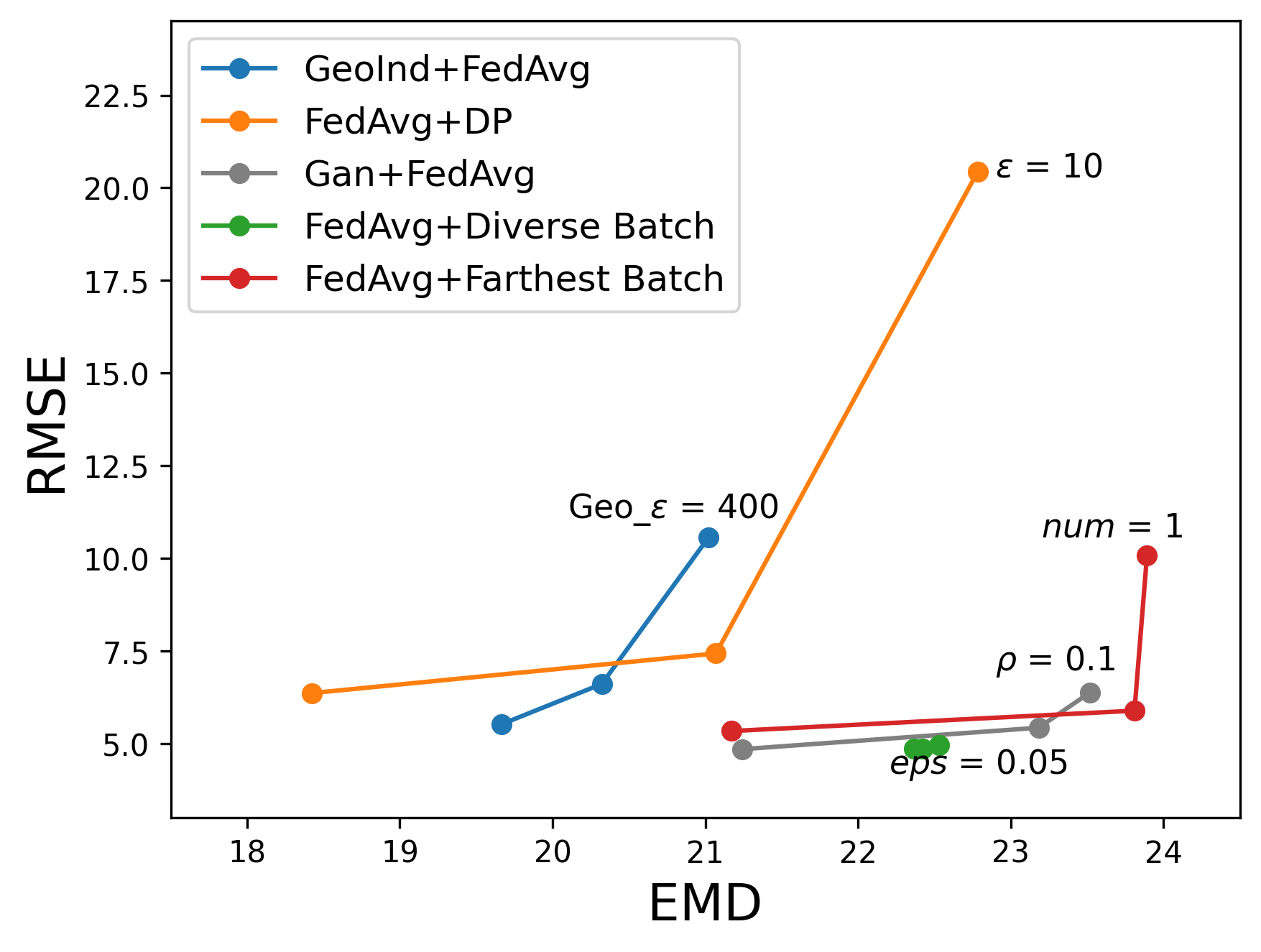}
    	\caption{\blue{\small RMSE vs. EMD.}}
    	\label{fig:appen_All_Comparison_1week_EMD_Avg}
    	\vspace{-5pt}
    \end{subfigure}
	\begin{subfigure}{0.4\linewidth}
	    \centering
        \includegraphics[width=0.95\linewidth]{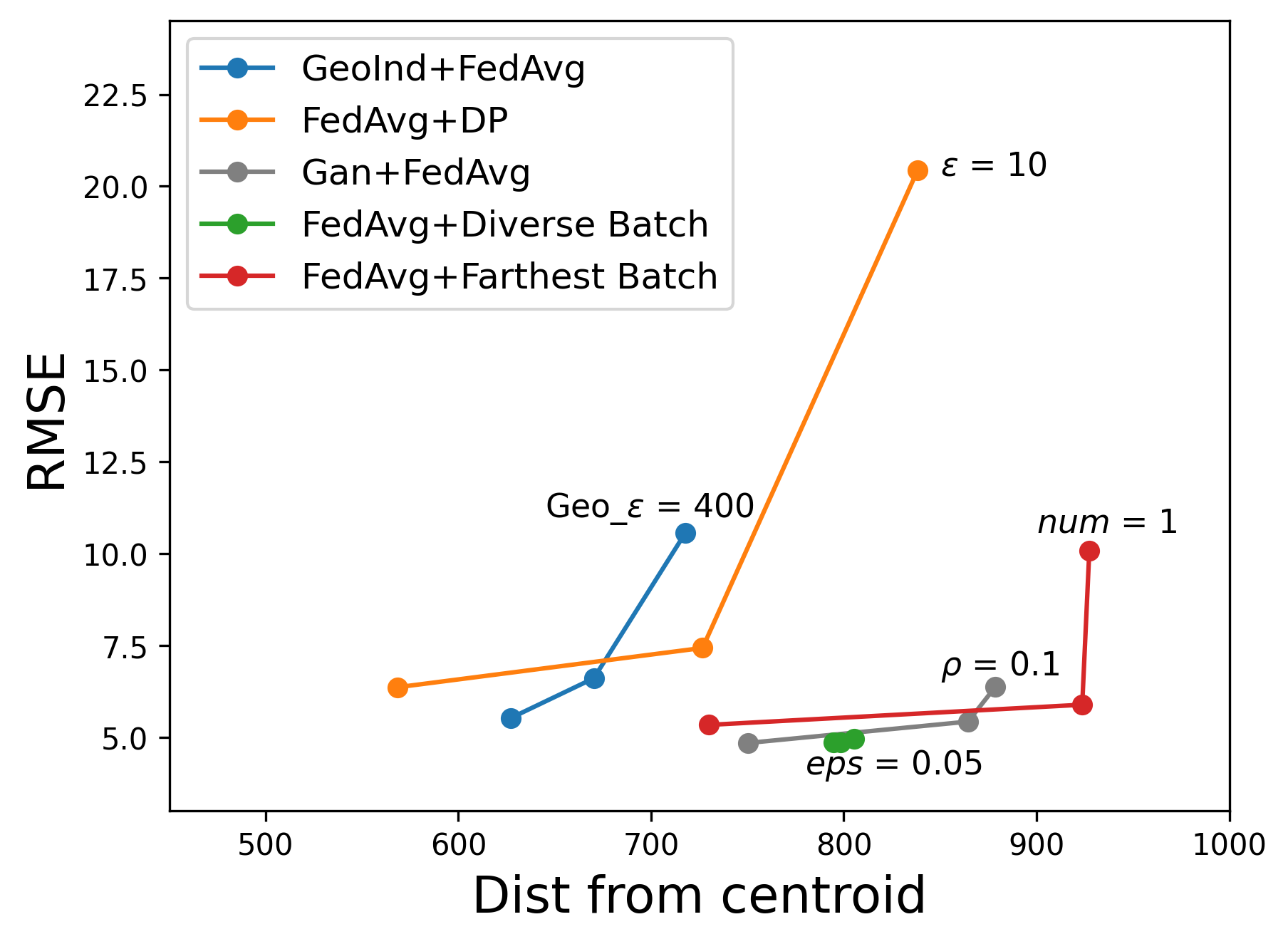}
    	\caption{\blue{\small RMSE vs. Distance from centroid.}}
    	\label{fig:appen_All_Comparison_1week_DIST_Avg}
    	\vspace{-5pt}
    \end{subfigure}
    \caption{\blue{\small Comparison between \algoname, \improalgoname, GeoInd, Gan and DP for 1-week interval using FedAvg.}}
    \label{fig:appendix_Comparison_all_1week_Avg}
    \vspace{-50pt}
\end{figure*}

\begin{figure}[t!]
    \begin{subfigure}{.4\textwidth}
        \centering
        \includegraphics[width=.99\linewidth]{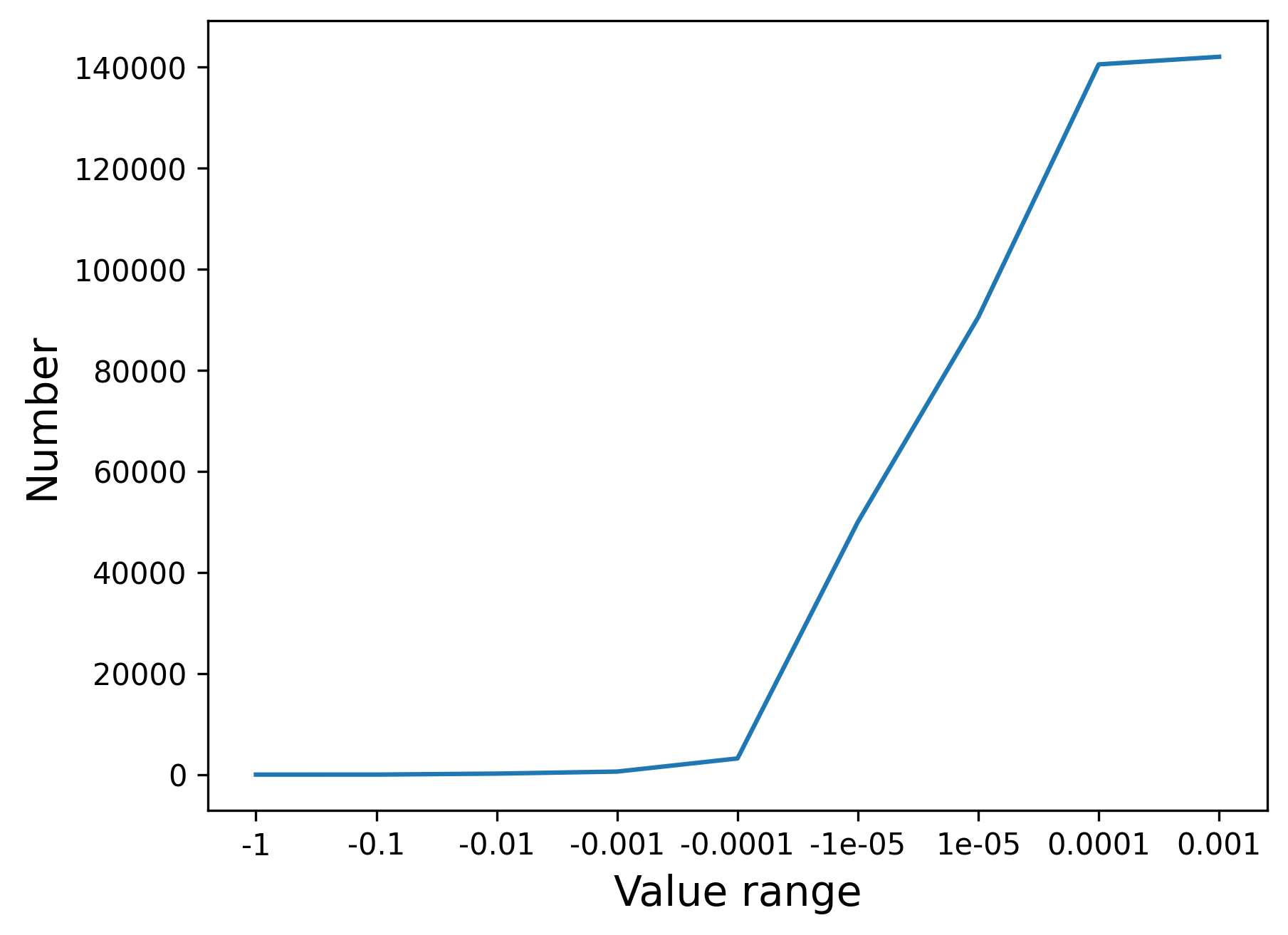}
        \caption{Without adding DP noise.}
        \label{fig:dpanalysis_1}
    \end{subfigure}
    \begin{subfigure}{.4\textwidth}
        \centering
        \includegraphics[width=.99\linewidth]{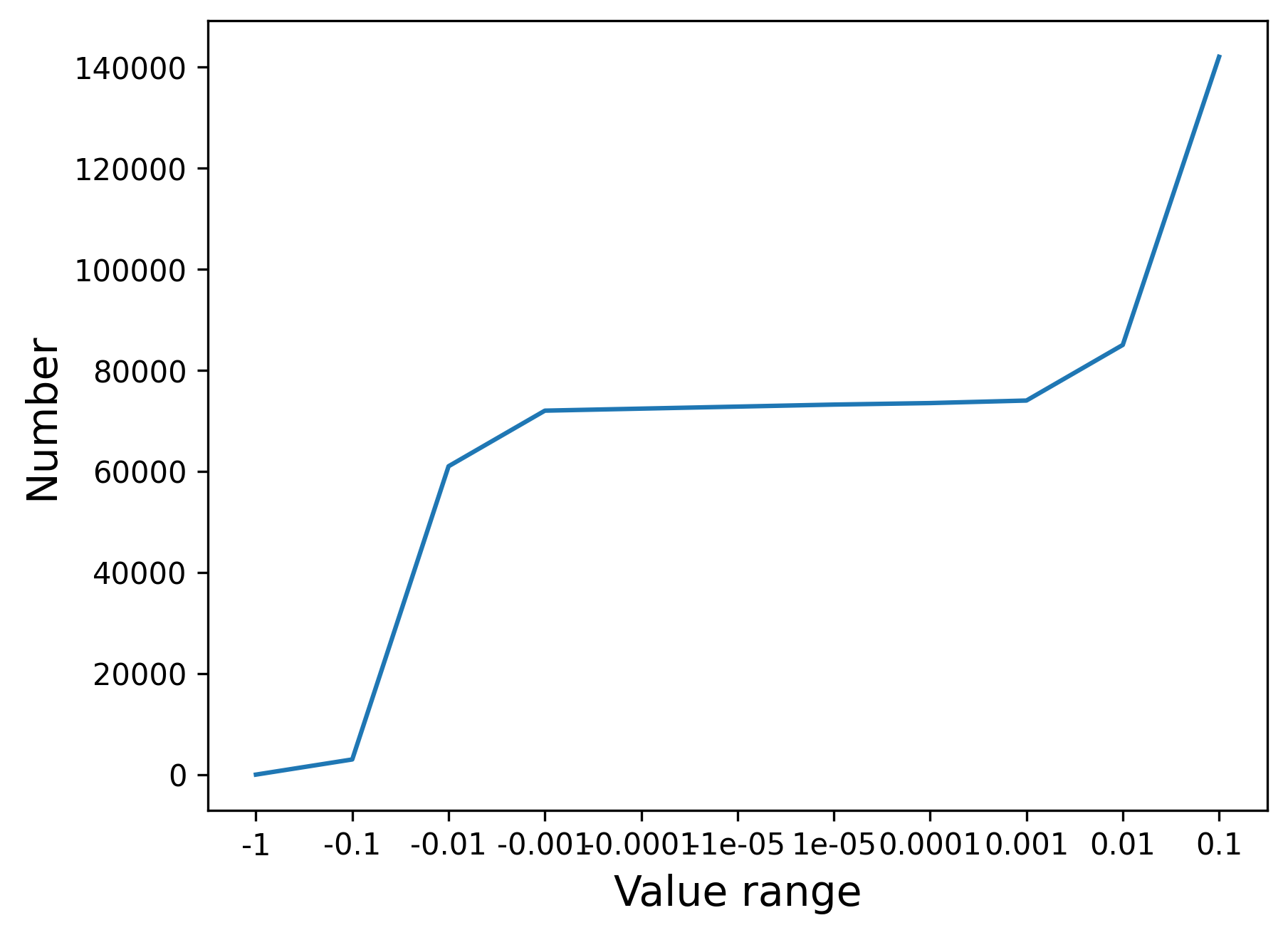}
        \caption{Adding DP noise when $epsilon=100$.}
        \vspace{-9pt}
        \label{fig:dpanalysis_2}
    \end{subfigure}
    \caption{CDF of gradient elements.}
\label{fig:dpanalysis}
\end{figure}

\subsection{\blue{Comparison to DP, Geoind and Gan}}
\blue{

As shown in Figure \ref{fig:appendix_Comparison_all_1day_sgd} and Figure \ref{fig:appendix_Comparison_all_1day_Avg}, \algoname and \improalgoname achieve a better tradeoff than differential privacy and GeoInd in both FedSGD and FedAvg for the 1-day interval. With the same level of privacy, in terms of $EMD$ and $Distance$ $from$ $centroid$, \algoname and \improalgoname could provide better utility than differential privacy and GeoInd. Figure \ref{fig:appendix_Comparison_all_1week_SGD} and Figure \ref{fig:appendix_Comparison_all_1week_Avg} show the privacy and utility tradeoff for the 1-week interval in FedSGD and FedAvg. \algoname and \improalgoname can partially achieve a better tradeoff than DP in the FedSGD setting (see Fig. \ref{fig:appendix_Comparison_all_1week_SGD}). In the FedAvg setting (see Fig. \ref{fig:appendix_Comparison_all_1week_Avg}), \algoname and \improalgoname outperforms DP thoroughly. Compared to GeoInd, \algoname and \improalgoname could achieve a better tradeoff in both FedSGD and FedAvg for 1-week interval. As mentioned in Sec. \ref{sec:DP}, although Gan could provide similar utility as \algoname and \improalgoname for the same privacy level, Gan requires more computation resources and our local batch selection approach (\algoname and \improalgoname) is more efficient and lightweight. Also as illustrated in Fig. \ref{fig:appendix_Comparison_all_1day_sgd}, Fig. \ref{fig:appendix_Comparison_all_1day_Avg}, Fig. \ref{fig:appendix_Comparison_all_1week_SGD}, and Fig. \ref{fig:appendix_Comparison_all_1week_Avg}, compared to \algoname, \improalgoname could better mislead the adversary and enhance the privacy. }

\subsection{Additional DP Analysis}
Comparing Figure \ref{fig:dpanalysis_1} and Figure \ref{fig:dpanalysis_2}, we can observe that there are lots of gradient elements with values smaller than 0.0001 after clipping. Hence, after adding DP noise with $\epsilon=100$, i.e., Gaussian noise with $\sigma=0.048$, it will disturb those gradient elements with small values. This explains why adding DP noise with $\epsilon=100$ can still reduce the utility in terms of RMSE. Moreover, as shown in Eq. (\ref{proof-eq3}), adding noise to the individual gradient is equivalent to adding noise into $x_i$, which will increase the variance of $x_i$ and hence increase the upper bound of DLG attacker's reconstruction error.

\vfill

\end{document}